\pdfoutput=1

\documentclass[11pt]{article}

\usepackage[final]{acl}

\usepackage{times}
\usepackage{latexsym}

\usepackage[T1]{fontenc}

\usepackage[utf8]{inputenc}

\usepackage{microtype}

\usepackage{inconsolata}
\usepackage{graphicx}
\usepackage{amsmath}
\usepackage{amssymb}
\usepackage{amsthm}
\usepackage{booktabs}
\usepackage{algorithm}
\usepackage{algorithmic}
\usepackage{pifont}
\usepackage{xcolor}
\usepackage{color}
\usepackage{colortbl}
\usepackage{lipsum}

\definecolor{Gray}{gray}{0.9}

\def \ie {\emph{i.e.}}
\def \eg {\emph{e.g.}}

\newcommand{\cmark}{\ding{51}}%
\newcommand{\xmark}{\ding{55}}%
\newcommand{\amark}{\ding{91}}%

\newcommand{\tit}[1]{\smallbreak\noindent\textbf{#1.}}
\newcommand{\tinytit}[1]{\noindent\textbf{#1.}}

\newcommand{\titt}[1]{\smallbreak\noindent\textbf{#1}}

\title{The Revolution of Multimodal Large Language Models: A Survey}

\author{Davide Caffagni\textsuperscript{1}\footnotemark[1], Federico Cocchi\textsuperscript{1,2}\footnotemark[1], Luca Barsellotti\textsuperscript{1}\footnotemark[1], \textbf{Nicholas Moratelli}\textsuperscript{1}\footnotemark[1],\\
\textbf{Sara Sarto\textsuperscript{1}\footnotemark[1]}, \textbf{Lorenzo Baraldi\textsuperscript{2}\footnotemark[1]}, \textbf{Lorenzo Baraldi\textsuperscript{1}}, \textbf{Marcella Cornia\textsuperscript{1}}, \and \textbf{Rita Cucchiara\textsuperscript{1,3}} \\
\textsuperscript{1}University of Modena and Reggio Emilia, Italy \\
\textsuperscript{2}University of Pisa, Italy \\
\textsuperscript{3}IIT-CNR, Italy \\
\textsuperscript{1}\texttt{\{name.surname\}@unimore.it} \quad \textsuperscript{2}\texttt{\{name.surname\}@phd.unipi.it}
}

\begin{document}
\maketitle
\renewcommand{\thefootnote}{\fnsymbol{footnote}}
\footnotetext[1]{Equal contribution.}

\maketitle
\begin{abstract}
Connecting text and visual modalities plays an essential role in generative intelligence. For this reason, inspired by the success of large language models, significant research efforts are being devoted to the development of Multimodal Large Language Models (MLLMs). These models can seamlessly integrate visual and textual modalities, while providing a dialogue-based interface and instruction-following capabilities. In this paper, we provide a comprehensive review of recent visual-based MLLMs, analyzing their architectural choices, multimodal alignment strategies, and training techniques. We also conduct a detailed analysis of these models across a wide range of tasks, including visual grounding, image generation and editing, visual understanding, and domain-specific applications. Additionally, we compile and describe training datasets and evaluation benchmarks, conducting comparisons among existing models in terms of performance and computational requirements. Overall, this survey offers a comprehensive overview of the current state of the art, laying the groundwork for future MLLMs.
\end{abstract}

\section{Introduction}
\label{sec:intro}
The introduction of the attention operation and the Transformer architecture~\cite{vaswani2017attention} has enabled the creation of models capable of handling various modalities on an increasingly large scale. This advancement is largely attributed to the versatility of the operator and the adaptability of the architecture. Initially, this breakthrough was leveraged for language-specific models~\cite{devlin2018bert,brown2020language} but quickly extended to support diverse modalities~\cite{li2019visualbert,lu2019vilbert} and facilitate their integration within unified embedding spaces~\cite{radford2021learning}.

\begin{figure}
\centering
\includegraphics[width=0.988\linewidth]{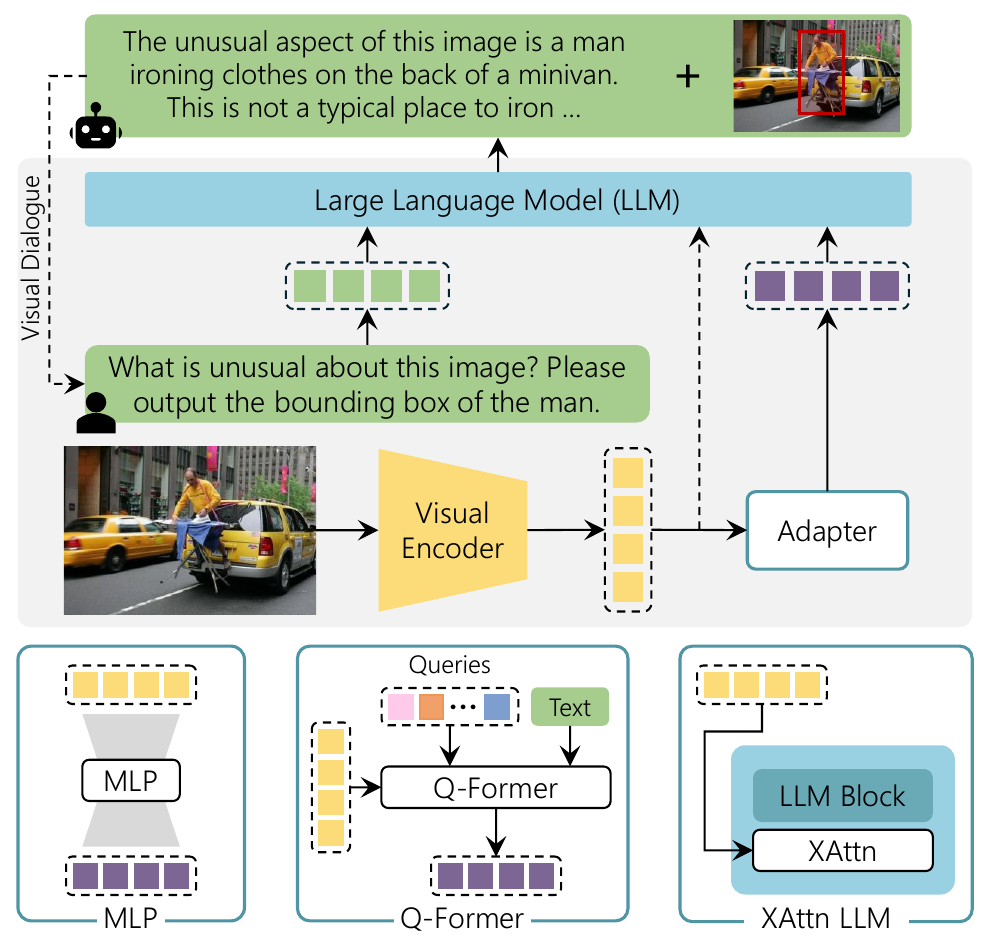}
\vspace{-.2cm}
\caption{General architecture of Multimodal Large Language Models (MLLMs), composed of a visual encoder, a language model, and an adapter module that connects visual inputs to the textual space.}
\label{fig:multimodal_llm}
\vspace{-.4cm}
\end{figure}

The surge in sophisticated Large Language Models (LLMs), particularly their capacity for in-context learning, has encouraged researchers to broaden the scope of these models to encompass multiple modalities, both as inputs and outputs. This expansion has led to the development of cutting-edge models such as GPT-4V~\cite{achiam2023gpt} and Gemini~\cite{team2023gemini}, showcasing state-of-the-art performance.

The development of Multimodal Large Language Models (MLLMs) entails merging single-modality architectures for vision and language, establishing effective connections between them through vision-to-language adapters, and devising innovative training approaches. These methodologies are crucial for ensuring modality alignment and the ability to follow instructions accurately.

In a context marked by the rapid release of new models, our goal is to offer an exhaustive overview of the MLLM landscape, with a focus on models exploiting the visual modality. This overview serves as both an update on the current state and a source of inspiration for future developments. We identify three core aspects that define these models: their architecture, training methodologies, and the tasks they are designed to perform. We begin by detailing the prevalent choices for vision encoders and adapter modules that equip LLMs with cross-modal capabilities. Following this, we delve into the training processes and data utilized. We then explore the range of tasks addressed by MLLMs. The review concludes with a discussion of the persisting challenges in the field and the promising directions for future research. Further details on training data, evaluation datasets, performance and computational requirements are reported in the supplementary material.

The motivation behind this survey stems from an emerging scientific interest in the field of MLLMs, as evidenced by the constant increase in published works. In comparison with existing surveys on the topic~\cite{yin2023survey,wu2023multimodal,huang2023visual}, our paper exhibits substantial differences. Notably, it addresses several critical areas that were overlooked in prior works, including visual grounding, image generation, and editing. Furthermore, our survey details the main components utilized by each discussed MLLM, such as the visual encoders and the specific LLM employed. Additionally, our analysis offers a comparative perspective on the performance and hardware requirements of the discussed papers, incorporating both quantitative results and detailed information on benchmarks. Through this comprehensive approach, our survey aims to fill the existing gaps and provide a more nuanced understanding of the current landscape in the field.

\section{Empowering LLMs with Multimodal Capabilities}
\label{sec:architectures}
\subsection{Preliminaries}
\label{sec:preliminaries}
\tinytit{Large Language Models}~\citet{brown2020language} discovered that in-context learning, \ie, prepending the prompt with a few examples demonstrating the desired output of an LLM~\cite{chowdhery2023palm,hoffmann2022training,tay2022unifying}, improves its performance, especially over unseen tasks. Generalization can be further enhanced by providing the LLM with the natural language description of the desired task for each training sample. This technique, called instruction-tuning~\cite{chung2022scaling,wang2022super,wang2022self, jiang2024mixtral}, turns out to be critical for aligning the behavior of an LLM with that of humans and currently empowers the most advanced LLMs, eventually boosted via reinforcement learning from human feedback (RLHF)~\cite{ouyang2022training, achiam2023gpt,chen2023dress,bai2023qwentc}.

\tit{PEFT} When a pre-trained LLM needs to be adapted to a specific domain or application, parameter-efficient fine-tuning (PEFT) schemes represent an important alternative to train the entire LLM, since these strategies only introduce a few new parameters. Among these, prompt-tuning~\cite{hambardzumyan2021warp,lester2021power,li2021prefix,liu2023gpt} learns a small set of vectors to be fed to the model as soft prompts before the input text. Differently, LoRA~\cite{hu2021lora} constrains the number of new weights by learning low-rank matrices. This technique is orthogonal to quantization methods such as QLoRA~\citep{dettmers2023qlora}, which further decreases the memory footprint of the LLM compared to the usual half-precision weights.

\tit{Towards Multimodal LLMs} The development of MLLMs follows a similar path to that of LLMs, with Flamingo~\cite{alayrac2022flamingo} being the first to explore in-context learning at scale in the vision-language field. Then, visual instruction-tuning~\cite{liu2023visual} quickly became the most prominent training paradigm also in the multimodal domain, as well as the use of PEFT techniques to fine-tune the LLM. Any MLLM contains at least three components (Fig.~\ref{fig:multimodal_llm}): an LLM backbone serving as an interface with the user, one (or more) visual encoders, and one or more vision-to-language adapter modules. Popular choices for the LLM backbone often fall into the LLaMA family~\cite{touvron2023llama,touvron2023llama2}, given that their weights are freely accessible, they have been trained on public data solely, and they boast different sizes to accommodate various use cases. In addition, their derivative versions are popular as well, such as Alpaca~\cite{taori2023stanford} and Vicuna~\cite{vicuna2023}. The former fine-tunes LLaMA on instructions written using GPT-3, while the latter exploits user-shared conversations with ChatGPT~\cite{chatgpt}. Alternatives are OPT~\cite{zhang2022opt}, Magneto~\cite{wang2023magneto}, MPT~\cite{Introduc1online}, and the instruction-tuned~\cite{chung2022scaling} or multilingual~\cite{xue2020mt5} flavors of T5~\cite{raffel2020exploring}, an encoder-decoder language model pre-trained for multiple tasks.

\tit{Pre-Training of Model Components}
The main components of MLLMs are the visual encoder and the language model. The visual encoder is designed to provide LLMs with visual information and the most used ones are CLIP-based architectures~\cite{radford2021learning,wortsman2022robust} whose pre-training objective is the alignment between CLIP embeddings, obtained thanks to a contrastive loss that aligns the correct image-text pairs. An exception is the EVA-CLIP models family~\cite{fang2023eva}, which exploits a MAE pre-training strategy~\cite{he2022masked} to reconstruct the masked-out image-text aligned visual features, conditioned on visible image patches. On the other hand, LLMs primarily rely on the widely employed Transformer model, although the Mamba architecture~\cite{gu2023mamba} has also emerged in recent times. This proposes to make a State-Space Model (SSM) time-dependent, effectively creating a selective SSM with favorable properties: (i) inference costs and memory requirements that scale linearly with the sequence length, and (ii) efficient parallel training thanks to a smart GPU implementation of the algorithm. Similar to Transformers, Mamba models for language modeling are pre-trained using the next token prediction task. Very recent studies propose MLLMs featuring Mamba as the language backbone~\cite{qiao2024vl,zhao2024cobra}.

\smallbreak
\noindent
A summary of the MLLMs covered in this survey is reported in Table~\ref{tab:models}, indicating for each model the LLM on which it is based, the visual encoder, the adapter used to connect visual and language components, whether the MLLM is trained with visual instruction tuning or not, and a short list of the main tasks and capabilities.

\begin{table*}
\centering
\setlength{\tabcolsep}{.2em}
\resizebox{\linewidth}{!}{
\begin{tabular}{l cccc cl}
\toprule
& & \textbf{Visual} & \textbf{V2L} & \textbf{VInstr.} \\
\textbf{Model} & \textbf{LLM} & \textbf{Encoder} & \textbf{Adapter} & \textbf{Tuning} & & \textbf{Main Tasks \& Capabilities} \\
\midrule
BLIP-2~\cite{li2023blip} & FlanT5-XXL-11B$^\bigstar$ & EVA ViT-g & Q-Former & \xmark & & Visual Dialogue, VQA, Captioning, Retrieval \\
FROMAGe~\cite{koh2023grounding} & OPT-6.7B$^\bigstar$ & CLIP ViT-L & Linear & \xmark & & Visual Dialogue, Captioning, Retrieval \\
Kosmos-1~\cite{huang2023language} & Magneto-1.3B$^\lozenge$ & CLIP ViT-L & Q-Former\textsuperscript{\amark} & \xmark & & Visual Dialogue, VQA, Captioning \\
LLaMA-Adapter V2~\cite{gao2023llama} & LLaMA-7B$^\blacktriangle$ & CLIP ViT-L & Linear & \xmark & & VQA, Captioning \\
OpenFlamingo~\cite{awadalla2023openflamingo} & MPT-7B$^\bigstar$ & CLIP ViT-L & XAttn LLM & \xmark & & VQA, Captioning \\
\rowcolor{Gray}
Flamingo~\cite{alayrac2022flamingo} & Chinchilla-70B$^\bigstar$ & NFNet-F6 & XAttn LLM & \xmark & & Visual Dialogue, VQA, Captioning \\
\rowcolor{Gray}
PaLI~\cite{chen2023pali} & mT5-XXL-13B$^\blacklozenge$ & ViT-e & XAttn LLM & \xmark & & Multilingual, VQA, Captioning, Retrieval \\
\rowcolor{Gray}
PaLI-X~\cite{chen2023palix} & UL2-32B$^\blacklozenge$ & ViT-22B & XAttn LLM & \xmark & & Multilingual, VQA, Captioning \\
\midrule
LLaVA~\cite{liu2023visual} & Vicuna-13B$^\blacklozenge$ & CLIP ViT-L & Linear & \cmark & & Visual Dialogue, VQA, Captioning \\
MiniGPT-4~\cite{zhu2023minigpt} & Vicuna-13B$^\bigstar$ & EVA ViT-g & Linear & \cmark & & VQA, Captioning \\
mPLUG-Owl~\cite{ye2023mplug} & LLaMA-7B$^\blacktriangle$ & CLIP ViT-L & Q-Former\textsuperscript{\amark} & \cmark & & Visual Dialogue, VQA \\
InstructBLIP~\cite{dai2023instructblip} & Vicuna-13B$^\bigstar$ & EVA ViT-g & Q-Former & \cmark & & Visual Dialogue, VQA, Captioning \\
MultiModal-GPT~\cite{gong2023multimodal} & LLaMA-7B$^\blacktriangle$ & CLIP ViT-L & XAttn LLM & \cmark & & Visual Dialogue, VQA, Captioning \\
LaVIN~\cite{luo2023cheap} & LLaMA-13B$^\blacktriangle$ & CLIP ViT-L & MLP & \cmark & & Visual Dialogue, VQA, Captioning \\
Otter~\cite{li2023otter} & LLaMA-7B$^\bigstar$ & CLIP ViT-L & XAttn LLM & \cmark & & VQA, Captioning \\
Kosmos-2~\cite{peng2023kosmos} & Magneto-1.3B$^\lozenge$ & CLIP ViT-L & Q-Former\textsuperscript{\amark} & \cmark & & Visual Dialogue, VQA, Captioning, Referring, REC \\
Shikra~\cite{chen2023shikra} & Vicuna-13B$^\blacklozenge$ & CLIP ViT-L & Linear & \cmark & & Visual Dialogue, VQA, Captioning, Referring, REC, GroundCap \\
Clever Flamingo~\cite{chen2023visual} & LLaMA-7B$^\blacktriangle$ & CLIP ViT-L & XAttn LLM & \cmark & & Visual Dialogue, VQA, Captioning \\
SVIT~\cite{zhao2023svit} &  Vicuna-13B$^\blacklozenge$ & CLIP ViT-L & MLP & \cmark & & Visual Dialogue, VQA, Captioning \\
BLIVA~\cite{hu2024bliva} & Vicuna-7B$^\bigstar$ & EVA ViT-g & Q-Former+Linear & \cmark & & Visual Dialogue, VQA, Captioning \\
IDEFICS~\cite{laurenccon2023obelisc} & LLaMA-65B$^\bigstar$ & OpenCLIP ViT-H & XAttn LLM & \cmark & & Visual Dialogue, VQA, Captioning \\
Qwen-VL~\cite{bai2023qwen} & Qwen-7B$^\blacklozenge$ & OpenCLIP ViT-bigG & Q-Former\textsuperscript{\amark} & \cmark & & Visual Dialogue, Multilingual, VQA, Captioning, REC \\ 
StableLLaVA~\cite{li2023stablellava} & Vicuna-13B$^\blacklozenge$ & CLIP ViT-L & Linear & \cmark & & Visual Dialogue, VQA, Captioning \\
Ferret~\cite{you2023ferret} & Vicuna-13B$^\blacklozenge$ & CLIP ViT-L & Linear & \cmark & & Visual Dialogue, Captioning, Referring, REC, GroundCap \\
LLaVA-1.5~\cite{liu2023improved} & Vicuna-13B$^\blacklozenge$ & CLIP ViT-L & MLP & \cmark & & Visual Dialogue, VQA, Captioning \\
MiniGPT-v2~\cite{chen2023minigpt} & LLaMA-2-7B$^\blacktriangle$ & EVA ViT-g & Linear & \cmark & & Visual Dialogue, VQA, Captioning, Referring, REC, GroundCap \\
Pink~\cite{xuan2023pink} & Vicuna-7B$^\blacktriangle$ & CLIP ViT-L & Linear & \cmark & & Visual Dialogue, VQA, Captioning, Referring, REC, GroundCap \\
CogVLM~\cite{wang2023cogvlm} & Vicuna-7B$^\blacklozenge$ & EVA ViT-E & MLP & \cmark & & Visual Dialogue, VQA, Captioning, REC \\
DRESS~\cite{chen2023dress} & Vicuna-13B$^\blacktriangle$ & EVA ViT-g & Linear & \cmark & & Visual Dialogue, VQA, Captioning \\
LION~\cite{chen2023lion} & FlanT5-XXL-11B$^\bigstar$ & EVA ViT-g & Q-Former+MLP & \cmark & & Visual Dialogue, VQA, Captioning, REC \\
mPLUG-Owl2~\cite{ye2023mplug2} & LLaMA-2-7B$^\blacklozenge$ & CLIP ViT-L & Q-Former\textsuperscript{\amark} & \cmark & & Visual Dialogue, VQA, Captioning \\
SPHINX~\cite{lin2023sphinx} & LLaMA-2-13B$^\blacklozenge$ & Mixture & Linear & \cmark & & Visual Dialogue, VQA, Captioning, Referring, REC, GroundCap \\ 
Honeybee~\cite{cha2023honeybee} & Vicuna-13B$^\blacklozenge$ & CLIP ViT-L & ResNet blocks & \cmark & & Visual Dialogue, VQA, Captioning \\
VILA~\cite{lin2023vila} & LLaMA-2-13B$^\blacklozenge$ & CLIP ViT-L & Linear & \cmark & & Visual Dialogue, VQA, Captioning \\
SPHINX-X~\cite{gao2024sphinxx} & Mixtral-8$\times$7B$^\blacklozenge$ & Mixture & Linear & \cmark & & Visual Dialogue, Multilingual, VQA, Captioning, Referring, REC \\
\bottomrule
\end{tabular}
}
\vspace{-.15cm}
\caption{\label{tab:models}
Summary of generalist MLLMs for vision-to-language tasks. For each model, we indicate the LLM used in its best configuration as shown in the original paper ($\lozenge$: LLM training from scratch; $\blacklozenge$: LLM fine-tuning; $\blacktriangle$: LLM fine-tuning with PEFT techniques; $\bigstar$: frozen LLM). The \amark~marker indicates variants to the reported vision-to-language adapter, while gray color indicates models not publicly available.
\vspace{-.35cm}
}
\end{table*}

\subsection{Visual Encoder}
\label{sec:visual_encoder}
In MLLMs, one of the key components is a visual encoder, which is specifically designed to provide the LLM with the visual extracted features. It is common to employ a frozen pre-trained visual encoder while training only a learnable interface that connects visual features with the underlying LLM. While this is usually done using low-resolution images with fixed aspect ratios, some attempts~\cite{xu2024llava,li2023monkey} involve adapting pre-trained visual backbones to handle images of different resolutions and aspect ratios. Further details on how to handle higher-resolution images are provided in the supplementary.

The most often employed visual encoders are based on pre-trained Vision Transformer (ViT) models with a CLIP-based objective to exploit the inherent alignment of CLIP embeddings. Popular choices are the ViT-L model from CLIP~\cite{radford2021learning}, the ViT-H backbone from OpenCLIP~\cite{wortsman2022robust}, and the ViT-g version from EVA-CLIP~\cite{fang2023eva}.

As shown in~\cite{li2023blip}, a stronger image encoder leads to better performance. Building on this insight,~\citet{lin2023sphinx} and~\citet{gao2024sphinxx} propose an ensemble of frozen visual backbones to capture robust visual representations and different levels of information granularity. Concurrently, PaLI models~\cite{chen2023pali,chen2023palix}, noticing an imbalance between language and visual parameters, propose scaling the visual backbone respectively to a 4- and 22-billion parameter ViT. 

The utilization of such large and powerful models is made feasible by the common practice of maintaining the visual encoder frozen during training, as observed in~\cite{li2023blip, huang2023language,gao2023llama,chen2023shikra}. However, employing a frozen visual encoder has some limitations, primarily due to the constrained number of parameters, resulting in an inadequate alignment between the visual and language modalities. Specifically, the dense features, extracted from the visual model, may fragment the fine-grained image information and bring large computation due to the lengthy sequence when fed into the language model. To mitigate this issue, other approaches~\cite{ye2023mplug,ye2023mplug2} employ a two-stage training paradigm. In the first stage, they incorporate a trainable visual backbone while maintaining the pre-trained LLM frozen. According to their findings, enabling the vision encoder to be trainable enhances performance on tasks such as visual question answering or visual description. However, it may lead to performance degradation in other tasks, indicating a degree of forgetting and damage to the general visual representation.

\subsection{Vision-to-Language Adapters}
The simultaneous presence of inputs from different modalities emphasizes the need to incorporate a module capable of delineating latent correspondences within these unimodal domains. These modules, termed as ``adapters'', are intended to facilitate interoperability between the visual and textual domains. A spectrum of different adapters are used in common MLLMs, ranging from elementary architectures such as linear layers or MLP to advanced methodologies such as Transformer-based solutions, exemplified by the Q-Former model, and conditioned cross-attention layers added to the LLM.

\tit{Linear and MLP Projections}
The most straightforward approach for projecting visual inputs into textual embeddings involves learning a linear mapping, which translates visual features to the same dimensionality as the textual counterpart. Some approaches like LLaMA-Adapter~\cite{gao2023llama} and FROMAGe~\cite{koh2023grounding} only employ a single linear layer to perform the multimodal connection, while LLaVA-1.5~\cite{liu2023improved} adopts a two-layer MLP, showing improved multimodal capabilities. Despite its widespread adoption in early MLLMs, the use of linear projections has proven highly effective even in recent methods with a more advanced understanding of the visual input~\cite{chen2023shikra,lin2023vila,wang2023cogvlm,you2023ferret,zhao2023svit}. It is, therefore, a simple yet effective technique for aligning visual features with textual counterparts. A different approach~\cite{cha2023honeybee} proposes to replace linear layers with convolutional ones, demonstrating moderate improvements.

\tit{Q-Former}
It is a Transformer-based model proposed in BLIP-2~\cite{li2023blip} and then used in several other approaches~\cite{chen2023lion,dai2023instructblip,hu2024bliva}. It is characterized by its adaptable architecture, which consists of two Transformer blocks that share mutual self-attention layers, facilitating the alignment process between visual and textual representations. It involves a set of learnable queries that interact within the self-attention layers and interface with visual features via a cross-attention mechanism. Textual and visual elements communicate via shared self-attention within the modules.

Drawing inspiration from the Q-Former, various modified versions have been introduced. In this regard, mPLUG-Owl models~\cite{ye2023mplug,ye2023mplug2} simplify the Q-Former architecture and propose a visual abstractor component that operates by condensing visual information into distinct learnable tokens to derive more semantically enriched visual representations. On the same line, Qwen-VL~\cite{bai2023qwen} compresses visual features using a single-layer cross-attention module with learnable queries also incorporating 2D positional encodings.

\tit{Additional Cross-Attention Layers}
This approach has been proposed in Flamingo~\cite{alayrac2022flamingo} with the integration of dense cross-attention blocks among the existing pre-trained layers of the LLM. The newly added layers are often combined with a zero-initialized tanh-gating mechanism to ensure that, upon initialization, the conditioned model acts as its original version. The use of additional cross-attention layers imposes the need to train them from scratch, increasing the number of trainable parameters compared to other alternatives. To reduce computational complexity, this strategy is usually paired with a Perceiver-based component~\cite{jaegle2021perceiver} that reduces the number of visual tokens before they are fed to the LLM. Since its introduction, several models~\cite{awadalla2023openflamingo,chen2023visual,laurenccon2023obelisc,li2023otter} employ this technique to connect the visual modality with the underlying LLM, demonstrating enhanced training stability and improved performance.

\subsection{Multimodal Training}
\label{sec:Multimodal_Training}
Starting from a pre-trained LLM, the training of an MLLM undergoes a single-stage or a two-stage process. In both cases, a standard cross-entropy loss is utilized for predicting the next token, serving as an auto-regressive objective.

\tit{Single-Stage Training}
This possibility is explored by LLaMA-Adapter~\cite{gao2023llama} which introduces additional trainable parameters to encapsulate the visual knowledge and manage text-only instruction learning at the same time. To achieve this, the model undergoes joint training using image-text pairs and instructions, operating on separate parameters. Concurrently, the model proposed in~\cite{koh2023grounding} adapts the final loss function by incorporating two contrastive losses for image-text retrieval. During the training, only three linear layers are updated.
On a different line, Kosmos-1~\cite{huang2023language} considers a frozen visual backbone and trains the language model of 1.3B parameters from scratch.

Flamingo~\cite{alayrac2022flamingo} and its open source variants~\cite{awadalla2023openflamingo,laurenccon2023obelisc}, instead, train the cross-attention layers and the Perceiver-based component to connect the visual features with the frozen LLM blocks.
Additionally, Otter~\cite{li2023otter} extends Flamingo's training to increment its in-context capabilities. Given the amount of training data currently available, approaches like SPHINX-X~\cite{gao2024sphinxx} opt to perform a single all-in-one training stage in which to update all model components, possibly also using text-only data to preserve the conversational capabilities of the LLM.

\tit{Two-Stage Training}
In the first of the two training stages, the objective is to align the image features with the text embedding space.
After this stage, the outputs tend to be fragmented and not coherent. Therefore, a second step is done to improve multimodal conversational capabilities. LLaVA~\cite{liu2023visual,liu2023improved} is among the first to introduce a visual instruction-following training scheme, which is performed as a second training stage updating the parameters of both the multimodal adapter and LLM. During the first stage, instead, only the multimodal adapter is trainable. 
Differently, MiniGPT-4~\cite{zhu2023minigpt} is notable for training solely the linear layer responsible for multimodal alignment across both stages. In the second stage, it uses filtered data, collected and refined through the model itself after the first stage.

Another approach, as demonstrated in InstructBLIP~\cite{dai2023instructblip}, involves the freezing of the visual encoder and LLM. In both training stages, only the Q-Former and the connection module are trainable. In contrast to previous approaches where the visual backbone remains frozen, mPLUG-Owl~\cite{ye2023mplug,ye2023mplug2} updates it in the initial stage, facilitating the capture of both low- and high-level visual information. Also, in the second stage text-only and multimodal data are used jointly to increase alignment. Differently, Shikra~\cite{chen2023shikra} updates all weights in both stages, with the only exception of the visual backbone which is kept frozen.

\tit{Training Data}
During the first (or single) training stage, the datasets predominantly consist of large-scale, publicly available, and uncurated data. For instance, the Conceptual Captions 3M (CC3M) dataset~\cite{sharma2018conceptual} is composed of 3M images paired with textual captions specifically designed for image captioning systems. Unlike the widely-used and curated MS-COCO~\cite{lin2014microsoft} dataset, which serves similar purposes, images and captions in CC3M are gathered from the web, showcasing a broader spectrum of styles and content. Similarly, the LAION family~\cite{schuhmann2021laion,schuhmann2022laion} represents an extended collection of non-curated image-text pairs sourced from web pages, providing a rich resource for pre-training multimodal language models. Additionally, the COYO-700M~\cite{kakaobrain2022coyo} dataset stands out as a significant resource, containing 747M image-text pairs. Notably, each alt-text in COYO-700M is linked to an image within HTML documents. Furthermore, DataComp~\cite{gadre2023datacomp} presents an extensive pool of 12.8B filtered image-text pairs sourced from common crawl.

It is important to highlight the distinction between datasets used in the initial phase of training, which typically comprise large-scale, uncurated data, and those selected for refinement in subsequent stages. While the former emphasizes diversity and scale, the latter focuses on specificity and task relevance, facilitating a more tailored approach to model optimization. Especially in single-training stage approaches, certain methods~\cite{alayrac2022flamingo,laurenccon2023obelisc} also leverage interleaved datasets, which contain images interleaved with text coming from the web, aiming to augment the dataset size for large models~\cite{hoffmann2022training}. Images within these datasets can be positioned at the beginning or in the middle of a sentence, allowing models to support arbitrarily interleaved sequences of images and text as input, thereby enhancing flexibility in input formats by blending textual and visual elements. Among these datasets, the most used are WebLI~\cite{chen2023pali}, composed of 10B images and image-text pairs, and MMC4~\cite{zhu2023multimodal}, an extension of the text-only C4~\cite{raffel2020exploring} dataset composed of 365M documents and 156B tokens relatives to different concepts, and OBELICS~\cite{laurenccon2023obelisc}, an open and curated collection of interleaved image-text web documents, containing 141M documents, 115B text tokens, and 353M images.

In the context of visual instruction tuning, which constitutes the second training stage for MLLMs, the available amount of data is limited. This limitation is mainly due to the creation process which is time-consuming and less well-defined. In this phase, different datasets are used to improve performances on a series of downstream tasks. Among them, LLaVA-Instruct~\cite{liu2023visual} is a collection of GPT-4 generated multimodal instruction-following data. It comprises 158k unique language-image descriptions, spanning various types of tasks including 58k conversations, 23k detailed descriptions, and 77k complex reasoning. Similarly, LRV-Instruction~\cite{liu2023mitigating} initially consisted of 400k visual instructions generated by GPT-4, and more recently, it has been updated with an additional set of 300k visual instructions. To enhance robustness in instruction tuning, LRV-Instruct also includes negative instructions organized across three semantic levels, showing that instruct-tuned MLLMs on this dataset suffer less from hallucination compared to the original versions. Moreover, LLaVAR~\cite{zhang2023llavar} considers publicly available OCR tools to collect results on 422k text-rich images from the LAION dataset. The pipeline first collects 422k noisy text-rich images and then extracts the text through OCR models. With the help of GPT-4, the results and captions are used to create 16k conversations, also including specific questions to create complex instructions which can be helpful to boost performance on new tasks. 

\begin{table*}
\centering
\setlength{\tabcolsep}{.28em}
\resizebox{\linewidth}{!}{
\begin{tabular}{l cc cc cl}
\toprule
\textbf{Model} & \textbf{LLM} & \textbf{Visual Encoder} & & \textbf{Supporting Model} & & \textbf{Main Tasks \& Capabilities} \\
\midrule
ContextDET~\cite{zang2023contextual} & OPT-6.7B$^\bigstar$ & Swin-B & & - & & Visual Dialogue, VQA, Captioning, Detection, REC, RES \\
DetGPT~\cite{pi2023detgpt} & Vicuna-13B$^\bigstar$ & EVA ViT-g & & G-DINO$^\bigstar$ & & Visual Dialogue, Detection \\
VisionLLM~\cite{wang2023visionllm} & Alpaca-7B$^\blacktriangle$ & Intern-H & & Deformable-DETR$^\blacktriangle$ & & VQA, Captioning, Detection, Segmentation, REC \\
BuboGPT~\cite{zhao2023bubogpt} & Vicuna-7B$^\bigstar$ & EVA ViT-g & & RAM, G-DINO, SAM$^\bigstar$ & & Visual Dialogue, Audio Understanding, Captioning, GroundCap  \\
ChatSpot~\cite{zhao2023chatspot} & Vicuna-7B$^\blacklozenge$ & CLIP ViT-L & & - & & Visual Dialogue, VQA, Captioning, Referring \\
GPT4RoI~\cite{zhang2023gpt4roi} & LLaVA-7B$^\blacklozenge$ & OpenCLIP ViT-H & & - & & Visual Dialogue, VQA, Captioning, Referring \\
ASM~\cite{wang2023all} & Husky-7B$^\blacktriangle$ & EVA ViT-g & & -  & & VQA, Captioning, Referring \\
LISA~\cite{lai2023lisa} & LLaVA-13B$^\blacktriangle$ & CLIP ViT-L & & SAM$^\blacklozenge$ & & Visual Dialogue, Captioning, RES \\
PVIT~\cite{chen2023position} & LLaVA-7B$^\blacklozenge$ & CLIP ViT-L & & RegionCLIP$^\bigstar$ & & Visual Dialogue, VQA, Captioning, Referring \\
GLaMM~\cite{rasheed2023glamm} & Vicuna-7B$^\blacktriangle$ & OpenCLIP ViT-H & & SAM$^\blacklozenge$ & & Visual Dialogue, Captioning, Referring, REC, RES, GroundCap \\
Griffon~\cite{zhan2023griffon} & LLaVA-13B$^\blacklozenge$ & CLIP ViT-L & & - & & REC, Detection, Phrase Grounding \\
LLaFS~\cite{zhu2023llafs} & CodeLLaMA-7B$^\blacktriangle$ & CLIP RN50 & & - & & Few-Shot Segmentation \\
NExT-Chat~\cite{zhang2023next} & Vicuna-7B$^\blacklozenge$ & CLIP ViT-L & & SAM$^\blacklozenge$ & & Visual Dialogue, Captioning, Referring, REC, RES, GroundCap \\
GSVA~\cite{xia2023gsva} & LLaVA-13B$^\blacktriangle$ & CLIP ViT-L & & SAM$^\blacklozenge$ & & VQA, Segmentation, REC, RES \\
Lenna~\cite{wei2023lenna} & LLaVA-7B$^\blacktriangle$ & CLIP ViT-L & & G-DINO$^\blacklozenge$ & & VQA, Captioning, REC \\
LISA++~\cite{yang2023improved} & LLaVA-13B$^\blacktriangle$ & CLIP ViT-L & & SAM$^\blacklozenge$ & & Visual Dialogue, Captioning, RES \\
LLaVA-G~\cite{zhang2023llava} & Vicuna-13B$^\blacklozenge$ & CLIP ViT-L & & OpenSeeD, S-SAM$^\blacklozenge$ & & Visual Dialogue, REC, RES, Grounding \\
PixelLLM~\cite{xu2023pixel} & FlanT5-XL-3B$^\blacktriangle$ & EVA ViT-L & & SAM$^\bigstar$ & & Referring, REC, RES, GroundCap \\
PixelLM~\cite{ren2023pixellm} & LLaVA-7B$^\blacktriangle$ & CLIP ViT-L & & - & & Visual Dialogue, RES \\
VistaLLM~\cite{pramanick2023jack} & Vicuna-13B$^\blacklozenge$ & EVA & & - & & Visual Dialogue, VQA, Referring, REC, RES, GroundCap \\
ChatterBox~\cite{tian2024chatterbox} & LLaVA-13B$^\blacktriangle$ & CLIP ViT-L & & iTPN-B$^\bigstar$, DINO$^\blacklozenge$ & & Visual Dialogue, Referring, REC, GroundCap \\
GELLA~\cite{qi2024generalizable} & LLaVA-13B$^\blacktriangle$ & CLIP ViT-L & & Mask2Former$^\blacklozenge$ & & Segmentation, RES, GroundCap \\
\rowcolor{Gray}
PaLI-3~\cite{chen2023pali3} & UL2-3B$^\blacklozenge$ & SigLIP ViT-g & & VQ-VAE$^\blacklozenge$ & & VQA, Captioning, Retrieval, RES \\
\bottomrule
\end{tabular}
}
\vspace{-.15cm}
\caption{\label{tab:grounding}
Summary of MLLMs with components specifically designed for visual grounding and region-level understanding. For each model, we indicate the LLM used in its best configuration, in some cases initialized with the weights of a pre-trained MLLM, and any supporting models used to perform the task ($\blacklozenge$: fine-tuning; $\blacktriangle$: fine-tuning with PEFT techniques; $\bigstar$: frozen). Gray color indicates models not publicly available.
\vspace{-.38cm}
}
\end{table*}

\section{Tackling Visual Tasks with MLLMs}
\label{sec:tasks}
Standard MLLMs can tackle visual understanding tasks, such as VQA, captioning and multi-turn conversation. However, recently there has been an interest in addressing more fine-grained visual tasks, such as visual grounding and image generation.

\subsection{Visual Grounding}
The visual grounding capabilities of an MLLM correspond to the ability to carry a dialogue with the user that includes the positioning of the content, also referred to as a referential dialogue~\cite{chen2023shikra}. In particular,~\citet{you2023ferret} introduce \textit{referring} as the ability to understand the content of an input region and can be evaluated on tasks such as region captioning and referring expression generation. 
Conversely, \textit{grounding} is associated with localizing regions of a given textual description and corresponds to tasks such as referring expression comprehension (REC), referring expression segmentation (RES), phrase grounding, and grounded captioning. 
Two main components are required to equip MLLMs with these capabilities: a region-to-sequence method to process input regions and a sequence-to-region method to ground nouns and phrases. A summary of the MLLMs with visual grounding capabilities is reported in Table~\ref{tab:grounding}. 

\tit{Region-as-Text} The most common way to output regions is to directly insert them into generated text as a series of coordinates, represented as numbers or as special tokens dedicated to location bins. Shikra~\cite{chen2023shikra}, Kosmos-2~\cite{peng2023kosmos}, MiniGPT-v2~\cite{chen2023minigpt}, Ferret~\cite{you2023ferret}, CogVLM~\cite{wang2023cogvlm}, SPHINX~\cite{lin2023sphinx}, Qwen-VL~\cite{bai2023qwen}, and Griffon~\cite{zhan2023griffon} convert bounding boxes into text by indicating two points. VisionLLM~\cite{wang2023visionllm}, VistaLLM~\cite{pramanick2023jack}, LLaFS~\cite{zhu2023llafs}, and ChatSpot~\cite{zhao2023chatspot} allow the MLLM to handle polygons by representing them as a series of points.

\tit{Embedding-as-Region} Another solution is to read input regions through region encoders and provide the output regions as embeddings extracted from the last layer of the MLLM to a decoder. For input regions, GLaMM~\cite{rasheed2023glamm}, GPT4RoI~\cite{zhang2023gpt4roi}, ASM~\cite{wang2023all} and ChatterBox~\cite{tian2024chatterbox} leverage features of the image encoder to perform ROI align on the bounding box, whereas PVIT~\cite{chen2023position} exploits RegionCLIP~\cite{zhong2022regionclip}. PixelLLM~\cite{xu2023pixel} and LLaVA-G~\cite{zhang2023llava} use the prompt encoder of SAM~\cite{kirillov2023segment} and Semantic-SAM~\cite{li2023semantic} respectively. For output regions, LISA~\cite{lai2023lisa}, GLaMM, GSVA~\cite{xia2023gsva}, NeXt-Chat~\cite{zhang2023next}, and LISA++~\cite{yang2023improved} send the embedding corresponding to special tokens to the mask decoder of SAM, LLaVA-G to OpenSeeD~\cite{zhang2023simple}, Lenna~\cite{wei2023lenna} to Grounding-DINO~\cite{liu2023grounding}, and PixelLM~\cite{ren2023pixellm} to a custom lightweight pixel decoder.

Differently, ContextDET~\cite{zang2023contextual} introduces a decoder that receives the latent embedding of the noun with learnable queries, performs a cross-attention with image features, and then uses a segmentation head. ChatterBox~\cite{tian2024chatterbox} combines features from the iTPN-B encoder~\cite{tian2023integrally} and the MLLM and provides them to the DINO detector~\cite{zhang2022dino}. GELLA~\cite{qi2024generalizable} presents a fusion module in Mask2Former~\cite{cheng2022masked} to propose masks based on multi-modal image features and an association module to assign latent embeddings to them. PaLI-3~\cite{chen2023pali3} converts embeddings into segmentation masks through a VQ-VAE~\cite{van2017neural} decoder.

\tit{Text-to-Grounding} Other approaches are based on open-vocabulary models that accept textual categories as input. DetGPT~\cite{pi2023detgpt} generates a list of categories for Grounding-DINO. BuboGPT~\cite{zhao2023bubogpt} leverages a combination of RAM, Grounding-DINO, and SAM and matches tags with nouns in the output sequence.

\begin{table*}
\centering
\setlength{\tabcolsep}{.28em}
\resizebox{\linewidth}{!}{
\begin{tabular}{l cc cc cl}
\toprule
\textbf{Model} & \textbf{LLM} & \textbf{Visual Encoder} & & \textbf{Supporting Model} & & \textbf{Main Tasks \& Capabilities} \\
\midrule
GILL~\cite{koh2023generating} & OPT-6.7B$^\bigstar$ & CLIP ViT-L & & SD v1.5$^\bigstar$ & & Visual Dialogue, Retrieval, Image Generation \\
Emu~\cite{sun2023generative} & LLaMA-13B$^\blacklozenge$ & EVA ViT-g & & SD v1.5$^\blacklozenge$ & & Visual Dialogue, VQA, Captioning, Image Generation \\
SEED~\cite{ge2023planting} & OPT-2.7B$^\blacktriangle$ & EVA ViT-g & & SD v1.4$^\bigstar$ & & VQA, Captioning, Image Generation \\
DreamLLM~\cite{dong2023dreamllm} & Vicuna-7B$^\blacklozenge$ & CLIP ViT-L & & SD v2.1$^\bigstar$ & & Visual Dialogue, VQA, Captioning, Image Generation, Interleaved Generation \\
LaVIT~\cite{jin2023unified} & LLaMA-7B$^\blacklozenge$ & EVA ViT-g & & SD v1.5$^\blacklozenge$ & & VQA, Captioning, Image Generation \\
MGIE~\cite{fu2023guiding} & LLaVA-7B$^\bigstar$ & CLIP ViT-L & & SD v1.5$^\blacklozenge$ & & Image Editing \\
TextBind~\cite{li2023textbind} & LLaMA-2-7B$^\blacklozenge$ & EVA ViT-g & & SD XL$^\bigstar$ & & Visual Dialogue, VQA, Captioning, Image Generation \\
Kosmos-G~\cite{pan2023kosmos} & Magneto-1.3B$^\lozenge$ & CLIP ViT-L & & SD v1.5$^\bigstar$ & & Image Generation, Compositional Image Generation  \\
MiniGPT-5~\cite{zheng2023minigpt} & Vicuna-7B$^\blacktriangle$ & EVA ViT-g & & SD v2.1$^\bigstar$ & & Visual Dialogue, Image Generation, Interleaved Generation \\
SEED-LLaMA~\cite{ge2023making} & LLaMA-2-13B$^\blacklozenge$ & EVA ViT-g & & SD unCLIP$^\bigstar$ & & Visual Dialogue, VQA, Captioning, Image Generation, Interleaved Generation\\
CoDi-2~\cite{tang2023codi} & LLaMA-2-7B$^\blacktriangle$ & ImageBind & & SD unCLIP$^\bigstar$ & & Visual Dialogue, Audio Understanding, Image Generation, Image Editing\\
Emu2~\cite{sun2023generative2} & LLaMA-33B$^\blacklozenge$ & EVA ViT-E & & SD XL$^\blacklozenge$ & & Visual Dialogue, VQA, Captioning, Image Generation, Image Editing \\
LLMGA~\cite{xia2023llmga} & LLaVA-13B$^\blacklozenge$ & CLIP ViT-L & & SD XL$^\blacklozenge$  & & Visual Dialogue, VQA, Image Generation, Image Editing \\
SmartEdit~\cite{huang2023smartedit} & LLaVA-13B$^\blacktriangle$& CLIP ViT-L & & SD$^\blacklozenge$ & & Image Editing \\
VL-GPT~\cite{zhu2023vl} & LLaMA-7B$^\blacktriangle$& CLIP ViT-L & & SD v1.5$^\bigstar$ & & Visual Dialogue, VQA, Captioning, Image Generation, Image Editing \\
MM-Interleaved~\cite{tian2024mm} & Vicuna-13B$^\blacklozenge$ & CLIP ViT-L & & SD v2.1$^\blacklozenge$ & & VQA, Captioning, REC, Image Generation, Interleaved Generation  \\
\rowcolor{Gray}
JAM~\cite{aiello2023jointly} & LLaMA\textsuperscript{\amark}-7B$^\blacklozenge$ & - & & CM3Leon$^\blacklozenge$ & & Image Generation, Interleaved Generation \\
\bottomrule
\end{tabular}
}
\vspace{-.15cm}
\caption{\label{tab:generation}
Summary of MLLMs with components specifically designed for image generation and editing. For each model, we indicate the LLM (\amark: LLM variants) used in its best configuration, in some cases initialized with the weights of a pre-trained MLLM, and any supporting models used to perform the task ($\lozenge$: training from scratch; $\blacklozenge$: fine-tuning; $\blacktriangle$: fine-tuning with PEFT techniques; $\bigstar$: frozen). Gray color indicates models not publicly available.
\vspace{-.38cm}
}
\end{table*}

\subsection{Image Generation and Editing}
While initial MLLMs excelled in extracting information from visual data, recent research included the generation of visual outputs. This advancement is realized through integrating MLLMs with image generation mechanisms, predominantly embodied by the Stable Diffusion (SD)~\cite{rombach2022high} models. These models feature a denoising U-Net~\cite{ronneberger2015u} architecture conditioned on textual or visual embeddings, through cross-attention layers. A complete list of the analyzed models is presented in Table~\ref{tab:generation}.

\tit{Connecting MLLMs with Diffusion Models} GILL~\cite{koh2023generating} is the pioneer in mapping the output embedding space of a frozen LLM to that of a frozen diffusion model. Specifically, inspired by Q-Former, a mapper component is trained by minimizing the $\ell_2$ distance between the image output representation of the language model and the expected conditioning embedding of SD. 

While GILL refrains from fine-tuning both the LLM and the diffusion U-Net, alternative methodologies fine-tune the language model to expand its multimodal generation capabilities. In this vein, Kosmos-G~\cite{pan2023kosmos} is developed through a training regime that integrates the output of the LLM with an encoder-decoder structure, leveraging a reconstruction loss and the minimization of the distance within a CLIP-text embedding. 
Similarly, MiniGPT-5~\cite{zheng2023minigpt} includes the reconstruction loss of diffusion models in addition to the alignment loss of GILL. Moreover, it divides the overall training process into two distinct phases: the initial phase concentrates on text-to-image generation, while the subsequent is focused on interleaved vision-and-language generation.
Distinctly, researchers have studied the alignment of discrete~\cite{jin2023unified,ge2023planting,ge2023making} and continuous visual tokens~\cite{zhu2023vl} extracted from input images with the SD conditioning embedding. This is usually achieved by fine-tuning the textual model~\cite{zhu2023vl,ge2023planting,ge2023making}. Conversely,~\citet{jin2023unified} fine-tune both the LLM and the SD U-Net.

A different approach has been studied by~\citet{li2023textbind} which proposes to fine-tune the LLM by adding two special tokens (\ie, \texttt{<start>} and \texttt{<end>}), and directly
encode the generated text between these two tokens using the text encoder in the SD model. Similarly, in~\cite{xia2023llmga} the LLM is trained to output detailed language-based generation prompts which are employed for generation or editing tasks. The U-Net is fine-tuned with longer and more detailed textual captions. Furthermore, in DreamLLM~\cite{dong2023dreamllm} an alignment loss is eschewed in favor of a score distillation loss while keeping the U-Net frozen. Additional research endeavors have been conducted to introduce MLLMs in the field of image editing~\cite{fu2023guiding,huang2023smartedit,tang2023codi}.  

\tit{End-to-End Pipelines}
A different direction is the development of end-to-end training strategies. Specifically, in~\cite{sun2023generative,sun2023generative2} the SD U-Net is directly fine-tuned with the continuous visual embeddings generated by the LLM. \citet{tian2024mm} employ a feature synchronizer, that intervenes in intermediate layers of the LLM and diffusion decoder to cross-attend multi-scale high-resolution image features.
Furthermore, end-to-end training approaches have been employed for non-diffusion-based generators, such as VQ-GAN~\cite{esser2021taming}, as demonstrated in the study by~\citet{lu2023unified}. Differently,~\citet{aiello2023jointly} propose a methodology to mix an LLM architecture with an autoregressive generator, CM3Leon~\cite{yu2023scaling}, via bi-directional cross-attention across the architectures of both models.

\subsection{Other Modalities and Applications}
\label{sec:others}

\tinytit{Video Understanding} Although much of the research focuses on images, some works propose MLLMs specifically designed to handle video sequences. These models process video frames independently, using CLIP-based backbones to extract frame-level features which are then combined using pooling mechanisms~\cite{li2023llama,maaz2023video} or Q-Former based solutions~\cite{li2023videochat,ren2023timechat}. The connection between visual features and the language model mainly follows the same trend as image-based MLLMs, with linear projections being the most common choice. However, there are also some attempts to develop video-specific adapters~\cite{liu2023one,ma2023vista} that can capture fine-grained temporal information. In addition to encoding video frames, some works~\cite{munasinghe2023pg,zhang2023video} also employ audio features to enrich the representation of input video sequences. Furthermore, effective strategies for visual instruction tuning are also designed in the video domain~\cite{song2024moviellm}, enabling more effective understanding of long video sequences.

\tit{Any-Modality Models}
Almost all models described so far treat a single modality as input to the LLM. However, a significant body of work focuses on designing effective solutions that can handle multiple modalities. This is usually achieved by aligning multimodal features through Transformer blocks such as Q-Former~\cite{chen2023x,panagopoulou2023x} and Perceiver~\cite{zhao2023chatbridge}, or by utilizing ImageBind to effectively extract features that are inherently multimodal~\cite{su2023pandagpt}. Images, videos, and audio are the most commonly treated modalities. Additionally, some works also effectively encode 3D data~\cite{yin2023lamm} and IMU sensor signals~\cite{moon2023anymal}. While all these solutions can manage multimodal inputs, approaches like NExT-GPT~\cite{wu2023next} and Unified-IO 2~\cite{lu2023unified} are also capable of generating outputs of different modalities.

\tit{Domain-Specific MLLMs} 
In addition to dealing with generic visual inputs, some research efforts are dedicated to developing MLLMs for specific domains and applications, either training the model starting from a pre-trained LLM or fine-tuning an existing MLLM with domain-specific data. Some examples are MLLMs designed for document analysis and text-intensive visual inputs~\cite{lv2023kosmos,ye2023mplugdoc}, those proposed for embodied AI and robotics~\cite{driess2023palm,mu2023embodiedgpt}, and those tailored for specific domains such as medicine~\cite{li2023llava} and autonomous driving~\cite{xu2023drivegpt4}. A complete list of domain-specific MLLMs is reported in the supplementary.

\section{Conclusion and Future Directions}
\vspace{-0.05cm}
\label{sec:challenges}
In this survey, we have provided a comprehensive overview of the recent evolution of MLLMs, first focusing on how to equip LLMs with multimodal capabilities and then exploring the main tasks addressed by these models. Based on the analysis presented, in the following, we outline important open challenges and promising future research directions to further empower MLLMs.

\tit{Multimodal Retrieval-Augmented Generation} While retrieval-augmented generation (RAG) is a consolidated technique in LLMs~\cite{lewis2020retrieval,asai2023self}, its application in MLLMs is still under-explored. We believe that the emergence of VQA datasets that require external retrieved knowledge~\cite{chen2023can,mensink2023encyclopedic} may enable the development of MLLMs with RAG capabilities~\cite{hu2023reveal,caffagni2024wiki}.

\tit{Correction of Hallucinations} 
Several studies~\cite{liu2023aligning,zhu2023minigpt} show that MLLMs tend to exhibit high hallucination rates, especially when generating longer captions. While some solutions are emerging to mitigate this problem~\cite{liu2023aligning, wang2023vigc,wu2023see,yin2023woodpecker,jing2023faithscore}, understanding and correcting the underlying causes of hallucinations remains an important open challenge that is worth addressing to allow the application of these models in more critical contexts (\eg, medicine) and guarantee their accuracy and trustworthiness.

\tit{Prevent Harmful and Biased Generation}
Ensuring the safety and fairness of large-scale models is of fundamental interest in the community. Recent works show that models trained on web-crawled data are prone to generate inappropriate and biased content. Although recent efforts are being made to reduce this phenomenon in text-to-image generative  models~\cite{schramowski2023safe,friedrich2023fair,poppi2024removing}, further exploration is needed to prevent the same behavior in MLLMs~\cite{pi2024mllm}.

\tit{Reduce Computational Load} As shown in the supplementary, MLLMs are highly computationally demanding. Effective strategies~\cite{chu2024mobilevlm} are needed to reduce computational requirements and enable more accessible development of MLLMs. Possible directions entail reducing training requirements both in terms of model scale and data quantity and optimizing the inference stage.

\clearpage
\section*{Limitations}
This survey provides a comprehensive review of visual-based MLLMs. Although we have made a significant effort to include all relevant works available to the date of submission, the review might have missed some minor works, and might not have a complete coverage of MLLMs treating modalities that are different from the visual one. Additionally, given the space constraints required by the submission venue, we have restricted our explanations of existing approaches so as to include only the most relevant novelty points. We encourage the reader to refer to the original papers for further technical details and implementation notes.

\section*{Acknowledgments}
This work has been partially supported by the projects: PNRR-M4C2 (PE00000013) ``FAIR - Future Artificial Intelligence Research'' funded by the European Commission, the PNRR project ``Italian Strengthening of Esfri RI Resilience'' (ITSERR) funded by the European Union - NextGenerationEU (CUP B53C22001770006), and the PRIN project ``CREATIVE: CRoss-modal understanding and gEnerATIon of Visual and tExtual content'' co-funded by the Italian Ministry of University and Research (CUP B87G22000460001).

\bibliography{bibliography}

\begin{thebibliography}{253}
\expandafter\ifx\csname natexlab\endcsname\relax\def\natexlab#1{#1}\fi

\bibitem[{Achiam et~al.(2023)Achiam, Adler, Agarwal, Ahmad, Akkaya, Aleman, Almeida, Altenschmidt, Altman, Anadkat et~al.}]{achiam2023gpt}
Josh Achiam, Steven Adler, Sandhini Agarwal, Lama Ahmad, Ilge Akkaya, Florencia~Leoni Aleman, Diogo Almeida, Janko Altenschmidt, Sam Altman, Shyamal Anadkat, et~al. 2023.
\newblock {GPT-4 Technical Report}.
\newblock \emph{arXiv preprint arXiv:2303.08774}.

\bibitem[{Agrawal et~al.(2019)Agrawal, Desai, Wang, Chen, Jain, Johnson, Batra, Parikh, Lee, and Anderson}]{agrawal2019nocaps}
Harsh Agrawal, Karan Desai, Yufei Wang, Xinlei Chen, Rishabh Jain, Mark Johnson, Dhruv Batra, Devi Parikh, Stefan Lee, and Peter Anderson. 2019.
\newblock nocaps: novel object captioning at scale.
\newblock In \emph{ICCV}.

\bibitem[{Aiello et~al.(2024)Aiello, Yu, Nie, Aghajanyan, and Oguz}]{aiello2023jointly}
Emanuele Aiello, Lili Yu, Yixin Nie, Armen Aghajanyan, and Barlas Oguz. 2024.
\newblock {Jointly Training Large Autoregressive Multimodal Models}.
\newblock In \emph{ICLR}.

\bibitem[{Alayrac et~al.(2022)Alayrac, Donahue, Luc, Miech, Barr, Hasson, Lenc, Mensch, Millican, Reynolds et~al.}]{alayrac2022flamingo}
Jean-Baptiste Alayrac, Jeff Donahue, Pauline Luc, Antoine Miech, Iain Barr, Yana Hasson, Karel Lenc, Arthur Mensch, Katherine Millican, Malcolm Reynolds, et~al. 2022.
\newblock {Flamingo: a Visual Language Model for Few-Shot Learning}.
\newblock In \emph{NeurIPS}.

\bibitem[{Anil et~al.(2023)Anil, Borgeaud, Wu, Alayrac, Yu, Soricut, Schalkwyk, Dai, Hauth et~al.}]{team2023gemini}
Rohan Anil, Sebastian Borgeaud, Yonghui Wu, Jean-Baptiste Alayrac, Jiahui Yu, Radu Soricut, Johan Schalkwyk, Andrew~M Dai, Anja Hauth, et~al. 2023.
\newblock {Gemini: A Family of Highly Capable Multimodal Models}.
\newblock \emph{arXiv preprint arXiv:2312.11805}.

\bibitem[{Antol et~al.(2015)Antol, Agrawal, Lu, Mitchell, Batra, Zitnick, and Parikh}]{antol2015vqa}
Stanislaw Antol, Aishwarya Agrawal, Jiasen Lu, Margaret Mitchell, Dhruv Batra, C~Lawrence Zitnick, and Devi Parikh. 2015.
\newblock {VQA: Visual Question Answering}.
\newblock In \emph{ICCV}.

\bibitem[{Asai et~al.(2023)Asai, Wu, Wang, Sil, and Hajishirzi}]{asai2023self}
Akari Asai, Zeqiu Wu, Yizhong Wang, Avirup Sil, and Hannaneh Hajishirzi. 2023.
\newblock {Self-RAG: Learning to Retrieve, Generate, and Critique through Self-Reflection}.
\newblock \emph{arXiv preprint arXiv:2310.11511}.

\bibitem[{Awadalla et~al.(2023)Awadalla, Gao, Gardner, Hessel, Hanafy, Zhu, Marathe, Bitton, Gadre, Sagawa et~al.}]{awadalla2023openflamingo}
Anas Awadalla, Irena Gao, Josh Gardner, Jack Hessel, Yusuf Hanafy, Wanrong Zhu, Kalyani Marathe, Yonatan Bitton, Samir Gadre, Shiori Sagawa, et~al. 2023.
\newblock {OpenFlamingo: An Open-Source Framework for Training Large Autoregressive Vision-Language Models}.
\newblock \emph{arXiv preprint arXiv:2308.01390}.

\bibitem[{Bai et~al.(2023{\natexlab{a}})Bai, Bai, Chu, Cui, Dang, Deng, Fan, Ge, Han, Huang et~al.}]{bai2023qwentc}
Jinze Bai, Shuai Bai, Yunfei Chu, Zeyu Cui, Kai Dang, Xiaodong Deng, Yang Fan, Wenbin Ge, Yu~Han, Fei Huang, et~al. 2023{\natexlab{a}}.
\newblock {Qwen technical report}.
\newblock \emph{arXiv preprint arXiv:2309.16609}.

\bibitem[{Bai et~al.(2023{\natexlab{b}})Bai, Bai, Yang, Wang, Tan, Wang, Lin, Zhou, and Zhou}]{bai2023qwen}
Jinze Bai, Shuai Bai, Shusheng Yang, Shijie Wang, Sinan Tan, Peng Wang, Junyang Lin, Chang Zhou, and Jingren Zhou. 2023{\natexlab{b}}.
\newblock {Qwen-VL: A Versatile Vision-Language Model for Understanding, Localization, Text Reading, and Beyond}.
\newblock \emph{arXiv preprint arXiv:2308.12966}.

\bibitem[{Bai et~al.(2023{\natexlab{c}})Bai, Yang, Bai, Wang, Zhang, Lin, Wang, Zhou, and Zhou}]{bai2023touchstone}
Shuai Bai, Shusheng Yang, Jinze Bai, Peng Wang, Xingxuan Zhang, Junyang Lin, Xinggang Wang, Chang Zhou, and Jingren Zhou. 2023{\natexlab{c}}.
\newblock {TouchStone: Evaluating Vision-Language Models by Language Models}.
\newblock \emph{arXiv preprint arXiv:2308.16890}.

\bibitem[{Banerjee and Lavie(2005)}]{banerjee2005meteor}
Satanjeev Banerjee and Alon Lavie. 2005.
\newblock {METEOR: An automatic metric for MT evaluation with improved correlation with human judgments}.
\newblock In \emph{ACL Workshops}.

\bibitem[{Brooks et~al.(2023)Brooks, Holynski, and Efros}]{brooks2023instructpix2pix}
Tim Brooks, Aleksander Holynski, and Alexei~A Efros. 2023.
\newblock {InstructPix2Pix: Learning to Follow Image Editing Instructions}.
\newblock In \emph{CVPR}.

\bibitem[{Brown et~al.(2020)Brown, Mann, Ryder, Subbiah, Kaplan, Dhariwal, Neelakantan, Shyam, Sastry, Askell et~al.}]{brown2020language}
Tom Brown, Benjamin Mann, Nick Ryder, Melanie Subbiah, Jared~D Kaplan, Prafulla Dhariwal, Arvind Neelakantan, Pranav Shyam, Girish Sastry, Amanda Askell, et~al. 2020.
\newblock {Language models are few-shot learners}.
\newblock In \emph{NeurIPS}.

\bibitem[{Byeon et~al.(2022)Byeon, Park, Kim, Lee, Baek, and Kim}]{kakaobrain2022coyo}
Minwoo Byeon, Beomhee Park, Haecheon Kim, Sungjun Lee, Woonhyuk Baek, and Saehoon Kim. 2022.
\newblock {COYO-700M: Image-Text Pair Dataset}.

\bibitem[{Caffagni et~al.(2024)Caffagni, Cocchi, Moratelli, Sarto, Cornia, Baraldi, and Cucchiara}]{caffagni2024wiki}
Davide Caffagni, Federico Cocchi, Nicholas Moratelli, Sara Sarto, Marcella Cornia, Lorenzo Baraldi, and Rita Cucchiara. 2024.
\newblock {Wiki-LLaVA: Hierarchical Retrieval-Augmented Generation for Multimodal LLMs}.
\newblock In \emph{CVPR Workshops}.

\bibitem[{Caron et~al.(2021)Caron, Touvron, Misra, J{\'e}gou, Mairal, Bojanowski, and Joulin}]{caron2021emerging}
Mathilde Caron, Hugo Touvron, Ishan Misra, Herv{\'e} J{\'e}gou, Julien Mairal, Piotr Bojanowski, and Armand Joulin. 2021.
\newblock {Emerging Properties in Self-Supervised Vision Transformers}.
\newblock In \emph{ICCV}.

\bibitem[{Cha et~al.(2023)Cha, Kang, Mun, and Roh}]{cha2023honeybee}
Junbum Cha, Wooyoung Kang, Jonghwan Mun, and Byungseok Roh. 2023.
\newblock {Honeybee: Locality-enhanced Projector for Multimodal LLM}.
\newblock \emph{arXiv preprint arXiv:2312.06742}.

\bibitem[{Chen et~al.(2023{\natexlab{a}})Chen, Qin, Luo, Mi, Li, Sun, and Liu}]{chen2023position}
Chi Chen, Ruoyu Qin, Fuwen Luo, Xiaoyue Mi, Peng Li, Maosong Sun, and Yang Liu. 2023{\natexlab{a}}.
\newblock {Position-Enhanced Visual Instruction Tuning for Multimodal Large Language Models}.
\newblock \emph{arXiv preprint arXiv:2308.13437}.

\bibitem[{Chen et~al.(2023{\natexlab{b}})Chen, Liu, Dai, and Wang}]{chen2023visual}
Delong Chen, Jianfeng Liu, Wenliang Dai, and Baoyuan Wang. 2023{\natexlab{b}}.
\newblock {Visual Instruction Tuning with Polite Flamingo}.
\newblock \emph{arXiv preprint arXiv:2307.01003}.

\bibitem[{Chen et~al.(2023{\natexlab{c}})Chen, Han, Zhao, Zhang, Shi, Xu, and Xu}]{chen2023x}
Feilong Chen, Minglun Han, Haozhi Zhao, Qingyang Zhang, Jing Shi, Shuang Xu, and Bo~Xu. 2023{\natexlab{c}}.
\newblock {X-LLM: Bootstrapping Advanced Large Language Models by Treating Multi-Modalities as Foreign Languages}.
\newblock \emph{arXiv preprint arXiv:2305.04160}.

\bibitem[{Chen et~al.(2023{\natexlab{d}})Chen, Shen, Shao, Deng, and Nie}]{chen2023lion}
Gongwei Chen, Leyang Shen, Rui Shao, Xiang Deng, and Liqiang Nie. 2023{\natexlab{d}}.
\newblock {LION: Empowering Multimodal Large Language Model with Dual-Level Visual Knowledge}.
\newblock \emph{arXiv preprint arXiv:2311.11860}.

\bibitem[{Chen et~al.(2023{\natexlab{e}})Chen, Zhu, Shen, Li, Liu, Zhang, Krishnamoorthi, Chandra, Xiong, and Elhoseiny}]{chen2023minigpt}
Jun Chen, Deyao Zhu, Xiaoqian Shen, Xiang Li, Zechun Liu, Pengchuan Zhang, Raghuraman Krishnamoorthi, Vikas Chandra, Yunyang Xiong, and Mohamed Elhoseiny. 2023{\natexlab{e}}.
\newblock {MiniGPT-v2: Large Language Model As a Unified Interface for Vision-Language Multi-task Learning}.
\newblock \emph{arXiv preprint arXiv:2310.09478}.

\bibitem[{Chen et~al.(2023{\natexlab{f}})Chen, Zhang, Zeng, Zhang, Zhu, and Zhao}]{chen2023shikra}
Keqin Chen, Zhao Zhang, Weili Zeng, Richong Zhang, Feng Zhu, and Rui Zhao. 2023{\natexlab{f}}.
\newblock {Shikra: Unleashing Multimodal LLM's Referential Dialogue Magic}.
\newblock \emph{arXiv preprint arXiv:2306.15195}.

\bibitem[{Chen et~al.(2023{\natexlab{g}})Chen, Chen, Zhang, Li, Yu, Fei, Zhu, Fan, and Chen}]{chen2023ll3da}
Sijin Chen, Xin Chen, Chi Zhang, Mingsheng Li, Gang Yu, Hao Fei, Hongyuan Zhu, Jiayuan Fan, and Tao Chen. 2023{\natexlab{g}}.
\newblock {LL3DA: Visual Interactive Instruction Tuning for Omni-3D Understanding, Reasoning, and Planning}.
\newblock \emph{arXiv preprint arXiv:2311.18651}.

\bibitem[{Chen et~al.(2023{\natexlab{h}})Chen, Djolonga, Padlewski, Mustafa, Changpinyo, Wu, Ruiz, Goodman, Wang, Tay et~al.}]{chen2023palix}
Xi~Chen, Josip Djolonga, Piotr Padlewski, Basil Mustafa, Soravit Changpinyo, Jialin Wu, Carlos~Riquelme Ruiz, Sebastian Goodman, Xiao Wang, Yi~Tay, et~al. 2023{\natexlab{h}}.
\newblock {PaLI-X: On Scaling up a Multilingual Vision and Language Model}.
\newblock \emph{arXiv preprint arXiv:2305.18565}.

\bibitem[{Chen et~al.(2023{\natexlab{i}})Chen, Wang, Beyer, Kolesnikov, Wu, Voigtlaender, Mustafa, Goodman, Alabdulmohsin, Padlewski et~al.}]{chen2023pali3}
Xi~Chen, Xiao Wang, Lucas Beyer, Alexander Kolesnikov, Jialin Wu, Paul Voigtlaender, Basil Mustafa, Sebastian Goodman, Ibrahim Alabdulmohsin, Piotr Padlewski, et~al. 2023{\natexlab{i}}.
\newblock {PaLI-3 Vision Language Models: Smaller, Faster, Stronger}.
\newblock \emph{arXiv preprint arXiv:2310.09199}.

\bibitem[{Chen et~al.(2023{\natexlab{j}})Chen, Wang, Changpinyo, Piergiovanni, Padlewski, Salz, Goodman, Grycner, Mustafa, Beyer et~al.}]{chen2023pali}
Xi~Chen, Xiao Wang, Soravit Changpinyo, AJ~Piergiovanni, Piotr Padlewski, Daniel Salz, Sebastian Goodman, Adam Grycner, Basil Mustafa, Lucas Beyer, et~al. 2023{\natexlab{j}}.
\newblock {PaLI: A Jointly-Scaled Multilingual Language-Image Model}.
\newblock In \emph{ICLR}.

\bibitem[{Chen et~al.(2023{\natexlab{k}})Chen, Hu, Luan, Sun, Changpinyo, Ritter, and Chang}]{chen2023can}
Yang Chen, Hexiang Hu, Yi~Luan, Haitian Sun, Soravit Changpinyo, Alan Ritter, and Ming-Wei Chang. 2023{\natexlab{k}}.
\newblock {Can Pre-trained Vision and Language Models Answer Visual Information-Seeking Questions?}
\newblock In \emph{EMNLP}.

\bibitem[{Chen et~al.(2023{\natexlab{l}})Chen, Sikka, Cogswell, Ji, and Divakaran}]{chen2023dress}
Yangyi Chen, Karan Sikka, Michael Cogswell, Heng Ji, and Ajay Divakaran. 2023{\natexlab{l}}.
\newblock {DRESS: Instructing Large Vision-Language Models to Align and Interact with Humans via Natural Language Feedback}.
\newblock \emph{arXiv preprint arXiv:2311.10081}.

\bibitem[{Cheng et~al.(2022)Cheng, Misra, Schwing, Kirillov, and Girdhar}]{cheng2022masked}
Bowen Cheng, Ishan Misra, Alexander~G Schwing, Alexander Kirillov, and Rohit Girdhar. 2022.
\newblock {Masked-Attention Mask Transformer for Universal Image Segmentation}.
\newblock In \emph{CVPR}.

\bibitem[{Chiang et~al.(2023)Chiang, Li, Lin, Sheng, Wu, Zhang, Zheng, Zhuang, Zhuang, Gonzalez, Stoica, and Xing}]{vicuna2023}
Wei-Lin Chiang, Zhuohan Li, Zi~Lin, Ying Sheng, Zhanghao Wu, Hao Zhang, Lianmin Zheng, Siyuan Zhuang, Yonghao Zhuang, Joseph~E. Gonzalez, Ion Stoica, and Eric~P. Xing. 2023.
\newblock \href {https://lmsys.org/blog/2023-03-30-vicuna/} {{Vicuna: An Open-Source Chatbot Impressing GPT-4 with 90\%* ChatGPT Quality}}.

\bibitem[{Chowdhery et~al.(2023)Chowdhery, Narang, Devlin, Bosma, Mishra, Roberts, Barham, Chung, Sutton, Gehrmann et~al.}]{chowdhery2023palm}
Aakanksha Chowdhery, Sharan Narang, Jacob Devlin, Maarten Bosma, Gaurav Mishra, Adam Roberts, Paul Barham, Hyung~Won Chung, Charles Sutton, Sebastian Gehrmann, et~al. 2023.
\newblock {Palm: Scaling language modeling with pathways}.
\newblock \emph{JMLR}, 24(240):1--113.

\bibitem[{Chu et~al.(2024)Chu, Qiao, Zhang, Xu, Wei, Yang, Sun, Hu, Lin, Zhang et~al.}]{chu2024mobilevlm}
Xiangxiang Chu, Limeng Qiao, Xinyu Zhang, Shuang Xu, Fei Wei, Yang Yang, Xiaofei Sun, Yiming Hu, Xinyang Lin, Bo~Zhang, et~al. 2024.
\newblock {MobileVLM V2: Faster and Stronger Baseline for Vision Language Model}.
\newblock \emph{arXiv preprint arXiv:2402.03766}.

\bibitem[{Chung et~al.(2022)Chung, Hou, Longpre, Zoph, Tay, Fedus, Li, Wang, Dehghani, Brahma et~al.}]{chung2022scaling}
Hyung~Won Chung, Le~Hou, Shayne Longpre, Barret Zoph, Yi~Tay, William Fedus, Yunxuan Li, Xuezhi Wang, Mostafa Dehghani, Siddhartha Brahma, et~al. 2022.
\newblock {Scaling Instruction-Finetuned Language Models}.
\newblock \emph{arXiv preprint arXiv:2210.11416}.

\bibitem[{Dai et~al.(2017)Dai, Chang, Savva, Halber, Funkhouser, and Nie{\ss}ner}]{dai2017scannet}
Angela Dai, Angel~X Chang, Manolis Savva, Maciej Halber, Thomas Funkhouser, and Matthias Nie{\ss}ner. 2017.
\newblock {Scannet: Richly-Annotated 3D Reconstructions of Indoor Scenes}.
\newblock In \emph{CVPR}.

\bibitem[{Dai et~al.(2023)Dai, Li, Li, Tiong, Zhao, Wang, Li, Fung, and Hoi}]{dai2023instructblip}
Wenliang Dai, Junnan Li, Dongxu Li, Anthony Meng~Huat Tiong, Junqi Zhao, Weisheng Wang, Boyang Li, Pascale Fung, and Steven Hoi. 2023.
\newblock {InstructBLIP: Towards General-purpose Vision-Language Models with Instruction Tuning}.
\newblock \emph{arXiv preprint arXiv:2305.06500}.

\bibitem[{Dettmers et~al.(2024)Dettmers, Pagnoni, Holtzman, and Zettlemoyer}]{dettmers2023qlora}
Tim Dettmers, Artidoro Pagnoni, Ari Holtzman, and Luke Zettlemoyer. 2024.
\newblock Qlora: Efficient finetuning of quantized llms.
\newblock In \emph{NeurIPS}.

\bibitem[{Devlin et~al.(2018)Devlin, Chang, Lee, and Toutanova}]{devlin2018bert}
Jacob Devlin, Ming-Wei Chang, Kenton Lee, and Kristina Toutanova. 2018.
\newblock {BERT: Pre-training of deep bidirectional transformers for language understanding}.
\newblock In \emph{NAACL}.

\bibitem[{Dong et~al.(2023)Dong, Han, Peng, Qi, Ge, Yang, Zhao, Sun, Zhou, Wei et~al.}]{dong2023dreamllm}
Runpei Dong, Chunrui Han, Yuang Peng, Zekun Qi, Zheng Ge, Jinrong Yang, Liang Zhao, Jianjian Sun, Hongyu Zhou, Haoran Wei, et~al. 2023.
\newblock {DreamLLM: Synergistic Multimodal Comprehension and Creation}.
\newblock \emph{arXiv preprint arXiv:2309.11499}.

\bibitem[{Driess et~al.(2023)Driess, Xia, Sajjadi, Lynch, Chowdhery, Ichter, Wahid, Tompson, Vuong, Yu et~al.}]{driess2023palm}
Danny Driess, Fei Xia, Mehdi~SM Sajjadi, Corey Lynch, Aakanksha Chowdhery, Brian Ichter, Ayzaan Wahid, Jonathan Tompson, Quan Vuong, Tianhe Yu, et~al. 2023.
\newblock {PaLM-E: An Embodied Multimodal Language Model}.
\newblock \emph{arXiv preprint arXiv:2303.03378}.

\bibitem[{Esser et~al.(2021)Esser, Rombach, and Ommer}]{esser2021taming}
Patrick Esser, Robin Rombach, and Bjorn Ommer. 2021.
\newblock {Taming Transformers for High-Resolution Image Synthesis}.
\newblock In \emph{CVPR}.

\bibitem[{Fang et~al.(2023)Fang, Wang, Xie, Sun, Wu, Wang, Huang, Wang, and Cao}]{fang2023eva}
Yuxin Fang, Wen Wang, Binhui Xie, Quan Sun, Ledell Wu, Xinggang Wang, Tiejun Huang, Xinlong Wang, and Yue Cao. 2023.
\newblock {Eva: Exploring the limits of masked visual representation learning at scale}.
\newblock In \emph{CVPR}.

\bibitem[{Feng et~al.(2023)Feng, Liu, Liu, Zhou, Li, and Huang}]{feng2023docpedia}
Hao Feng, Qi~Liu, Hao Liu, Wengang Zhou, Houqiang Li, and Can Huang. 2023.
\newblock {DocPedia: Unleashing the Power of Large Multimodal Model in the Frequency Domain for Versatile Document Understanding}.
\newblock \emph{arXiv preprint arXiv:2311.11810}.

\bibitem[{Friedrich et~al.(2023)Friedrich, Schramowski, Brack, Struppek, Hintersdorf, Luccioni, and Kersting}]{friedrich2023fair}
Felix Friedrich, Patrick Schramowski, Manuel Brack, Lukas Struppek, Dominik Hintersdorf, Sasha Luccioni, and Kristian Kersting. 2023.
\newblock {Fair Diffusion: Instructing Text-to-Image Generation Models on Fairness}.
\newblock \emph{arXiv preprint arXiv:2302.10893}.

\bibitem[{Fu et~al.(2023)Fu, Chen, Shen, Qin, Zhang, Lin, Yang, Zheng, Li, Sun et~al.}]{fu2023mme}
Chaoyou Fu, Peixian Chen, Yunhang Shen, Yulei Qin, Mengdan Zhang, Xu~Lin, Jinrui Yang, Xiawu Zheng, Ke~Li, Xing Sun, et~al. 2023.
\newblock {MME: A Comprehensive Evaluation Benchmark for Multimodal Large Language Models}.
\newblock \emph{arXiv preprint arXiv:2306.13394}.

\bibitem[{Fu et~al.(2024)Fu, Hu, Du, Wang, Yang, and Gan}]{fu2023guiding}
Tsu-Jui Fu, Wenze Hu, Xianzhi Du, William~Yang Wang, Yinfei Yang, and Zhe Gan. 2024.
\newblock {Guiding Instruction-based Image Editing via Multimodal Large Language Models}.
\newblock In \emph{ICLR}.

\bibitem[{Gadre et~al.(2023)Gadre, Ilharco, Fang, Hayase, Smyrnis, Nguyen, Marten, Wortsman, Ghosh, Zhang et~al.}]{gadre2023datacomp}
Samir~Yitzhak Gadre, Gabriel Ilharco, Alex Fang, Jonathan Hayase, Georgios Smyrnis, Thao Nguyen, Ryan Marten, Mitchell Wortsman, Dhruba Ghosh, Jieyu Zhang, et~al. 2023.
\newblock Datacomp: In search of the next generation of multimodal datasets.
\newblock In \emph{NeurIPS}.

\bibitem[{Gao et~al.(2023)Gao, Han, Zhang, Lin, Geng, Zhou, Zhang, Lu, He, Yue et~al.}]{gao2023llama}
Peng Gao, Jiaming Han, Renrui Zhang, Ziyi Lin, Shijie Geng, Aojun Zhou, Wei Zhang, Pan Lu, Conghui He, Xiangyu Yue, et~al. 2023.
\newblock {LLaMA-Adapter V2: Parameter-Efficient Visual Instruction Model}.
\newblock \emph{arXiv preprint arXiv:2304.15010}.

\bibitem[{Gao et~al.(2024)Gao, Zhang, Liu, Qiu, Huang, Lin, Zhao, Geng, Lin, Jin, Zhang, Shao, Xu, He, He, Shao, Lu, Li, and Qiao}]{gao2024sphinxx}
Peng Gao, Renrui Zhang, Chris Liu, Longtian Qiu, Siyuan Huang, Weifeng Lin, Shitian Zhao, Shijie Geng, Ziyi Lin, Peng Jin, Kaipeng Zhang, Wenqi Shao, Chao Xu, Conghui He, Junjun He, Hao Shao, Pan Lu, Hongsheng Li, and Yu~Qiao. 2024.
\newblock {SPHINX-X: Scaling Data and Parameters for a Family of Multi-modal Large Language Models}.
\newblock \emph{arXiv preprint arXiv:2402.05935}.

\bibitem[{Ge et~al.(2023{\natexlab{a}})Ge, Ge, Zeng, Wang, and Shan}]{ge2023planting}
Yuying Ge, Yixiao Ge, Ziyun Zeng, Xintao Wang, and Ying Shan. 2023{\natexlab{a}}.
\newblock {Planting a SEED of Vision in Large Language Model}.
\newblock \emph{arXiv preprint arXiv:2307.08041}.

\bibitem[{Ge et~al.(2023{\natexlab{b}})Ge, Zhao, Zeng, Ge, Li, Wang, and Shan}]{ge2023making}
Yuying Ge, Sijie Zhao, Ziyun Zeng, Yixiao Ge, Chen Li, Xintao Wang, and Ying Shan. 2023{\natexlab{b}}.
\newblock {Making LLaMA SEE and Draw with SEED Tokenizer}.
\newblock \emph{arXiv preprint arXiv:2310.01218}.

\bibitem[{Girdhar et~al.(2023)Girdhar, El-Nouby, Liu, Singh, Alwala, Joulin, and Misra}]{girdhar2023imagebind}
Rohit Girdhar, Alaaeldin El-Nouby, Zhuang Liu, Mannat Singh, Kalyan~Vasudev Alwala, Armand Joulin, and Ishan Misra. 2023.
\newblock {ImageBind: One Embedding Space To Bind Them All}.
\newblock In \emph{CVPR}.

\bibitem[{Gong et~al.(2023)Gong, Lyu, Zhang, Wang, Zheng, Zhao, Liu, Zhang, Luo, and Chen}]{gong2023multimodal}
Tao Gong, Chengqi Lyu, Shilong Zhang, Yudong Wang, Miao Zheng, Qian Zhao, Kuikun Liu, Wenwei Zhang, Ping Luo, and Kai Chen. 2023.
\newblock {MultiModal-GPT: A Vision and Language Model for Dialogue with Humans}.
\newblock \emph{arXiv preprint arXiv:2305.04790}.

\bibitem[{Goyal et~al.(2017)Goyal, Khot, Summers-Stay, Batra, and Parikh}]{goyal2017making}
Yash Goyal, Tejas Khot, Douglas Summers-Stay, Dhruv Batra, and Devi Parikh. 2017.
\newblock {Making the v in VQA Matter: Elevating the Role of Image Understanding in Visual Question Answering}.
\newblock In \emph{CVPR}.

\bibitem[{Gu and Dao(2023)}]{gu2023mamba}
Albert Gu and Tri Dao. 2023.
\newblock {Mamba: Linear-Time Sequence Modeling with Selective State Spaces}.
\newblock \emph{arXiv preprint arXiv:2312.00752}.

\bibitem[{Guo et~al.(2023)Guo, Zhang, Zhu, Tang, Ma, Han, Chen, Gao, Li, Li et~al.}]{guo2023point}
Ziyu Guo, Renrui Zhang, Xiangyang Zhu, Yiwen Tang, Xianzheng Ma, Jiaming Han, Kexin Chen, Peng Gao, Xianzhi Li, Hongsheng Li, et~al. 2023.
\newblock {Point-Bind \& Point-LLM: Aligning Point Cloud with Multi-modality for 3D Understanding, Generation, and Instruction Following}.
\newblock \emph{arXiv preprint arXiv:2309.00615}.

\bibitem[{Gupta et~al.(2019)Gupta, Dollar, and Girshick}]{gupta2019lvis}
Agrim Gupta, Piotr Dollar, and Ross Girshick. 2019.
\newblock {LVIS: A dataset for large vocabulary instance segmentation}.
\newblock In \emph{CVPR}.

\bibitem[{Gurari et~al.(2018)Gurari, Li, Stangl, Guo, Lin, Grauman, Luo, and Bigham}]{gurari2018vizwiz}
Danna Gurari, Qing Li, Abigale~J Stangl, Anhong Guo, Chi Lin, Kristen Grauman, Jiebo Luo, and Jeffrey~P Bigham. 2018.
\newblock {VizWiz Grand Challenge: Answering Visual Questions From Blind People}.
\newblock In \emph{CVPR}.

\bibitem[{Hambardzumyan et~al.(2021)Hambardzumyan, Khachatrian, and May}]{hambardzumyan2021warp}
Karen Hambardzumyan, Hrant Khachatrian, and Jonathan May. 2021.
\newblock {Warp: Word-level adversarial reprogramming}.
\newblock \emph{arXiv preprint arXiv:2101.00121}.

\bibitem[{Han et~al.(2024)Han, Gong, Zhang, Wang, Zhang, Lin, Qiao, Gao, and Yue}]{han2024onellm}
Jiaming Han, Kaixiong Gong, Yiyuan Zhang, Jiaqi Wang, Kaipeng Zhang, Dahua Lin, Yu~Qiao, Peng Gao, and Xiangyu Yue. 2024.
\newblock {OneLLM: One Framework to Align All Modalities with Language}.
\newblock In \emph{CVPR}.

\bibitem[{He et~al.(2022)He, Chen, Xie, Li, Doll{\'a}r, and Girshick}]{he2022masked}
Kaiming He, Xinlei Chen, Saining Xie, Yanghao Li, Piotr Doll{\'a}r, and Ross Girshick. 2022.
\newblock {Masked Autoencoders Are Scalable Vision Learners}.
\newblock In \emph{CVPR}.

\bibitem[{Heusel et~al.(2017)Heusel, Ramsauer, Unterthiner, Nessler, and Hochreiter}]{heusel2017gans}
Martin Heusel, Hubert Ramsauer, Thomas Unterthiner, Bernhard Nessler, and Sepp Hochreiter. 2017.
\newblock {GANs Trained by a Two Time-Scale Update Rule Converge to a Local Nash Equilibrium}.
\newblock In \emph{NeurIPS}.

\bibitem[{Hoffmann et~al.(2022)Hoffmann, Borgeaud, Mensch, Buchatskaya, Cai, Rutherford, Casas, Hendricks, Welbl, Clark et~al.}]{hoffmann2022training}
Jordan Hoffmann, Sebastian Borgeaud, Arthur Mensch, Elena Buchatskaya, Trevor Cai, Eliza Rutherford, Diego de~Las Casas, Lisa~Anne Hendricks, Johannes Welbl, Aidan Clark, et~al. 2022.
\newblock {Training compute-optimal large language models}.
\newblock \emph{arXiv preprint arXiv:2203.15556}.

\bibitem[{Hong et~al.(2023)Hong, Zhen, Chen, Zheng, Du, Chen, and Gan}]{hong20233d}
Yining Hong, Haoyu Zhen, Peihao Chen, Shuhong Zheng, Yilun Du, Zhenfang Chen, and Chuang Gan. 2023.
\newblock {3D-LLM: Injecting the 3D World into Large Language Models}.
\newblock In \emph{NeurIPS}.

\bibitem[{Horawalavithana et~al.(2023)Horawalavithana, Munikoti, Stewart, and Kvinge}]{horawalavithana2023scitune}
Sameera Horawalavithana, Sai Munikoti, Ian Stewart, and Henry Kvinge. 2023.
\newblock {SCITUNE: Aligning Large Language Models with Scientific Multimodal Instructions}.
\newblock \emph{arXiv preprint arXiv:2307.01139}.

\bibitem[{Hu et~al.(2023{\natexlab{a}})Hu, Shi, Xu, Ye, Ye, Yan, Li, Qian, Zhang, and Huang}]{hu2023mplugpaper}
Anwen Hu, Yaya Shi, Haiyang Xu, Jiabo Ye, Qinghao Ye, Ming Yan, Chenliang Li, Qi~Qian, Ji~Zhang, and Fei Huang. 2023{\natexlab{a}}.
\newblock {mPLUG-PaperOwl: Scientific Diagram Analysis with the Multimodal Large Language Model}.
\newblock \emph{arXiv preprint arXiv:2311.18248}.

\bibitem[{Hu et~al.(2021)Hu, Wallis, Allen-Zhu, Li, Wang, Wang, Chen et~al.}]{hu2021lora}
Edward~J Hu, Phillip Wallis, Zeyuan Allen-Zhu, Yuanzhi Li, Shean Wang, Lu~Wang, Weizhu Chen, et~al. 2021.
\newblock {LoRA: Low-Rank Adaptation of Large Language Models}.
\newblock In \emph{ICLR}.

\bibitem[{Hu et~al.(2024)Hu, Xu, Li, Li, Chen, and Tu}]{hu2024bliva}
Wenbo Hu, Yifan Xu, Y~Li, W~Li, Z~Chen, and Z~Tu. 2024.
\newblock {BLIVA: A Simple Multimodal LLM for Better Handling of Text-Rich Visual Questions}.
\newblock In \emph{AAAI}.

\bibitem[{Hu et~al.(2023{\natexlab{b}})Hu, Iscen, Sun, Wang, Chang, Sun, Schmid, Ross, and Fathi}]{hu2023reveal}
Ziniu Hu, Ahmet Iscen, Chen Sun, Zirui Wang, Kai-Wei Chang, Yizhou Sun, Cordelia Schmid, David~A Ross, and Alireza Fathi. 2023{\natexlab{b}}.
\newblock {REVEAL: Retrieval-Augmented Visual-Language Pre-Training With Multi-Source Multimodal Knowledge Memory}.
\newblock In \emph{CVPR}.

\bibitem[{Huang et~al.(2023{\natexlab{a}})Huang, Zhang, Jiang, Qiu, and Lu}]{huang2023visual}
Jiaxing Huang, Jingyi Zhang, Kai Jiang, Han Qiu, and Shijian Lu. 2023{\natexlab{a}}.
\newblock {Visual Instruction Tuning towards General-Purpose Multimodal Model: A Survey}.
\newblock \emph{arXiv preprint arXiv:2312.16602}.

\bibitem[{Huang et~al.(2023{\natexlab{b}})Huang, Dong, Wang, Hao, Singhal, Ma, Lv, Cui, Mohammed, Liu et~al.}]{huang2023language}
Shaohan Huang, Li~Dong, Wenhui Wang, Yaru Hao, Saksham Singhal, Shuming Ma, Tengchao Lv, Lei Cui, Owais~Khan Mohammed, Qiang Liu, et~al. 2023{\natexlab{b}}.
\newblock {Language Is Not All You Need: Aligning Perception with Language Models}.
\newblock \emph{arXiv preprint arXiv:2302.14045}.

\bibitem[{Huang et~al.(2016)Huang, Ferraro, Mostafazadeh, Misra, Devlin, Agrawal, Girshick, He, Kohli, Batra et~al.}]{huang2016visual}
Ting-Hao~K. Huang, Francis Ferraro, Nasrin Mostafazadeh, Ishan Misra, Jacob Devlin, Aishwarya Agrawal, Ross Girshick, Xiaodong He, Pushmeet Kohli, Dhruv Batra, et~al. 2016.
\newblock {Visual Storytelling}.
\newblock In \emph{NAACL}.

\bibitem[{Huang et~al.(2023{\natexlab{c}})Huang, Xie, Wang, Yuan, Cun, Ge, Zhou, Dong, Huang, Zhang et~al.}]{huang2023smartedit}
Yuzhou Huang, Liangbin Xie, Xintao Wang, Ziyang Yuan, Xiaodong Cun, Yixiao Ge, Jiantao Zhou, Chao Dong, Rui Huang, Ruimao Zhang, et~al. 2023{\natexlab{c}}.
\newblock {SmartEdit: Exploring Complex Instruction-based Image Editing with Multimodal Large Language Models}.
\newblock \emph{arXiv preprint arXiv:2312.06739}.

\bibitem[{Hudson and Manning(2019)}]{hudson2019gqa}
Drew~A Hudson and Christopher~D Manning. 2019.
\newblock {GQA: A New Dataset for Real-World Visual Reasoning and Compositional Question Answering}.
\newblock In \emph{CVPR}.

\bibitem[{Hussain et~al.(2023)Hussain, Liu, Sun, and Shan}]{hussain2023m}
Atin~Sakkeer Hussain, Shansong Liu, Chenshuo Sun, and Ying Shan. 2023.
\newblock {M$^2$UGen: Multi-modal Music Understanding and Generation with the Power of Large Language Models}.
\newblock \emph{arXiv preprint arXiv:2311.11255}.

\bibitem[{Jaegle et~al.(2021)Jaegle, Gimeno, Brock, Vinyals, Zisserman, and Carreira}]{jaegle2021perceiver}
Andrew Jaegle, Felix Gimeno, Andy Brock, Oriol Vinyals, Andrew Zisserman, and Joao Carreira. 2021.
\newblock Perceiver: General perception with iterative attention.
\newblock In \emph{ICML}.

\bibitem[{Jiang et~al.(2024)Jiang, Sablayrolles, Roux, Mensch, Savary, Bamford, Chaplot, Casas, Hanna, Bressand et~al.}]{jiang2024mixtral}
Albert~Q Jiang, Alexandre Sablayrolles, Antoine Roux, Arthur Mensch, Blanche Savary, Chris Bamford, Devendra~Singh Chaplot, Diego de~las Casas, Emma~Bou Hanna, Florian Bressand, et~al. 2024.
\newblock {Mixtral of Experts}.
\newblock \emph{arXiv preprint arXiv:2401.04088}.

\bibitem[{Jin et~al.(2023)Jin, Xu, Chen, Liao, Tan, Chen, Lei, Liu, Song, Lei et~al.}]{jin2023unified}
Yang Jin, Kun Xu, Liwei Chen, Chao Liao, Jianchao Tan, Bin Chen, Chenyi Lei, An~Liu, Chengru Song, Xiaoqiang Lei, et~al. 2023.
\newblock {Unified Language-Vision Pretraining with Dynamic Discrete Visual Tokenization}.
\newblock \emph{arXiv preprint arXiv:2309.04669}.

\bibitem[{Jing et~al.(2023)Jing, Li, Chen, Jia, and Du}]{jing2023faithscore}
Liqiang Jing, Ruosen Li, Yunmo Chen, Mengzhao Jia, and Xinya Du. 2023.
\newblock {FAITHSCORE: Evaluating Hallucinations in Large Vision-Language Models}.
\newblock \emph{arXiv preprint arXiv:2311.01477}.

\bibitem[{Karpathy and Fei-Fei(2015)}]{karpathy2015deep}
Andrej Karpathy and Li~Fei-Fei. 2015.
\newblock {Deep visual-semantic alignments for generating image descriptions}.
\newblock In \emph{CVPR}.

\bibitem[{Kazemzadeh et~al.(2014)Kazemzadeh, Ordonez, Matten, and Berg}]{kazemzadeh2014referitgame}
Sahar Kazemzadeh, Vicente Ordonez, Mark Matten, and Tamara Berg. 2014.
\newblock {ReferItGame: Referring to Objects in Photographs of Natural Scenes}.
\newblock In \emph{EMNLP}.

\bibitem[{Kirillov et~al.(2023)Kirillov, Mintun, Ravi, Mao, Rolland, Gustafson, Xiao, Whitehead, Berg, Lo et~al.}]{kirillov2023segment}
Alexander Kirillov, Eric Mintun, Nikhila Ravi, Hanzi Mao, Chloe Rolland, Laura Gustafson, Tete Xiao, Spencer Whitehead, Alexander~C Berg, Wan-Yen Lo, et~al. 2023.
\newblock {Segment Anything}.
\newblock \emph{arXiv preprint arXiv:2304.02643}.

\bibitem[{Koh et~al.(2023{\natexlab{a}})Koh, Fried, and Salakhutdinov}]{koh2023generating}
Jing~Yu Koh, Daniel Fried, and Ruslan Salakhutdinov. 2023{\natexlab{a}}.
\newblock {Generating Images with Multimodal Language Models}.
\newblock In \emph{NeurIPS}.

\bibitem[{Koh et~al.(2023{\natexlab{b}})Koh, Salakhutdinov, and Fried}]{koh2023grounding}
Jing~Yu Koh, Ruslan Salakhutdinov, and Daniel Fried. 2023{\natexlab{b}}.
\newblock {Grounding Language Models to Images for Multimodal Inputs and Outputs}.
\newblock In \emph{ICML}.

\bibitem[{Krishna et~al.(2017)Krishna, Zhu, Groth, Johnson, Hata, Kravitz, Chen, Kalantidis, Li, Shamma et~al.}]{krishna2017visual}
Ranjay Krishna, Yuke Zhu, Oliver Groth, Justin Johnson, Kenji Hata, Joshua Kravitz, Stephanie Chen, Yannis Kalantidis, Li-Jia Li, David~A Shamma, et~al. 2017.
\newblock {Visual Genome: Connecting Language and Vision Using Crowdsourced Dense Image Annotations}.
\newblock \emph{IJCV}, 123:32--73.

\bibitem[{Kuznetsova et~al.(2020)Kuznetsova, Rom, Alldrin, Uijlings, Krasin, Pont-Tuset, Kamali, Popov, Malloci, Kolesnikov et~al.}]{kuznetsova2020open}
Alina Kuznetsova, Hassan Rom, Neil Alldrin, Jasper Uijlings, Ivan Krasin, Jordi Pont-Tuset, Shahab Kamali, Stefan Popov, Matteo Malloci, Alexander Kolesnikov, et~al. 2020.
\newblock {The Open Images Dataset V4: Unified Image Classification, Object Detection, and Visual Relationship Detection at Scale}.
\newblock \emph{IJCV}, 128:1956--1981.

\bibitem[{Lai et~al.(2023)Lai, Tian, Chen, Li, Yuan, Liu, and Jia}]{lai2023lisa}
Xin Lai, Zhuotao Tian, Yukang Chen, Yanwei Li, Yuhui Yuan, Shu Liu, and Jiaya Jia. 2023.
\newblock {LISA: Reasoning Segmentation via Large Language Model}.
\newblock \emph{arXiv preprint arXiv:2308.00692}.

\bibitem[{Lauren{\c{c}}on et~al.(2024)Lauren{\c{c}}on, Saulnier, Tronchon, Bekman, Singh, Lozhkov, Wang, Karamcheti, Rush, Kiela et~al.}]{laurenccon2023obelisc}
Hugo Lauren{\c{c}}on, Lucile Saulnier, L{\'e}o Tronchon, Stas Bekman, Amanpreet Singh, Anton Lozhkov, Thomas Wang, Siddharth Karamcheti, Alexander Rush, Douwe Kiela, et~al. 2024.
\newblock Obelics: An open web-scale filtered dataset of interleaved image-text documents.
\newblock In \emph{NeurIPS}.

\bibitem[{Lester et~al.(2021)Lester, Al-Rfou, and Constant}]{lester2021power}
Brian Lester, Rami Al-Rfou, and Noah Constant. 2021.
\newblock {The power of scale for parameter-efficient prompt tuning}.
\newblock \emph{arXiv preprint arXiv:2104.08691}.

\bibitem[{Lewis et~al.(2020)Lewis, Perez, Piktus, Petroni, Karpukhin, Goyal, K{\"u}ttler, Lewis, Yih, Rockt{\"a}schel et~al.}]{lewis2020retrieval}
Patrick Lewis, Ethan Perez, Aleksandra Piktus, Fabio Petroni, Vladimir Karpukhin, Naman Goyal, Heinrich K{\"u}ttler, Mike Lewis, Wen-tau Yih, Tim Rockt{\"a}schel, et~al. 2020.
\newblock {Retrieval-Augmented Generation for Knowledge-Intensive NLP Tasks}.
\newblock In \emph{NeurIPS}.

\bibitem[{Li et~al.(2023{\natexlab{a}})Li, Zhang, Yang, Zhang, Pu, and Liu}]{li2023otterhd}
Bo~Li, Peiyuan Zhang, Jingkang Yang, Yuanhan Zhang, Fanyi Pu, and Ziwei Liu. 2023{\natexlab{a}}.
\newblock {OtterHD: A High-Resolution Multi-modality Model}.
\newblock \emph{arXiv preprint arXiv:2311.04219}.

\bibitem[{Li et~al.(2023{\natexlab{b}})Li, Zhang, Chen, Wang, Yang, and Liu}]{li2023otter}
Bo~Li, Yuanhan Zhang, Liangyu Chen, Jinghao Wang, Jingkang Yang, and Ziwei Liu. 2023{\natexlab{b}}.
\newblock {Otter: A Multi-Modal Model with In-Context Instruction Tuning}.
\newblock \emph{arXiv preprint arXiv:2305.03726}.

\bibitem[{Li et~al.(2023{\natexlab{c}})Li, Wang, Wang, Ge, Ge, and Shan}]{li2023seed}
Bohao Li, Rui Wang, Guangzhi Wang, Yuying Ge, Yixiao Ge, and Ying Shan. 2023{\natexlab{c}}.
\newblock {SEED-Bench: Benchmarking Multimodal LLMs with Generative Comprehension}.
\newblock \emph{arXiv preprint arXiv:2307.16125}.

\bibitem[{Li et~al.(2023{\natexlab{d}})Li, Wong, Zhang, Usuyama, Liu, Yang, Naumann, Poon, and Gao}]{li2023llava}
Chunyuan Li, Cliff Wong, Sheng Zhang, Naoto Usuyama, Haotian Liu, Jianwei Yang, Tristan Naumann, Hoifung Poon, and Jianfeng Gao. 2023{\natexlab{d}}.
\newblock {LLaVA-Med: Training a Large Language-and-Vision Assistant for Biomedicine in One Day}.
\newblock \emph{arXiv preprint arXiv:2306.00890}.

\bibitem[{Li et~al.(2023{\natexlab{e}})Li, Zhang, Sun, Zou, Liu, Yang, Li, Zhang, and Gao}]{li2023semantic}
Feng Li, Hao Zhang, Peize Sun, Xueyan Zou, Shilong Liu, Jianwei Yang, Chunyuan Li, Lei Zhang, and Jianfeng Gao. 2023{\natexlab{e}}.
\newblock {Semantic-SAM: Segment and recognize anything at any granularity}.
\newblock \emph{arXiv preprint arXiv:2307.04767}.

\bibitem[{Li et~al.(2023{\natexlab{f}})Li, Li, Cai, Wang, Liu, Watanabe, Yang, and Shi}]{li2023textbind}
Huayang Li, Siheng Li, Deng Cai, Longyue Wang, Lemao Liu, Taro Watanabe, Yujiu Yang, and Shuming Shi. 2023{\natexlab{f}}.
\newblock {TextBind: Multi-turn Interleaved Multimodal Instruction-following in the Wild}.
\newblock \emph{arXiv preprint arXiv:2309.08637}.

\bibitem[{Li et~al.(2023{\natexlab{g}})Li, Li, Savarese, and Hoi}]{li2023blip}
Junnan Li, Dongxu Li, Silvio Savarese, and Steven Hoi. 2023{\natexlab{g}}.
\newblock {BLIP-2: Bootstrapping Language-Image Pre-training with Frozen Image Encoders and Large Language Models}.
\newblock \emph{arXiv preprint arXiv:2301.12597}.

\bibitem[{Li et~al.(2023{\natexlab{h}})Li, He, Wang, Li, Wang, Luo, Wang, Wang, and Qiao}]{li2023videochat}
KunChang Li, Yinan He, Yi~Wang, Yizhuo Li, Wenhai Wang, Ping Luo, Yali Wang, Limin Wang, and Yu~Qiao. 2023{\natexlab{h}}.
\newblock {VideoChat: Chat-Centric Video Understanding}.
\newblock \emph{arXiv preprint arXiv:2305.06355}.

\bibitem[{Li et~al.(2019)Li, Yatskar, Yin, Hsieh, and Chang}]{li2019visualbert}
Liunian~Harold Li, Mark Yatskar, Da~Yin, Cho-Jui Hsieh, and Kai-Wei Chang. 2019.
\newblock {VisualBERT: A Simple and Performant Baseline for Vision and Language}.
\newblock \emph{arXiv preprint arXiv:1908.03557}.

\bibitem[{Li et~al.(2022)Li, Zhang, Zhang, Yang, Li, Zhong, Wang, Yuan, Zhang, Hwang et~al.}]{li2022grounded}
Liunian~Harold Li, Pengchuan Zhang, Haotian Zhang, Jianwei Yang, Chunyuan Li, Yiwu Zhong, Lijuan Wang, Lu~Yuan, Lei Zhang, Jenq-Neng Hwang, et~al. 2022.
\newblock {Grounded language-image pre-training}.
\newblock In \emph{CVPR}.

\bibitem[{Li and Liang(2021)}]{li2021prefix}
Xiang~Lisa Li and Percy Liang. 2021.
\newblock {Prefix-tuning: Optimizing continuous prompts for generation}.
\newblock \emph{arXiv preprint arXiv:2101.00190}.

\bibitem[{Li et~al.(2023{\natexlab{i}})Li, Zhang, Yu, Wang, Fu, Lin, Shen, Chen, and Wei}]{li2023stablellava}
Yanda Li, Chi Zhang, Gang Yu, Zhibin Wang, Bin Fu, Guosheng Lin, Chunhua Shen, Ling Chen, and Yunchao Wei. 2023{\natexlab{i}}.
\newblock {StableLLaVA: Enhanced Visual Instruction Tuning with Synthesized Image-Dialogue Data}.
\newblock \emph{arXiv preprint arXiv:2308.10253}.

\bibitem[{Li et~al.(2023{\natexlab{j}})Li, Wang, and Jia}]{li2023llama}
Yanwei Li, Chengyao Wang, and Jiaya Jia. 2023{\natexlab{j}}.
\newblock {LLaMA-VID: An Image is Worth 2 Tokens in Large Language Models}.
\newblock \emph{arXiv preprint arXiv:2311.17043}.

\bibitem[{Li et~al.(2023{\natexlab{k}})Li, Du, Zhou, Wang, Zhao, and Wen}]{li2023evaluating}
Yifan Li, Yifan Du, Kun Zhou, Jinpeng Wang, Wayne~Xin Zhao, and Ji-Rong Wen. 2023{\natexlab{k}}.
\newblock {Evaluating Object Hallucination in Large Vision-Language Models}.
\newblock \emph{arXiv preprint arXiv:2305.10355}.

\bibitem[{Li et~al.(2023{\natexlab{l}})Li, Yang, Liu, Ma, Zhang, Yang, Sun, Liu, and Bai}]{li2023monkey}
Zhang Li, Biao Yang, Qiang Liu, Zhiyin Ma, Shuo Zhang, Jingxu Yang, Yabo Sun, Yuliang Liu, and Xiang Bai. 2023{\natexlab{l}}.
\newblock {Monkey: Image Resolution and Text Label Are Important Things for Large Multi-modal Models}.
\newblock \emph{arXiv preprint arXiv:2311.06607}.

\bibitem[{Lin et~al.(2023{\natexlab{a}})Lin, Yin, Ping, Lu, Molchanov, Tao, Mao, Kautz, Shoeybi, and Han}]{lin2023vila}
Ji~Lin, Hongxu Yin, Wei Ping, Yao Lu, Pavlo Molchanov, Andrew Tao, Huizi Mao, Jan Kautz, Mohammad Shoeybi, and Song Han. 2023{\natexlab{a}}.
\newblock {VILA: On Pre-training for Visual Language Models}.
\newblock \emph{arXiv preprint arXiv:2312.07533}.

\bibitem[{Lin et~al.(2014)Lin, Maire, Belongie, Hays, Perona, Ramanan, Doll{\'a}r, and Zitnick}]{lin2014microsoft}
Tsung-Yi Lin, Michael Maire, Serge Belongie, James Hays, Pietro Perona, Deva Ramanan, Piotr Doll{\'a}r, and C~Lawrence Zitnick. 2014.
\newblock {Microsoft COCO: Common Objects in Context}.
\newblock In \emph{ECCV}.

\bibitem[{Lin et~al.(2023{\natexlab{b}})Lin, Liu, Zhang, Gao, Qiu, Xiao, Qiu, Lin, Shao, Chen et~al.}]{lin2023sphinx}
Ziyi Lin, Chris Liu, Renrui Zhang, Peng Gao, Longtian Qiu, Han Xiao, Han Qiu, Chen Lin, Wenqi Shao, Keqin Chen, et~al. 2023{\natexlab{b}}.
\newblock {SPHINX: The Joint Mixing of Weights, Tasks, and Visual Embeddings for Multi-modal Large Language Models}.
\newblock \emph{arXiv preprint arXiv:2311.07575}.

\bibitem[{Liu et~al.(2023{\natexlab{a}})Liu, Emerson, and Collier}]{liu2023vsr}
Fangyu Liu, Guy Emerson, and Nigel Collier. 2023{\natexlab{a}}.
\newblock {Visual Spatial Reasoning}.
\newblock \emph{TACL}, 11:635--651.

\bibitem[{Liu et~al.(2023{\natexlab{b}})Liu, Lin, Li, Wang, Yacoob, and Wang}]{liu2023aligning}
Fuxiao Liu, Kevin Lin, Linjie Li, Jianfeng Wang, Yaser Yacoob, and Lijuan Wang. 2023{\natexlab{b}}.
\newblock {Aligning Large Multi-Modal Model with Robust Instruction Tuning}.
\newblock \emph{arXiv preprint arXiv:2306.14565}.

\bibitem[{Liu et~al.(2023{\natexlab{c}})Liu, Lin, Li, Wang, Yacoob, and Wang}]{liu2023mitigating}
Fuxiao Liu, Kevin Lin, Linjie Li, Jianfeng Wang, Yaser Yacoob, and Lijuan Wang. 2023{\natexlab{c}}.
\newblock {Mitigating Hallucination in Large Multi-Modal Models via Robust Instruction Tuning}.
\newblock \emph{arXiv preprint arXiv:2306.14565}.

\bibitem[{Liu et~al.(2024{\natexlab{a}})Liu, You, Han, Wang, Zhai, Liu, Tao, Huang, He, and Yang}]{liu2024infimm}
Haogeng Liu, Quanzeng You, Xiaotian Han, Yiqi Wang, Bohan Zhai, Yongfei Liu, Yunzhe Tao, Huaibo Huang, Ran He, and Hongxia Yang. 2024{\natexlab{a}}.
\newblock {InfiMM-HD: A Leap Forward in High-Resolution Multimodal Understanding}.
\newblock \emph{arXiv preprint arXiv:2403.01487}.

\bibitem[{Liu et~al.(2023{\natexlab{d}})Liu, Li, Li, and Lee}]{liu2023improved}
Haotian Liu, Chunyuan Li, Yuheng Li, and Yong~Jae Lee. 2023{\natexlab{d}}.
\newblock {Improved Baselines with Visual Instruction Tuning}.
\newblock \emph{arXiv preprint arXiv:2310.03744}.

\bibitem[{Liu et~al.(2024{\natexlab{b}})Liu, Li, Li, Li, Zhang, Shen, and Lee}]{liu2024llava}
Haotian Liu, Chunyuan Li, Yuheng Li, Bo~Li, Yuanhan Zhang, Sheng Shen, and Yong~Jae Lee. 2024{\natexlab{b}}.
\newblock {LLaVA-NeXT: Improved reasoning, OCR, and world knowledge}.

\bibitem[{Liu et~al.(2023{\natexlab{e}})Liu, Li, Wu, and Lee}]{liu2023visual}
Haotian Liu, Chunyuan Li, Qingyang Wu, and Yong~Jae Lee. 2023{\natexlab{e}}.
\newblock {Visual Instruction Tuning}.
\newblock In \emph{NeurIPS}.

\bibitem[{Liu et~al.(2023{\natexlab{f}})Liu, Wang, Ye, Chong, Zhou, and Hua}]{liu2023qilin}
Junling Liu, Ziming Wang, Qichen Ye, Dading Chong, Peilin Zhou, and Yining Hua. 2023{\natexlab{f}}.
\newblock {Qilin-Med-VL: Towards Chinese Large Vision-Language Model for General Healthcare}.
\newblock \emph{arXiv preprint arXiv:2310.17956}.

\bibitem[{Liu et~al.(2023{\natexlab{g}})Liu, Li, Ge, Shan, Li, and Li}]{liu2023one}
Ruyang Liu, Chen Li, Yixiao Ge, Ying Shan, Thomas~H Li, and Ge~Li. 2023{\natexlab{g}}.
\newblock {One For All: Video Conversation is Feasible Without Video Instruction Tuning}.
\newblock \emph{arXiv preprint arXiv:2309.15785}.

\bibitem[{Liu et~al.(2023{\natexlab{h}})Liu, Cheng, Liu, Zhang, Li, Ren, Zou, Yang, Su, Zhu et~al.}]{liu2023llava}
Shilong Liu, Hao Cheng, Haotian Liu, Hao Zhang, Feng Li, Tianhe Ren, Xueyan Zou, Jianwei Yang, Hang Su, Jun Zhu, et~al. 2023{\natexlab{h}}.
\newblock {LLaVA-Plus: Learning to Use Tools for Creating Multimodal Agents}.
\newblock \emph{arXiv preprint arXiv:2311.05437}.

\bibitem[{Liu et~al.(2023{\natexlab{i}})Liu, Zeng, Ren, Li, Zhang, Yang, Li, Yang, Su, Zhu et~al.}]{liu2023grounding}
Shilong Liu, Zhaoyang Zeng, Tianhe Ren, Feng Li, Hao Zhang, Jie Yang, Chunyuan Li, Jianwei Yang, Hang Su, Jun Zhu, et~al. 2023{\natexlab{i}}.
\newblock {Grounding DINO: Marrying dino with grounded pre-training for open-set object detection}.
\newblock \emph{arXiv preprint arXiv:2303.05499}.

\bibitem[{Liu et~al.(2023{\natexlab{j}})Liu, Zheng, Du, Ding, Qian, Yang, and Tang}]{liu2023gpt}
Xiao Liu, Yanan Zheng, Zhengxiao Du, Ming Ding, Yujie Qian, Zhilin Yang, and Jie Tang. 2023{\natexlab{j}}.
\newblock {GPT understands, too}.
\newblock \emph{AI Open}.

\bibitem[{Liu et~al.(2023{\natexlab{k}})Liu, Duan, Zhang, Li, Zhang, Zhao, Yuan, Wang, He, Liu et~al.}]{liu2023mmbench}
Yuan Liu, Haodong Duan, Yuanhan Zhang, Bo~Li, Songyang Zhang, Wangbo Zhao, Yike Yuan, Jiaqi Wang, Conghui He, Ziwei Liu, et~al. 2023{\natexlab{k}}.
\newblock {MMBench: Is Your Multi-modal Model an All-around Player?}
\newblock \emph{arXiv preprint arXiv:2307.06281}.

\bibitem[{Liu et~al.(2023{\natexlab{l}})Liu, He, Wang, Wang, Wang, Chen, Zhang, Lai, Yang, Li et~al.}]{liu2023interngpt}
Zhaoyang Liu, Yinan He, Wenhai Wang, Weiyun Wang, Yi~Wang, Shoufa Chen, Qinglong Zhang, Zeqiang Lai, Yang Yang, Qingyun Li, et~al. 2023{\natexlab{l}}.
\newblock {InternGPT: Solving Vision-Centric Tasks by Interacting with ChatGPT Beyond Language}.
\newblock \emph{arXiv preprint arXiv:2305.05662}.

\bibitem[{Lu et~al.(2019)Lu, Batra, Parikh, and Lee}]{lu2019vilbert}
Jiasen Lu, Dhruv Batra, Devi Parikh, and Stefan Lee. 2019.
\newblock {ViLBERT: Pretraining Task-Agnostic Visiolinguistic Representations for Vision-and-Language Tasks}.
\newblock In \emph{NeurIPS}.

\bibitem[{Lu et~al.(2023{\natexlab{a}})Lu, Clark, Lee, Zhang, Khosla, Marten, Hoiem, and Kembhavi}]{lu2023unified}
Jiasen Lu, Christopher Clark, Sangho Lee, Zichen Zhang, Savya Khosla, Ryan Marten, Derek Hoiem, and Aniruddha Kembhavi. 2023{\natexlab{a}}.
\newblock {Unified-IO 2: Scaling Autoregressive Multimodal Models with Vision, Language, Audio, and Action}.
\newblock \emph{arXiv preprint arXiv:2312.17172}.

\bibitem[{Lu et~al.(2023{\natexlab{b}})Lu, Bansal, Xia, Liu, Li, Hajishirzi, Cheng, Chang, Galley, and Gao}]{lu2023mathvista}
Pan Lu, Hritik Bansal, Tony Xia, Jiacheng Liu, Chunyuan Li, Hannaneh Hajishirzi, Hao Cheng, Kai-Wei Chang, Michel Galley, and Jianfeng Gao. 2023{\natexlab{b}}.
\newblock {MathVista: Evaluating Mathematical Reasoning of Foundation Models in Visual Contexts}.
\newblock \emph{arXiv preprint arXiv:2310.02255}.

\bibitem[{Lu et~al.(2022)Lu, Mishra, Xia, Qiu, Chang, Zhu, Tafjord, Clark, and Kalyan}]{lu2022learn}
Pan Lu, Swaroop Mishra, Tanglin Xia, Liang Qiu, Kai-Wei Chang, Song-Chun Zhu, Oyvind Tafjord, Peter Clark, and Ashwin Kalyan. 2022.
\newblock {Learn to Explain: Multimodal Reasoning via Thought Chains for Science Question Answering}.
\newblock In \emph{NeurIPS}.

\bibitem[{Lu et~al.(2021)Lu, Qiu, Chen, Xia, Zhao, Zhang, Yu, Liang, and Zhu}]{lu2021iconqa}
Pan Lu, Liang Qiu, Jiaqi Chen, Tony Xia, Yizhou Zhao, Wei Zhang, Zhou Yu, Xiaodan Liang, and Song-Chun Zhu. 2021.
\newblock {IconQA: A New Benchmark for Abstract Diagram Understanding and Visual Language Reasoning}.
\newblock In \emph{NeurIPS}.

\bibitem[{Luo et~al.(2023)Luo, Zhou, Ren, Chen, Sun, and Ji}]{luo2023cheap}
Gen Luo, Yiyi Zhou, Tianhe Ren, Shengxin Chen, Xiaoshuai Sun, and Rongrong Ji. 2023.
\newblock {Cheap and Quick: Efficient Vision-Language Instruction Tuning for Large Language Models}.
\newblock \emph{arXiv preprint arXiv:2305.15023}.

\bibitem[{Luo et~al.(2024)Luo, Zhou, Zhang, Zheng, Sun, and Ji}]{luo2024feast}
Gen Luo, Yiyi Zhou, Yuxin Zhang, Xiawu Zheng, Xiaoshuai Sun, and Rongrong Ji. 2024.
\newblock {Feast Your Eyes: Mixture-of-Resolution Adaptation for Multimodal Large Language Models}.
\newblock \emph{arXiv preprint arXiv:2403.03003}.

\bibitem[{Lv et~al.(2023)Lv, Huang, Chen, Cui, Ma, Chang, Huang, Wang, Dong, Luo et~al.}]{lv2023kosmos}
Tengchao Lv, Yupan Huang, Jingye Chen, Lei Cui, Shuming Ma, Yaoyao Chang, Shaohan Huang, Wenhui Wang, Li~Dong, Weiyao Luo, et~al. 2023.
\newblock {Kosmos-2.5: A Multimodal Literate Model}.
\newblock \emph{arXiv preprint arXiv:2309.11419}.

\bibitem[{Ma et~al.(2023{\natexlab{a}})Ma, Jin, Wang, Xian, Feng, and Yang}]{ma2023vista}
Fan Ma, Xiaojie Jin, Heng Wang, Yuchen Xian, Jiashi Feng, and Yi~Yang. 2023{\natexlab{a}}.
\newblock {Vista-LLaMA: Reliable Video Narrator via Equal Distance to Visual Tokens}.
\newblock \emph{arXiv preprint arXiv:2312.08870}.

\bibitem[{Ma et~al.(2023{\natexlab{b}})Ma, Cao, Sun, Pavone, and Xiao}]{ma2023dolphins}
Yingzi Ma, Yulong Cao, Jiachen Sun, Marco Pavone, and Chaowei Xiao. 2023{\natexlab{b}}.
\newblock {Dolphins: Multimodal Language Model for Driving}.
\newblock \emph{arXiv preprint arXiv:2312.00438}.

\bibitem[{Maaz et~al.(2023)Maaz, Rasheed, Khan, and Khan}]{maaz2023video}
Muhammad Maaz, Hanoona Rasheed, Salman Khan, and Fahad~Shahbaz Khan. 2023.
\newblock {Video-ChatGPT: Towards Detailed Video Understanding via Large Vision and Language Models}.
\newblock \emph{arXiv preprint arXiv:2306.05424}.

\bibitem[{Mao et~al.(2016)Mao, Huang, Toshev, Camburu, Yuille, and Murphy}]{mao2016generation}
Junhua Mao, Jonathan Huang, Alexander Toshev, Oana Camburu, Alan~L Yuille, and Kevin Murphy. 2016.
\newblock {Generation and Comprehension of Unambiguous Object Descriptions}.
\newblock In \emph{CVPR}.

\bibitem[{Marino et~al.(2019)Marino, Rastegari, Farhadi, and Mottaghi}]{marino2019ok}
Kenneth Marino, Mohammad Rastegari, Ali Farhadi, and Roozbeh Mottaghi. 2019.
\newblock {OK-VQA: A Visual Question Answering Benchmark Requiring External Knowledge}.
\newblock In \emph{CVPR}.

\bibitem[{Mensink et~al.(2023)Mensink, Uijlings, Castrejon, Goel, Cadar, Zhou, Sha, Araujo, and Ferrari}]{mensink2023encyclopedic}
Thomas Mensink, Jasper Uijlings, Lluis Castrejon, Arushi Goel, Felipe Cadar, Howard Zhou, Fei Sha, Andr{\'e} Araujo, and Vittorio Ferrari. 2023.
\newblock {Encyclopedic VQA: Visual questions about detailed properties of fine-grained categories}.
\newblock In \emph{ICCV}.

\bibitem[{Mishra et~al.(2019)Mishra, Shekhar, Singh, and Chakraborty}]{mishra2019ocr}
Anand Mishra, Shashank Shekhar, Ajeet~Kumar Singh, and Anirban Chakraborty. 2019.
\newblock {Ocr-vqa: Visual question answering by reading text in images}.
\newblock In \emph{ICDAR}.

\bibitem[{Moon et~al.(2023)Moon, Madotto, Lin, Nagarajan, Smith, Jain, Yeh, Murugesan, Heidari, Liu et~al.}]{moon2023anymal}
Seungwhan Moon, Andrea Madotto, Zhaojiang Lin, Tushar Nagarajan, Matt Smith, Shashank Jain, Chun-Fu Yeh, Prakash Murugesan, Peyman Heidari, Yue Liu, et~al. 2023.
\newblock {AnyMAL: An Efficient and Scalable Any-Modality Augmented Language Model}.
\newblock \emph{arXiv preprint arXiv:2309.16058}.

\bibitem[{MosaicML(2023)}]{Introduc1online}
MosaicML. 2023.
\newblock {Introducing MPT-7B: A New Standard for Open-Source, Commercially Usable LLMs}.

\bibitem[{Mu et~al.(2023)Mu, Zhang, Hu, Wang, Ding, Jin, Wang, Dai, Qiao, and Luo}]{mu2023embodiedgpt}
Yao Mu, Qinglong Zhang, Mengkang Hu, Wenhai Wang, Mingyu Ding, Jun Jin, Bin Wang, Jifeng Dai, Yu~Qiao, and Ping Luo. 2023.
\newblock {EmbodiedGPT: Vision-Language Pre-Training via Embodied Chain of Thought}.
\newblock \emph{arXiv preprint arXiv:2305.15021}.

\bibitem[{Munasinghe et~al.(2023)Munasinghe, Thushara, Maaz, Rasheed, Khan, Shah, and Khan}]{munasinghe2023pg}
Shehan Munasinghe, Rusiru Thushara, Muhammad Maaz, Hanoona~Abdul Rasheed, Salman Khan, Mubarak Shah, and Fahad Khan. 2023.
\newblock {PG-Video-LLaVA: Pixel Grounding Large Video-Language Models}.
\newblock \emph{arXiv preprint arXiv:2311.13435}.

\bibitem[{OpenAI(2022)}]{chatgpt}
OpenAI. 2022.
\newblock {Introducing ChatGPT}.

\bibitem[{Ouyang et~al.(2022)Ouyang, Wu, Jiang, Almeida, Wainwright, Mishkin, Zhang, Agarwal, Slama, Ray et~al.}]{ouyang2022training}
Long Ouyang, Jeffrey Wu, Xu~Jiang, Diogo Almeida, Carroll Wainwright, Pamela Mishkin, Chong Zhang, Sandhini Agarwal, Katarina Slama, Alex Ray, et~al. 2022.
\newblock Training language models to follow instructions with human feedback.
\newblock In \emph{NeurIPS}.

\bibitem[{Pan et~al.(2023)Pan, Dong, Huang, Peng, Chen, and Wei}]{pan2023kosmos}
Xichen Pan, Li~Dong, Shaohan Huang, Zhiliang Peng, Wenhu Chen, and Furu Wei. 2023.
\newblock {Kosmos-G: Generating Images in Context with Multimodal Large Language Models}.
\newblock \emph{arXiv preprint arXiv:2310.02992}.

\bibitem[{Panagopoulou et~al.(2023)Panagopoulou, Xue, Yu, Li, Li, Joty, Xu, Savarese, Xiong, and Niebles}]{panagopoulou2023x}
Artemis Panagopoulou, Le~Xue, Ning Yu, Junnan Li, Dongxu Li, Shafiq Joty, Ran Xu, Silvio Savarese, Caiming Xiong, and Juan~Carlos Niebles. 2023.
\newblock {X-InstructBLIP: A Framework for aligning X-Modal instruction-aware representations to LLMs and Emergent Cross-modal Reasoning}.
\newblock \emph{arXiv preprint arXiv:2311.18799}.

\bibitem[{Peng et~al.(2023)Peng, Wang, Dong, Hao, Huang, Ma, and Wei}]{peng2023kosmos}
Zhiliang Peng, Wenhui Wang, Li~Dong, Yaru Hao, Shaohan Huang, Shuming Ma, and Furu Wei. 2023.
\newblock {Kosmos-2: Grounding Multimodal Large Language Models to the World}.
\newblock \emph{arXiv preprint arXiv:2306.14824}.

\bibitem[{Pi et~al.(2023)Pi, Gao, Diao, Pan, Dong, Zhang, Yao, Han, Xu, and Zhang}]{pi2023detgpt}
Renjie Pi, Jiahui Gao, Shizhe Diao, Rui Pan, Hanze Dong, Jipeng Zhang, Lewei Yao, Jianhua Han, Hang Xu, and Lingpeng Kong~Tong Zhang. 2023.
\newblock {DetGPT: Detect What You Need via Reasoning}.
\newblock \emph{arXiv preprint arXiv:2305.14167}.

\bibitem[{Pi et~al.(2024)Pi, Han, Xie, Pan, Lian, Dong, Zhang, and Zhang}]{pi2024mllm}
Renjie Pi, Tianyang Han, Yueqi Xie, Rui Pan, Qing Lian, Hanze Dong, Jipeng Zhang, and Tong Zhang. 2024.
\newblock {MLLM-Protector: Ensuring MLLM's Safety without Hurting Performance}.
\newblock \emph{arXiv preprint arXiv:2401.02906}.

\bibitem[{Podell et~al.(2023)Podell, English, Lacey, Blattmann, Dockhorn, M{\"u}ller, Penna, and Rombach}]{podell2023sdxl}
Dustin Podell, Zion English, Kyle Lacey, Andreas Blattmann, Tim Dockhorn, Jonas M{\"u}ller, Joe Penna, and Robin Rombach. 2023.
\newblock {SDXL: Improving Latent Diffusion Models for High-Resolution Image Synthesis}.
\newblock \emph{arXiv preprint arXiv:2307.01952}.

\bibitem[{Poppi et~al.(2024)Poppi, Poppi, Cocchi, Cornia, Baraldi, and Cucchiara}]{poppi2024removing}
Samuele Poppi, Tobia Poppi, Federico Cocchi, Marcella Cornia, Lorenzo Baraldi, and Rita Cucchiara. 2024.
\newblock {Safe-CLIP: Removing NSFW Concepts from Vision-and-Language Models}.
\newblock \emph{arXiv preprint arXiv:2311.16254}.

\bibitem[{Pramanick et~al.(2023)Pramanick, Han, Hou, Nag, Lim, Ballas, Wang, Chellappa, and Almahairi}]{pramanick2023jack}
Shraman Pramanick, Guangxing Han, Rui Hou, Sayan Nag, Ser-Nam Lim, Nicolas Ballas, Qifan Wang, Rama Chellappa, and Amjad Almahairi. 2023.
\newblock {Jack of All Tasks, Master of Many: Designing General-purpose Coarse-to-Fine Vision-Language Model}.
\newblock \emph{arXiv preprint arXiv:2312.12423}.

\bibitem[{Qi et~al.(2024)Qi, Chen, Yang, Shen, Li, Guo, Xu, and Yang}]{qi2024generalizable}
Lu~Qi, Yi-Wen Chen, Lehan Yang, Tiancheng Shen, Xiangtai Li, Weidong Guo, Yu~Xu, and Ming-Hsuan Yang. 2024.
\newblock {Generalizable Entity Grounding via Assistance of Large Language Model}.
\newblock \emph{arXiv preprint arXiv:2402.02555}.

\bibitem[{Qiao et~al.(2024)Qiao, Yu, Guo, Chen, Zhao, Sun, Wu, and Liu}]{qiao2024vl}
Yanyuan Qiao, Zheng Yu, Longteng Guo, Sihan Chen, Zijia Zhao, Mingzhen Sun, Qi~Wu, and Jing Liu. 2024.
\newblock {VL-Mamba: Exploring State Space Models for Multimodal Learning}.
\newblock \emph{arXiv preprint arXiv:2403.13600}.

\bibitem[{Radford et~al.(2021)Radford, Kim, Hallacy, Ramesh, Goh, Agarwal, Sastry, Askell, Mishkin, Clark et~al.}]{radford2021learning}
Alec Radford, Jong~Wook Kim, Chris Hallacy, Aditya Ramesh, Gabriel Goh, Sandhini Agarwal, Girish Sastry, Amanda Askell, Pamela Mishkin, Jack Clark, et~al. 2021.
\newblock Learning transferable visual models from natural language supervision.
\newblock In \emph{ICML}.

\bibitem[{Raffel et~al.(2020)Raffel, Shazeer, Roberts, Lee, Narang, Matena, Zhou, Li, and Liu}]{raffel2020exploring}
Colin Raffel, Noam Shazeer, Adam Roberts, Katherine Lee, Sharan Narang, Michael Matena, Yanqi Zhou, Wei Li, and Peter~J Liu. 2020.
\newblock {Exploring the limits of transfer learning with a unified text-to-text transformer}.
\newblock \emph{JMLR}, 21(1):5485--5551.

\bibitem[{Rasheed et~al.(2023)Rasheed, Maaz, Shaji, Shaker, Khan, Cholakkal, Anwer, Xing, Yang, and Khan}]{rasheed2023glamm}
Hanoona Rasheed, Muhammad Maaz, Sahal Shaji, Abdelrahman Shaker, Salman Khan, Hisham Cholakkal, Rao~M Anwer, Erix Xing, Ming-Hsuan Yang, and Fahad~S Khan. 2023.
\newblock {GLaMM : Pixel Grounding Large Multimodal Model}.
\newblock \emph{arXiv preprint arXiv:2311.03356}.

\bibitem[{Ren et~al.(2023{\natexlab{a}})Ren, Yao, Li, Sun, and Hou}]{ren2023timechat}
Shuhuai Ren, Linli Yao, Shicheng Li, Xu~Sun, and Lu~Hou. 2023{\natexlab{a}}.
\newblock {TimeChat: A Time-sensitive Multimodal Large Language Model for Long Video Understanding}.
\newblock \emph{arXiv preprint arXiv:2312.02051}.

\bibitem[{Ren et~al.(2023{\natexlab{b}})Ren, Huang, Wei, Zhao, Fu, Feng, and Jin}]{ren2023pixellm}
Zhongwei Ren, Zhicheng Huang, Yunchao Wei, Yao Zhao, Dongmei Fu, Jiashi Feng, and Xiaojie Jin. 2023{\natexlab{b}}.
\newblock {PixelLM: Pixel Reasoning with Large Multimodal Model}.
\newblock \emph{arXiv preprint arXiv:2312.02228}.

\bibitem[{Rombach et~al.(2022)Rombach, Blattmann, Lorenz, Esser, and Ommer}]{rombach2022high}
Robin Rombach, Andreas Blattmann, Dominik Lorenz, Patrick Esser, and Bj{\"o}rn Ommer. 2022.
\newblock {High-Resolution Image Synthesis with Latent Diffusion Models}.
\newblock In \emph{CVPR}.

\bibitem[{Ronneberger et~al.(2015)Ronneberger, Fischer, and Brox}]{ronneberger2015u}
Olaf Ronneberger, Philipp Fischer, and Thomas Brox. 2015.
\newblock {U-Net: Convolutional Networks for Biomedical Image Segmentation}.
\newblock In \emph{MICCAI}.

\bibitem[{Ruiz et~al.(2023)Ruiz, Li, Jampani, Pritch, Rubinstein, and Aberman}]{ruiz2023dreambooth}
Nataniel Ruiz, Yuanzhen Li, Varun Jampani, Yael Pritch, Michael Rubinstein, and Kfir Aberman. 2023.
\newblock {DreamBooth: Fine Tuning Text-to-Image Diffusion Models for Subject-Driven Generation}.
\newblock In \emph{CVPR}.

\bibitem[{Schramowski et~al.(2023)Schramowski, Brack, Deiseroth, and Kersting}]{schramowski2023safe}
Patrick Schramowski, Manuel Brack, Bj{\"o}rn Deiseroth, and Kristian Kersting. 2023.
\newblock {Safe Latent Diffusion: Mitigating Inappropriate Degeneration in Diffusion Models}.
\newblock In \emph{CVPR}.

\bibitem[{Schuhmann et~al.(2022)Schuhmann, Beaumont, Vencu, Gordon, Wightman, Cherti, Coombes, Katta, Mullis, Wortsman et~al.}]{schuhmann2022laion}
Christoph Schuhmann, Romain Beaumont, Richard Vencu, Cade Gordon, Ross Wightman, Mehdi Cherti, Theo Coombes, Aarush Katta, Clayton Mullis, Mitchell Wortsman, et~al. 2022.
\newblock {LAION-5B: An open large-scale dataset for training next generation image-text models}.
\newblock In \emph{NeurIPS}.

\bibitem[{Schuhmann et~al.(2021)Schuhmann, Vencu, Beaumont, Kaczmarczyk, Mullis, Katta, Coombes, Jitsev, and Komatsuzaki}]{schuhmann2021laion}
Christoph Schuhmann, Richard Vencu, Romain Beaumont, Robert Kaczmarczyk, Clayton Mullis, Aarush Katta, Theo Coombes, Jenia Jitsev, and Aran Komatsuzaki. 2021.
\newblock {LAION-400M: Open Dataset of CLIP-Filtered 400 Million Image-Text Pairs}.
\newblock In \emph{NeurIPS Workshops}.

\bibitem[{Shao et~al.(2023)Shao, Hu, Gao, Lei, Zhang, Meng, Xu, Huang, Li, Qiao et~al.}]{shao2023tiny}
Wenqi Shao, Yutao Hu, Peng Gao, Meng Lei, Kaipeng Zhang, Fanqing Meng, Peng Xu, Siyuan Huang, Hongsheng Li, Yu~Qiao, et~al. 2023.
\newblock {Tiny LVLM-eHub: Early Multimodal Experiments with Bard}.
\newblock \emph{arXiv preprint arXiv:2308.03729}.

\bibitem[{Sharma et~al.(2018)Sharma, Ding, Goodman, and Soricut}]{sharma2018conceptual}
Piyush Sharma, Nan Ding, Sebastian Goodman, and Radu Soricut. 2018.
\newblock Conceptual captions: A cleaned, hypernymed, image alt-text dataset for automatic image captioning.
\newblock In \emph{ACL}.

\bibitem[{Shukor et~al.(2023)Shukor, Dancette, Rame, and Cord}]{shukor2023unival}
Mustafa Shukor, Corentin Dancette, Alexandre Rame, and Matthieu Cord. 2023.
\newblock {UnIVAL: Unified Model for Image, Video, Audio and Language Tasks}.
\newblock \emph{TMLR}.

\bibitem[{Sidorov et~al.(2020)Sidorov, Hu, Rohrbach, and Singh}]{sidorov2020textcaps}
Oleksii Sidorov, Ronghang Hu, Marcus Rohrbach, and Amanpreet Singh. 2020.
\newblock {TextCaps: A Dataset for Image Captioning with Reading Comprehension}.
\newblock In \emph{ECCV}.

\bibitem[{Singh et~al.(2019)Singh, Natarajan, Shah, Jiang, Chen, Batra, Parikh, and Rohrbach}]{singh2019towards}
Amanpreet Singh, Vivek Natarajan, Meet Shah, Yu~Jiang, Xinlei Chen, Dhruv Batra, Devi Parikh, and Marcus Rohrbach. 2019.
\newblock {Towards VQA Models That Can Read}.
\newblock In \emph{CVPR}.

\bibitem[{Song et~al.(2024)Song, Wang, Sheng, Zhang, Yu, Fan, and Chen}]{song2024moviellm}
Zhende Song, Chenchen Wang, Jiamu Sheng, Chi Zhang, Gang Yu, Jiayuan Fan, and Tao Chen. 2024.
\newblock {MovieLLM: Enhancing Long Video Understanding with AI-Generated Movies}.
\newblock \emph{arXiv preprint arXiv:2403.01422}.

\bibitem[{Su et~al.(2023)Su, Lan, Li, Xu, Wang, and Cai}]{su2023pandagpt}
Yixuan Su, Tian Lan, Huayang Li, Jialu Xu, Yan Wang, and Deng Cai. 2023.
\newblock {PandaGPT: One Model To Instruction-Follow Them All}.
\newblock \emph{arXiv preprint arXiv:2305.16355}.

\bibitem[{Sun et~al.(2023{\natexlab{a}})Sun, Cui, Zhang, Zhang, Yu, Luo, Wang, Rao, Liu, Huang et~al.}]{sun2023generative2}
Quan Sun, Yufeng Cui, Xiaosong Zhang, Fan Zhang, Qiying Yu, Zhengxiong Luo, Yueze Wang, Yongming Rao, Jingjing Liu, Tiejun Huang, et~al. 2023{\natexlab{a}}.
\newblock {Generative Multimodal Models are In-Context Learners}.
\newblock \emph{arXiv preprint arXiv:2312.13286}.

\bibitem[{Sun et~al.(2023{\natexlab{b}})Sun, Yu, Cui, Zhang, Zhang, Wang, Gao, Liu, Huang, and Wang}]{sun2023generative}
Quan Sun, Qiying Yu, Yufeng Cui, Fan Zhang, Xiaosong Zhang, Yueze Wang, Hongcheng Gao, Jingjing Liu, Tiejun Huang, and Xinlong Wang. 2023{\natexlab{b}}.
\newblock {Generative Pretraining in Multimodality}.
\newblock \emph{arXiv preprint arXiv:2307.05222}.

\bibitem[{Tang et~al.(2023)Tang, Yang, Khademi, Liu, Zhu, and Bansal}]{tang2023codi}
Zineng Tang, Ziyi Yang, Mahmoud Khademi, Yang Liu, Chenguang Zhu, and Mohit Bansal. 2023.
\newblock {CoDi-2: In-Context, Interleaved, and Interactive Any-to-Any Generation}.
\newblock \emph{arXiv preprint arXiv:2311.18775}.

\bibitem[{Taori et~al.(2023)Taori, Gulrajani, Zhang, Dubois, Li, Guestrin, Liang, and Hashimoto}]{taori2023stanford}
Rohan Taori, Ishaan Gulrajani, Tianyi Zhang, Yann Dubois, Xuechen Li, Carlos Guestrin, Percy Liang, and Tatsunori~B Hashimoto. 2023.
\newblock {Stanford Alpaca: An Instruction-Following LLaMA Model}.

\bibitem[{Tay et~al.(2022)Tay, Dehghani, Tran, Garcia, Bahri, Schuster, Zheng, Houlsby, and Metzler}]{tay2022unifying}
Yi~Tay, Mostafa Dehghani, Vinh~Q Tran, Xavier Garcia, Dara Bahri, Tal Schuster, Huaixiu~Steven Zheng, Neil Houlsby, and Donald Metzler. 2022.
\newblock {Unifying Language Learning Paradigms}.
\newblock \emph{arXiv preprint arXiv:2205.05131}.

\bibitem[{Tian et~al.(2024{\natexlab{a}})Tian, Zhu, Xiong, Wang, Chen, Wang, Chen, Lu, Lu, Zhou et~al.}]{tian2024mm}
Changyao Tian, Xizhou Zhu, Yuwen Xiong, Weiyun Wang, Zhe Chen, Wenhai Wang, Yuntao Chen, Lewei Lu, Tong Lu, Jie Zhou, et~al. 2024{\natexlab{a}}.
\newblock {MM-Interleaved: Interleaved Image-Text Generative Modeling via Multi-modal Feature Synchronizer}.
\newblock \emph{arXiv preprint arXiv:2401.10208}.

\bibitem[{Tian et~al.(2024{\natexlab{b}})Tian, Ma, Xie, Qiu, Tang, Zhang, Jiao, Tian, and Ye}]{tian2024chatterbox}
Yunjie Tian, Tianren Ma, Lingxi Xie, Jihao Qiu, Xi~Tang, Yuan Zhang, Jianbin Jiao, Qi~Tian, and Qixiang Ye. 2024{\natexlab{b}}.
\newblock {ChatterBox: Multi-round Multimodal Referring and Grounding}.
\newblock \emph{arXiv preprint arXiv:2401.13307}.

\bibitem[{Tian et~al.(2023)Tian, Xie, Wang, Wei, Zhang, Jiao, Wang, Tian, and Ye}]{tian2023integrally}
Yunjie Tian, Lingxi Xie, Zhaozhi Wang, Longhui Wei, Xiaopeng Zhang, Jianbin Jiao, Yaowei Wang, Qi~Tian, and Qixiang Ye. 2023.
\newblock {Integrally Pre-Trained Transformer Pyramid Networks}.
\newblock In \emph{CVPR}.

\bibitem[{Touvron et~al.(2023{\natexlab{a}})Touvron, Lavril, Izacard, Martinet, Lachaux, Lacroix, Rozi{\`e}re, Goyal, Hambro, Azhar et~al.}]{touvron2023llama}
Hugo Touvron, Thibaut Lavril, Gautier Izacard, Xavier Martinet, Marie-Anne Lachaux, Timoth{\'e}e Lacroix, Baptiste Rozi{\`e}re, Naman Goyal, Eric Hambro, Faisal Azhar, et~al. 2023{\natexlab{a}}.
\newblock {LLaMA: Open and Efficient Foundation Language Models}.
\newblock \emph{arXiv preprint arXiv:2302.13971}.

\bibitem[{Touvron et~al.(2023{\natexlab{b}})Touvron, Martin, Stone, Albert, Almahairi, Babaei, Bashlykov, Batra, Bhargava, Bhosale et~al.}]{touvron2023llama2}
Hugo Touvron, Louis Martin, Kevin Stone, Peter Albert, Amjad Almahairi, Yasmine Babaei, Nikolay Bashlykov, Soumya Batra, Prajjwal Bhargava, Shruti Bhosale, et~al. 2023{\natexlab{b}}.
\newblock {Llama 2: Open foundation and fine-tuned chat models}.
\newblock \emph{arXiv preprint arXiv:2307.09288}.

\bibitem[{Van Den~Oord et~al.(2017)Van Den~Oord, Vinyals et~al.}]{van2017neural}
Aaron Van Den~Oord, Oriol Vinyals, et~al. 2017.
\newblock {Neural Discrete Representation Learning}.
\newblock In \emph{NeurIPS}.

\bibitem[{Vaswani et~al.(2017)Vaswani, Shazeer, Parmar, Uszkoreit, Jones, Gomez, Kaiser, and Polosukhin}]{vaswani2017attention}
Ashish Vaswani, Noam Shazeer, Niki Parmar, Jakob Uszkoreit, Llion Jones, Aidan~N Gomez, {\L}ukasz Kaiser, and Illia Polosukhin. 2017.
\newblock Attention is all you need.
\newblock In \emph{NeurIPS}.

\bibitem[{Vedantam et~al.(2015)Vedantam, Lawrence~Zitnick, and Parikh}]{vedantam2015cider}
Ramakrishna Vedantam, C~Lawrence~Zitnick, and Devi Parikh. 2015.
\newblock {CIDEr: Consensus-Based Image Description Evaluation}.
\newblock In \emph{CVPR}.

\bibitem[{Wang et~al.(2023{\natexlab{a}})Wang, Wu, Han, Peng, Zhong, Zhang, Dong, Li, Li, Wang et~al.}]{wang2023vigc}
Bin Wang, Fan Wu, Xiao Han, Jiahui Peng, Huaping Zhong, Pan Zhang, Xiaoyi Dong, Weijia Li, Wei Li, Jiaqi Wang, et~al. 2023{\natexlab{a}}.
\newblock {VIGC: Visual instruction generation and correction}.
\newblock \emph{arXiv preprint arXiv:2308.12714}.

\bibitem[{Wang et~al.(2023{\natexlab{b}})Wang, Ma, Huang, Dong, Wang, Peng, Wu, Bajaj, Singhal, Benhaim et~al.}]{wang2023magneto}
Hongyu Wang, Shuming Ma, Shaohan Huang, Li~Dong, Wenhui Wang, Zhiliang Peng, Yu~Wu, Payal Bajaj, Saksham Singhal, Alon Benhaim, et~al. 2023{\natexlab{b}}.
\newblock {Magneto: A Foundation Transformer}.
\newblock In \emph{ICML}.

\bibitem[{Wang et~al.(2023{\natexlab{c}})Wang, Lv, Yu, Hong, Qi, Wang, Ji, Yang, Zhao, Song et~al.}]{wang2023cogvlm}
Weihan Wang, Qingsong Lv, Wenmeng Yu, Wenyi Hong, Ji~Qi, Yan Wang, Junhui Ji, Zhuoyi Yang, Lei Zhao, Xixuan Song, et~al. 2023{\natexlab{c}}.
\newblock {CogVLM: Visual Expert for Pretrained Language Models}.
\newblock \emph{arXiv preprint arXiv:2311.03079}.

\bibitem[{Wang et~al.(2023{\natexlab{d}})Wang, Shi, Li, Wang, Huang, Xing, Chen, Li, Zhu, Cao et~al.}]{wang2023all}
Weiyun Wang, Min Shi, Qingyun Li, Wenhai Wang, Zhenhang Huang, Linjie Xing, Zhe Chen, Hao Li, Xizhou Zhu, Zhiguo Cao, et~al. 2023{\natexlab{d}}.
\newblock {The All-Seeing Project: Towards Panoptic Visual Recognition and Understanding of the Open World}.
\newblock \emph{arXiv preprint arXiv:2308.01907}.

\bibitem[{Wang et~al.(2023{\natexlab{e}})Wang, Chen, Chen, Wu, Zhu, Zeng, Luo, Lu, Zhou, Qiao et~al.}]{wang2023visionllm}
Wenhai Wang, Zhe Chen, Xiaokang Chen, Jiannan Wu, Xizhou Zhu, Gang Zeng, Ping Luo, Tong Lu, Jie Zhou, Yu~Qiao, et~al. 2023{\natexlab{e}}.
\newblock {VisionLLM: Large Language Model is also an Open-Ended Decoder for Vision-Centric Tasks}.
\newblock \emph{arXiv preprint arXiv:2305.11175}.

\bibitem[{Wang et~al.(2022{\natexlab{a}})Wang, Kordi, Mishra, Liu, Smith, Khashabi, and Hajishirzi}]{wang2022self}
Yizhong Wang, Yeganeh Kordi, Swaroop Mishra, Alisa Liu, Noah~A Smith, Daniel Khashabi, and Hannaneh Hajishirzi. 2022{\natexlab{a}}.
\newblock {Self-instruct: Aligning language model with self generated instructions}.
\newblock \emph{arXiv preprint arXiv:2212.10560}.

\bibitem[{Wang et~al.(2022{\natexlab{b}})Wang, Mishra, Alipoormolabashi, Kordi, Mirzaei, Arunkumar, Ashok, Dhanasekaran, Naik, Stap et~al.}]{wang2022super}
Yizhong Wang, Swaroop Mishra, Pegah Alipoormolabashi, Yeganeh Kordi, Amirreza Mirzaei, Anjana Arunkumar, Arjun Ashok, Arut~Selvan Dhanasekaran, Atharva Naik, David Stap, et~al. 2022{\natexlab{b}}.
\newblock {Super-NaturalInstructions: Generalization via Declarative Instructions on 1600+ NLP Tasks}.
\newblock \emph{arXiv preprint arXiv:2204.07705}.

\bibitem[{Wei et~al.(2023)Wei, Zhang, Zhang, Zhang, and Chu}]{wei2023lenna}
Fei Wei, Xinyu Zhang, Ailing Zhang, Bo~Zhang, and Xiangxiang Chu. 2023.
\newblock {Lenna: Language enhanced reasoning detection assistant}.
\newblock \emph{arXiv preprint arXiv:2312.02433}.

\bibitem[{Wortsman et~al.(2022)Wortsman, Ilharco, Kim, Li, Kornblith, Roelofs, Lopes, Hajishirzi, Farhadi, Namkoong, and Schmidt}]{wortsman2022robust}
Mitchell Wortsman, Gabriel Ilharco, Jong~Wook Kim, Mike Li, Simon Kornblith, Rebecca Roelofs, Raphael~Gontijo Lopes, Hannaneh Hajishirzi, Ali Farhadi, Hongseok Namkoong, and Ludwig Schmidt. 2022.
\newblock {Robust Fine-Tuning of Zero-Shot Models}.
\newblock In \emph{CVPR}.

\bibitem[{Wu et~al.(2023{\natexlab{a}})Wu, Yin, Qi, Wang, Tang, and Duan}]{wu2023visual}
Chenfei Wu, Shengming Yin, Weizhen Qi, Xiaodong Wang, Zecheng Tang, and Nan Duan. 2023{\natexlab{a}}.
\newblock {Visual ChatGPT: Talking, Drawing and Editing with Visual Foundation Models}.
\newblock \emph{arXiv preprint arXiv:2303.04671}.

\bibitem[{Wu et~al.(2023{\natexlab{b}})Wu, Gan, Chen, Wan, and Philip}]{wu2023multimodal}
Jiayang Wu, Wensheng Gan, Zefeng Chen, Shicheng Wan, and S~Yu Philip. 2023{\natexlab{b}}.
\newblock {Multimodal Large Language Models: A Survey}.
\newblock In \emph{BigData}.

\bibitem[{Wu et~al.(2023{\natexlab{c}})Wu, Fei, Qu, Ji, and Chua}]{wu2023next}
Shengqiong Wu, Hao Fei, Leigang Qu, Wei Ji, and Tat-Seng Chua. 2023{\natexlab{c}}.
\newblock {NExT-GPT: Any-to-Any Multimodal LLM}.
\newblock \emph{arXiv preprint arXiv:2309.05519}.

\bibitem[{Wu et~al.(2023{\natexlab{d}})Wu, Biamby, Chan, Dunlap, Gupta, Wang, Gonzalez, and Darrell}]{wu2023see}
Tsung-Han Wu, Giscard Biamby, David Chan, Lisa Dunlap, Ritwik Gupta, Xudong Wang, Joseph~E Gonzalez, and Trevor Darrell. 2023{\natexlab{d}}.
\newblock {See, Say, and Segment: Teaching LMMs to Overcome False Premises}.
\newblock \emph{arXiv preprint arXiv:2312.08366}.

\bibitem[{Xia et~al.(2023{\natexlab{a}})Xia, Wang, Tao, Wang, and Jia}]{xia2023llmga}
Bin Xia, Shiyin Wang, Yingfan Tao, Yitong Wang, and Jiaya Jia. 2023{\natexlab{a}}.
\newblock {LLMGA: Multimodal Large Language Model based Generation Assistant}.
\newblock \emph{arXiv preprint arXiv:2311.16500}.

\bibitem[{Xia et~al.(2023{\natexlab{b}})Xia, Han, Han, Pan, Song, and Huang}]{xia2023gsva}
Zhuofan Xia, Dongchen Han, Yizeng Han, Xuran Pan, Shiji Song, and Gao Huang. 2023{\natexlab{b}}.
\newblock {GSVA: Generalized Segmentation via Multimodal Large Language Models}.
\newblock \emph{arXiv preprint arXiv:2312.10103}.

\bibitem[{Xu et~al.(2023{\natexlab{a}})Xu, Zhou, Yan, Gu, Arnab, Sun, Wang, and Schmid}]{xu2023pixel}
Jiarui Xu, Xingyi Zhou, Shen Yan, Xiuye Gu, Anurag Arnab, Chen Sun, Xiaolong Wang, and Cordelia Schmid. 2023{\natexlab{a}}.
\newblock {Pixel Aligned Language Models}.
\newblock \emph{arXiv preprint arXiv:2312.09237}.

\bibitem[{Xu et~al.(2023{\natexlab{b}})Xu, Wang, Wang, Chen, Pang, and Lin}]{xu2023pointllm}
Runsen Xu, Xiaolong Wang, Tai Wang, Yilun Chen, Jiangmiao Pang, and Dahua Lin. 2023{\natexlab{b}}.
\newblock {PointLLM: Empowering Large Language Models to Understand Point Clouds}.
\newblock \emph{arXiv preprint arXiv:2308.16911}.

\bibitem[{Xu et~al.(2024)Xu, Yao, Guo, Cui, Ni, Ge, Chua, Liu, Sun, and Huang}]{xu2024llava}
Ruyi Xu, Yuan Yao, Zonghao Guo, Junbo Cui, Zanlin Ni, Chunjiang Ge, Tat-Seng Chua, Zhiyuan Liu, Maosong Sun, and Gao Huang. 2024.
\newblock {LLaVA-UHD: an LMM Perceiving Any Aspect Ratio and High-Resolution Images}.
\newblock \emph{arXiv preprint arXiv:2403.11703}.

\bibitem[{Xu et~al.(2023{\natexlab{c}})Xu, Zhang, Xie, Zhao, Guo, Wong, Li, and Zhao}]{xu2023drivegpt4}
Zhenhua Xu, Yujia Zhang, Enze Xie, Zhen Zhao, Yong Guo, Kenneth~KY Wong, Zhenguo Li, and Hengshuang Zhao. 2023{\natexlab{c}}.
\newblock {DriveGPT4: Interpretable End-to-end Autonomous Driving via Large Language Model}.
\newblock \emph{arXiv preprint arXiv:2310.01412}.

\bibitem[{Xuan et~al.(2023)Xuan, Guo, Yang, and Zhang}]{xuan2023pink}
Shiyu Xuan, Qingpei Guo, Ming Yang, and Shiliang Zhang. 2023.
\newblock {Pink: Unveiling the Power of Referential Comprehension for Multi-modal LLMs}.
\newblock \emph{arXiv preprint arXiv:2310.00582}.

\bibitem[{Xue et~al.(2020)Xue, Constant, Roberts, Kale, Al-Rfou, Siddhant, Barua, and Raffel}]{xue2020mt5}
Linting Xue, Noah Constant, Adam Roberts, Mihir Kale, Rami Al-Rfou, Aditya Siddhant, Aditya Barua, and Colin Raffel. 2020.
\newblock {mT5: A massively multilingual pre-trained text-to-text transformer}.
\newblock \emph{arXiv preprint arXiv:2010.11934}.

\bibitem[{Yang et~al.(2022)Yang, Ang, Guo, Zhou, Zhang, and Liu}]{yang2022panoptic}
Jingkang Yang, Yi~Zhe Ang, Zujin Guo, Kaiyang Zhou, Wayne Zhang, and Ziwei Liu. 2022.
\newblock Panoptic scene graph generation.
\newblock In \emph{ECCV}.

\bibitem[{Yang et~al.(2023{\natexlab{a}})Yang, Song, Li, Zhao, Ge, Li, and Shan}]{yang2023gpt4tools}
Rui Yang, Lin Song, Yanwei Li, Sijie Zhao, Yixiao Ge, Xiu Li, and Ying Shan. 2023{\natexlab{a}}.
\newblock {GPT4Tools: Teaching Large Language Model to Use Tools via Self-instruction}.
\newblock \emph{arXiv preprint arXiv:2305.18752}.

\bibitem[{Yang et~al.(2023{\natexlab{b}})Yang, Qu, Lai, Tian, Peng, Liu, and Jia}]{yang2023improved}
Senqiao Yang, Tianyuan Qu, Xin Lai, Zhuotao Tian, Bohao Peng, Shu Liu, and Jiaya Jia. 2023{\natexlab{b}}.
\newblock {LISA++: An Improved Baseline for Reasoning Segmentation with Large Language Model}.
\newblock \emph{arXiv preprint arXiv:2312.17240}.

\bibitem[{Yang et~al.(2023{\natexlab{c}})Yang, Li, Wang, Lin, Azarnasab, Ahmed, Liu, Liu, Zeng, and Wang}]{yang2023mm}
Zhengyuan Yang, Linjie Li, Jianfeng Wang, Kevin Lin, Ehsan Azarnasab, Faisal Ahmed, Zicheng Liu, Ce~Liu, Michael Zeng, and Lijuan Wang. 2023{\natexlab{c}}.
\newblock {MM-REACT: Prompting ChatGPT for Multimodal Reasoning and Action}.
\newblock \emph{arXiv preprint arXiv:2303.11381}.

\bibitem[{Ye et~al.(2023{\natexlab{a}})Ye, Hu, Xu, Ye, Yan, Dan, Zhao, Xu, Li, Tian et~al.}]{ye2023mplugdoc}
Jiabo Ye, Anwen Hu, Haiyang Xu, Qinghao Ye, Ming Yan, Yuhao Dan, Chenlin Zhao, Guohai Xu, Chenliang Li, Junfeng Tian, et~al. 2023{\natexlab{a}}.
\newblock {mPLUG-DocOwl: Modularized Multimodal Large Language Model for Document Understanding}.
\newblock \emph{arXiv preprint arXiv:2307.02499}.

\bibitem[{Ye et~al.(2023{\natexlab{b}})Ye, Hu, Xu, Ye, Yan, Xu, Li, Tian, Qian, Zhang et~al.}]{ye2023ureader}
Jiabo Ye, Anwen Hu, Haiyang Xu, Qinghao Ye, Ming Yan, Guohai Xu, Chenliang Li, Junfeng Tian, Qi~Qian, Ji~Zhang, et~al. 2023{\natexlab{b}}.
\newblock {UReader: Universal OCR-free Visually-situated Language Understanding with Multimodal Large Language Model}.
\newblock In \emph{EMNLP}.

\bibitem[{Ye et~al.(2024)Ye, Yu, Shao, Xie, Torr, and Cao}]{ye2024cat}
Qilang Ye, Zitong Yu, Rui Shao, Xinyu Xie, Philip Torr, and Xiaochun Cao. 2024.
\newblock {CAT: Enhancing Multimodal Large Language Model to Answer Questions in Dynamic Audio-Visual Scenarios}.
\newblock \emph{arXiv preprint arXiv:2403.04640}.

\bibitem[{Ye et~al.(2023{\natexlab{c}})Ye, Xu, Xu, Ye, Yan, Zhou, Wang, Hu, Shi, Shi et~al.}]{ye2023mplug}
Qinghao Ye, Haiyang Xu, Guohai Xu, Jiabo Ye, Ming Yan, Yiyang Zhou, Junyang Wang, Anwen Hu, Pengcheng Shi, Yaya Shi, et~al. 2023{\natexlab{c}}.
\newblock {mPLUG-Owl: Modularization Empowers Large Language Models with Multimodality}.
\newblock \emph{arXiv preprint arXiv:2304.14178}.

\bibitem[{Ye et~al.(2023{\natexlab{d}})Ye, Xu, Ye, Yan, Liu, Qian, Zhang, Huang, and Zhou}]{ye2023mplug2}
Qinghao Ye, Haiyang Xu, Jiabo Ye, Ming Yan, Haowei Liu, Qi~Qian, Ji~Zhang, Fei Huang, and Jingren Zhou. 2023{\natexlab{d}}.
\newblock {mPLUG-Owl2: Revolutionizing Multi-modal Large Language Model with Modality Collaboration}.
\newblock \emph{arXiv preprint arXiv:2311.04257}.

\bibitem[{Yin et~al.(2023{\natexlab{a}})Yin, Fu, Zhao, Li, Sun, Xu, and Chen}]{yin2023survey}
Shukang Yin, Chaoyou Fu, Sirui Zhao, Ke~Li, Xing Sun, Tong Xu, and Enhong Chen. 2023{\natexlab{a}}.
\newblock {A Survey on Multimodal Large Language Models}.
\newblock \emph{arXiv preprint arXiv:2306.13549}.

\bibitem[{Yin et~al.(2023{\natexlab{b}})Yin, Fu, Zhao, Xu, Wang, Sui, Shen, Li, Sun, and Chen}]{yin2023woodpecker}
Shukang Yin, Chaoyou Fu, Sirui Zhao, Tong Xu, Hao Wang, Dianbo Sui, Yunhang Shen, Ke~Li, Xing Sun, and Enhong Chen. 2023{\natexlab{b}}.
\newblock {Woodpecker: Hallucination correction for multimodal large language models}.
\newblock \emph{arXiv preprint arXiv:2310.16045}.

\bibitem[{Yin et~al.(2023{\natexlab{c}})Yin, Qi, Zhu, Chen, Jiang, and Ngo}]{yin2023foodlmm}
Yuehao Yin, Huiyan Qi, Bin Zhu, Jingjing Chen, Yu-Gang Jiang, and Chong-Wah Ngo. 2023{\natexlab{c}}.
\newblock {FoodLMM: A Versatile Food Assistant using Large Multi-modal Model}.
\newblock \emph{arXiv preprint arXiv:2312.14991}.

\bibitem[{Yin et~al.(2023{\natexlab{d}})Yin, Wang, Cao, Shi, Liu, Li, Sheng, Bai, Huang, Wang et~al.}]{yin2023lamm}
Zhenfei Yin, Jiong Wang, Jianjian Cao, Zhelun Shi, Dingning Liu, Mukai Li, Lu~Sheng, Lei Bai, Xiaoshui Huang, Zhiyong Wang, et~al. 2023{\natexlab{d}}.
\newblock {LAMM: Language-Assisted Multi-Modal Instruction-Tuning Dataset, Framework, and Benchmark}.
\newblock In \emph{NeurIPS}.

\bibitem[{You et~al.(2023)You, Zhang, Gan, Du, Zhang, Wang, Cao, Chang, and Yang}]{you2023ferret}
Haoxuan You, Haotian Zhang, Zhe Gan, Xianzhi Du, Bowen Zhang, Zirui Wang, Liangliang Cao, Shih-Fu Chang, and Yinfei Yang. 2023.
\newblock {Ferret: Refer and Ground Anything Anywhere at Any Granularity}.
\newblock \emph{arXiv preprint arXiv:2310.07704}.

\bibitem[{Young et~al.(2014)Young, Lai, Hodosh, and Hockenmaier}]{young2014image}
Peter Young, Alice Lai, Micah Hodosh, and Julia Hockenmaier. 2014.
\newblock {From image descriptions to visual denotations: New similarity metrics for semantic inference over event descriptions}.
\newblock \emph{TACL}, 2:67--78.

\bibitem[{Yu et~al.(2016)Yu, Poirson, Yang, Berg, and Berg}]{yu2016modeling}
Licheng Yu, Patrick Poirson, Shan Yang, Alexander~C Berg, and Tamara~L Berg. 2016.
\newblock {Modeling Context in Referring Expressions}.
\newblock In \emph{ECCV}.

\bibitem[{Yu et~al.(2023{\natexlab{a}})Yu, Shi, Pasunuru, Muller, Golovneva, Wang, Babu, Tang, Karrer, Sheynin et~al.}]{yu2023scaling}
Lili Yu, Bowen Shi, Ramakanth Pasunuru, Benjamin Muller, Olga Golovneva, Tianlu Wang, Arun Babu, Binh Tang, Brian Karrer, Shelly Sheynin, et~al. 2023{\natexlab{a}}.
\newblock {Scaling Autoregressive Multi-Modal Models: Pretraining and Instruction Tuning}.
\newblock \emph{arXiv preprint arXiv:2309.02591}.

\bibitem[{Yu et~al.(2023{\natexlab{b}})Yu, Yang, Li, Wang, Lin, Liu, Wang, and Wang}]{yu2023mm}
Weihao Yu, Zhengyuan Yang, Linjie Li, Jianfeng Wang, Kevin Lin, Zicheng Liu, Xinchao Wang, and Lijuan Wang. 2023{\natexlab{b}}.
\newblock {MM-Vet: Evaluating Large Multimodal Models for Integrated Capabilities}.
\newblock \emph{arXiv preprint arXiv:2308.02490}.

\bibitem[{Yu et~al.(2022)Yu, Tang, Rao, Huang, Zhou, and Lu}]{yu2022point}
Xumin Yu, Lulu Tang, Yongming Rao, Tiejun Huang, Jie Zhou, and Jiwen Lu. 2022.
\newblock {Point-BERT: Pre-Training 3D Point Cloud Transformers With Masked Point Modeling}.
\newblock In \emph{CVPR}.

\bibitem[{Yue et~al.(2023)Yue, Ni, Zhang, Zheng, Liu, Zhang, Stevens, Jiang, Ren, Sun et~al.}]{yue2023mmmu}
Xiang Yue, Yuansheng Ni, Kai Zhang, Tianyu Zheng, Ruoqi Liu, Ge~Zhang, Samuel Stevens, Dongfu Jiang, Weiming Ren, Yuxuan Sun, et~al. 2023.
\newblock {MMMU: A Massive Multi-discipline Multimodal Understanding and Reasoning Benchmark for Expert AGI}.
\newblock \emph{arXiv preprint arXiv:2311.16502}.

\bibitem[{Zang et~al.(2023)Zang, Li, Han, Zhou, and Loy}]{zang2023contextual}
Yuhang Zang, Wei Li, Jun Han, Kaiyang Zhou, and Chen~Change Loy. 2023.
\newblock {Contextual Object Detection with Multimodal Large Language Models}.
\newblock \emph{arXiv preprint arXiv:2305.18279}.

\bibitem[{Zhan et~al.(2024)Zhan, Dai, Ye, Zhou, Zhang, Liu, Zhang, Yuan, Zhang, Li et~al.}]{zhan2024anygpt}
Jun Zhan, Junqi Dai, Jiasheng Ye, Yunhua Zhou, Dong Zhang, Zhigeng Liu, Xin Zhang, Ruibin Yuan, Ge~Zhang, Linyang Li, et~al. 2024.
\newblock {AnyGPT: Unified Multimodal LLM with Discrete Sequence Modeling}.
\newblock \emph{arXiv preprint arXiv:2402.12226}.

\bibitem[{Zhan et~al.(2023)Zhan, Zhu, Chen, Yang, Tang, and Wang}]{zhan2023griffon}
Yufei Zhan, Yousong Zhu, Zhiyang Chen, Fan Yang, Ming Tang, and Jinqiao Wang. 2023.
\newblock {Griffon: Spelling out All Object Locations at Any Granularity with Large Language Models}.
\newblock \emph{arXiv preprint arXiv:2311.14552}.

\bibitem[{Zhang et~al.(2023{\natexlab{a}})Zhang, Zhao, Xie, Zheng, Ji, and Chua}]{zhang2023next}
Ao~Zhang, Liming Zhao, Chen-Wei Xie, Yun Zheng, Wei Ji, and Tat-Seng Chua. 2023{\natexlab{a}}.
\newblock {NExT-Chat: An LMM for Chat, Detection and Segmentation}.
\newblock \emph{arXiv preprint arXiv:2311.04498}.

\bibitem[{Zhang et~al.(2023{\natexlab{b}})Zhang, Li, and Bing}]{zhang2023video}
Hang Zhang, Xin Li, and Lidong Bing. 2023{\natexlab{b}}.
\newblock {Video-LLaMA: An Instruction-tuned Audio-Visual Language Model for Video Understanding}.
\newblock In \emph{EMNLP}.

\bibitem[{Zhang et~al.(2022{\natexlab{a}})Zhang, Li, Liu, Zhang, Su, Zhu, Ni, and Shum}]{zhang2022dino}
Hao Zhang, Feng Li, Shilong Liu, Lei Zhang, Hang Su, Jun Zhu, Lionel~M Ni, and Heung-Yeung Shum. 2022{\natexlab{a}}.
\newblock {DINO: DETR with Improved DeNoising Anchor Boxes for End-to-End Object Detection}.
\newblock \emph{arXiv preprint arXiv:2203.03605}.

\bibitem[{Zhang et~al.(2023{\natexlab{c}})Zhang, Li, Zou, Liu, Li, Yang, and Zhang}]{zhang2023simple}
Hao Zhang, Feng Li, Xueyan Zou, Shilong Liu, Chunyuan Li, Jianwei Yang, and Lei Zhang. 2023{\natexlab{c}}.
\newblock A simple framework for open-vocabulary segmentation and detection.
\newblock In \emph{CVPR}.

\bibitem[{Zhang et~al.(2023{\natexlab{d}})Zhang, Li, Li, Ren, Zou, Liu, Huang, Gao, Zhang, Li et~al.}]{zhang2023llava}
Hao Zhang, Hongyang Li, Feng Li, Tianhe Ren, Xueyan Zou, Shilong Liu, Shijia Huang, Jianfeng Gao, Lei Zhang, Chunyuan Li, et~al. 2023{\natexlab{d}}.
\newblock {LLaVA-Grounding: Grounded Visual Chat with Large Multimodal Models}.
\newblock \emph{arXiv preprint arXiv:2312.02949}.

\bibitem[{Zhang et~al.(2023{\natexlab{e}})Zhang, Mo, Chen, Sun, and Su}]{Zhang2023MagicBrush}
Kai Zhang, Lingbo Mo, Wenhu Chen, Huan Sun, and Yu~Su. 2023{\natexlab{e}}.
\newblock {MagicBrush: A Manually Annotated Dataset for Instruction-Guided Image Editing}.
\newblock In \emph{NeurIPS}.

\bibitem[{Zhang et~al.(2023{\natexlab{f}})Zhang, Wang, Qiao, Gao, and Li}]{zhang2023learning}
Renrui Zhang, Liuhui Wang, Yu~Qiao, Peng Gao, and Hongsheng Li. 2023{\natexlab{f}}.
\newblock {Learning 3D Representations From 2D Pre-Trained Models via Image-to-Point Masked Autoencoders}.
\newblock In \emph{CVPR}.

\bibitem[{Zhang et~al.(2023{\natexlab{g}})Zhang, Sun, Chen, Xiao, Shao, Zhang, Chen, and Luo}]{zhang2023gpt4roi}
Shilong Zhang, Peize Sun, Shoufa Chen, Min Xiao, Wenqi Shao, Wenwei Zhang, Kai Chen, and Ping Luo. 2023{\natexlab{g}}.
\newblock {GPT4RoI: Instruction Tuning Large Language Model on Region-of-Interest}.
\newblock \emph{arXiv preprint arXiv:2307.03601}.

\bibitem[{Zhang et~al.(2022{\natexlab{b}})Zhang, Roller, Goyal, Artetxe, Chen, Chen, Dewan, Diab, Li, Lin et~al.}]{zhang2022opt}
Susan Zhang, Stephen Roller, Naman Goyal, Mikel Artetxe, Moya Chen, Shuohui Chen, Christopher Dewan, Mona Diab, Xian Li, Xi~Victoria Lin, et~al. 2022{\natexlab{b}}.
\newblock {Opt: Open pre-trained transformer language models}.
\newblock \emph{arXiv preprint arXiv:2205.01068}.

\bibitem[{Zhang et~al.(2023{\natexlab{h}})Zhang, Wu, Zhao, Lin, Zhang, Wang, and Xie}]{zhang2023pmc}
Xiaoman Zhang, Chaoyi Wu, Ziheng Zhao, Weixiong Lin, Ya~Zhang, Yanfeng Wang, and Weidi Xie. 2023{\natexlab{h}}.
\newblock {PMC-VQA: Visual Instruction Tuning for Medical Visual Question Answering}.
\newblock \emph{arXiv preprint arXiv:2305.10415}.

\bibitem[{Zhang et~al.(2023{\natexlab{i}})Zhang, Zhang, Gu, Zhou, Lipka, Yang, and Sun}]{zhang2023llavar}
Yanzhe Zhang, Ruiyi Zhang, Jiuxiang Gu, Yufan Zhou, Nedim Lipka, Diyi Yang, and Tong Sun. 2023{\natexlab{i}}.
\newblock {LLaVAR: Enhanced Visual Instruction Tuning for Text-Rich Image Understanding}.
\newblock \emph{arXiv preprint arXiv:2306.17107}.

\bibitem[{Zhao et~al.(2023{\natexlab{a}})Zhao, Wu, and Huang}]{zhao2023svit}
Bo~Zhao, Boya Wu, and Tiejun Huang. 2023{\natexlab{a}}.
\newblock {SVIT: Scaling up Visual Instruction Tuning}.
\newblock \emph{arXiv preprint arXiv:2307.04087}.

\bibitem[{Zhao et~al.(2024)Zhao, Zhang, Zhao, Ding, Huang, and Wang}]{zhao2024cobra}
Han Zhao, Min Zhang, Wei Zhao, Pengxiang Ding, Siteng Huang, and Donglin Wang. 2024.
\newblock {Cobra: Extending Mamba to Multi-Modal Large Language Model for Efficient Inference}.
\newblock \emph{arXiv preprint arXiv:2403.14520}.

\bibitem[{Zhao et~al.(2023{\natexlab{b}})Zhao, Yu, Ge, Yang, Wei, Zhou, Sun, Peng, Dong, Han et~al.}]{zhao2023chatspot}
Liang Zhao, En~Yu, Zheng Ge, Jinrong Yang, Haoran Wei, Hongyu Zhou, Jianjian Sun, Yuang Peng, Runpei Dong, Chunrui Han, et~al. 2023{\natexlab{b}}.
\newblock {ChatSpot: Bootstrapping Multimodal LLMs via Precise Referring Instruction Tuning}.
\newblock \emph{arXiv preprint arXiv:2307.09474}.

\bibitem[{Zhao et~al.(2023{\natexlab{c}})Zhao, Lin, Zhou, Huang, Feng, and Kang}]{zhao2023bubogpt}
Yang Zhao, Zhijie Lin, Daquan Zhou, Zilong Huang, Jiashi Feng, and Bingyi Kang. 2023{\natexlab{c}}.
\newblock {BuboGPT: Enabling Visual Grounding in Multi-Modal LLMs}.
\newblock \emph{arXiv preprint arXiv:2307.08581}.

\bibitem[{Zhao et~al.(2023{\natexlab{d}})Zhao, Guo, Yue, Chen, Shao, Zhu, Yuan, and Liu}]{zhao2023chatbridge}
Zijia Zhao, Longteng Guo, Tongtian Yue, Sihan Chen, Shuai Shao, Xinxin Zhu, Zehuan Yuan, and Jing Liu. 2023{\natexlab{d}}.
\newblock {ChatBridge: Bridging Modalities with Large Language Model as a Language Catalyst}.
\newblock \emph{arXiv preprint arXiv:2305.16103}.

\bibitem[{Zheng et~al.(2023)Zheng, He, and Wang}]{zheng2023minigpt}
Kaizhi Zheng, Xuehai He, and Xin~Eric Wang. 2023.
\newblock {MiniGPT-5: Interleaved Vision-and-Language Generation via Generative Vokens}.
\newblock \emph{arXiv preprint arXiv:2310.02239}.

\bibitem[{Zhong et~al.(2022)Zhong, Yang, Zhang, Li, Codella, Li, Zhou, Dai, Yuan, Li et~al.}]{zhong2022regionclip}
Yiwu Zhong, Jianwei Yang, Pengchuan Zhang, Chunyuan Li, Noel Codella, Liunian~Harold Li, Luowei Zhou, Xiyang Dai, Lu~Yuan, Yin Li, et~al. 2022.
\newblock {RegionCLIP: Region-based Language-Image Pretraining}.
\newblock In \emph{CVPR}.

\bibitem[{Zhu et~al.(2024)Zhu, Jin, Ning, Lin, Huang, Song, Pan, and Yuan}]{zhu2024llmbind}
Bin Zhu, Peng Jin, Munan Ning, Bin Lin, Jinfa Huang, Qi~Song, Mingjun Pan, and Li~Yuan. 2024.
\newblock {LLMBind: A Unified Modality-Task Integration Framework}.
\newblock \emph{arXiv preprint arXiv:2402.14891}.

\bibitem[{Zhu et~al.(2023{\natexlab{a}})Zhu, Chen, Shen, Li, and Elhoseiny}]{zhu2023minigpt}
Deyao Zhu, Jun Chen, Xiaoqian Shen, Xiang Li, and Mohamed Elhoseiny. 2023{\natexlab{a}}.
\newblock {MiniGPT-4: Enhancing Vision-Language Understanding with Advanced Large Language Models}.
\newblock \emph{arXiv preprint arXiv:2304.10592}.

\bibitem[{Zhu et~al.(2023{\natexlab{b}})Zhu, Ding, Ge, Ge, Zhao, Zhao, Wang, and Shan}]{zhu2023vl}
Jinguo Zhu, Xiaohan Ding, Yixiao Ge, Yuying Ge, Sijie Zhao, Hengshuang Zhao, Xiaohua Wang, and Ying Shan. 2023{\natexlab{b}}.
\newblock {VL-GPT: A Generative Pre-trained Transformer for Vision and Language Understanding and Generation}.
\newblock \emph{arXiv preprint arXiv:2312.09251}.

\bibitem[{Zhu et~al.(2023{\natexlab{c}})Zhu, Chen, Ji, Ye, and Liu}]{zhu2023llafs}
Lanyun Zhu, Tianrun Chen, Deyi Ji, Jieping Ye, and Jun Liu. 2023{\natexlab{c}}.
\newblock {LLaFS: When Large-Language Models Meet Few-Shot Segmentation}.
\newblock \emph{arXiv preprint arXiv:2311.16926}.

\bibitem[{Zhu et~al.(2023{\natexlab{d}})Zhu, Hessel, Awadalla, Gadre, Dodge, Fang, Yu, Schmidt, Wang, and Choi}]{zhu2023multimodal}
Wanrong Zhu, Jack Hessel, Anas Awadalla, Samir~Yitzhak Gadre, Jesse Dodge, Alex Fang, Youngjae Yu, Ludwig Schmidt, William~Yang Wang, and Yejin Choi. 2023{\natexlab{d}}.
\newblock Multimodal c4: An open, billion-scale corpus of images interleaved with text.
\newblock \emph{arXiv preprint arXiv:2304.06939}.

\bibitem[{Zhu et~al.(2016)Zhu, Groth, Bernstein, and Fei-Fei}]{zhu2016visual7w}
Yuke Zhu, Oliver Groth, Michael Bernstein, and Li~Fei-Fei. 2016.
\newblock {Visual7W: Grounded Question Answering in Images}.
\newblock In \emph{CVPR}.

\end{thebibliography}

\appendix
\section{Handling Images of Different Resolutions and Aspect Ratios}
\label{sec:high_res_supp}

Most existing MLLMs perceive images in a low resolution and a fixed squared aspect ratio. Some works~\cite{liu2023improved,you2023ferret,chen2023pali} have demonstrated that adopting visual backbones trained on higher resolutions leads to fewer hallucinations and improved multimodal understanding abilities, translating into better performance over tasks that require fine-grained details. However, scaling an MLLM to arbitrary input resolutions and aspect ratios raises two important concerns: (i) the adaptation issue of switching from small images seen during training to larger ones at inference time and (ii) computational costs provided by the increased number of tokens in both the visual encoder and the LLM, given by the quadratic complexity of the attention-based architectures. In the following, we distinguish three different approaches to address these problems.

\tit{Positional-Encoding Interpolation} These models interpolate the positional encoding of their visual backbones, trained at low resolutions, to handle high-resolution images. While being simple, these methods are prone to adaptation issues. As a consequence, they partially mitigate this issue by performing at least one high-resolution training stage. To reduce the input sequence length to the LLM, and thus the computational cost, MiniGPT-v2~\cite{chen2023minigpt} and VILA~\cite{lin2023vila} propose to project multiple visual tokens together into the same token within the embedding space of the LLM. For the same reason, mPLUG-Owl2~\cite{ye2023mplug2} and Qwen-VL~\cite{bai2023qwen} compress the visual features into fixed-length sequences, independent of the resolution, using learnable queries. The latter further saves computation in most layers of the ViT backbone due to a window attention mechanism.

\tit{Sub-Images Slicing} To avoid the adaptation issue, some methods propose to slice a high-resolution image into multiple sub-images of fixed size according to the native resolution of their visual encoder. Then, each sub-image is processed independently by the visual backbone, along with the whole image downsized at the same resolution, and the features are concatenated to obtain the global representation. SPHINX~\cite{lin2023sphinx} divides the input image in a squared grid of sub-images (\ie, $2\times2$ or $3\times3$) at the training resolution of the visual backbone. Moreover, to handle rectangular aspect ratios, SPHINX pads the image to reach the desired square size.
For extreme aspect ratios, the padding leads to sub-images which are only composed of padding. Hence, in SPHINX-X~\cite{gao2024sphinxx} a skip token is introduced to replace noisy tokens associated with only padding sub-images and reduce the sequence length provided to the LLM, increasing efficiency. Similarly, LLaVA-NeXT~\cite{liu2024llava} ignores sub-images composed only of padding and handles different shapes of grids, by introducing a special token that indicates when a row of sub-images ends. Monkey~\cite{li2023monkey} uses a Perceiver-like resampler to extract fixed-length sequences from each sub-image and trains on an image-text dataset curated by several vision expert models integrated by ChatGPT. InfiMM-HD~\cite{liu2024infimm} proposes a dynamic resolution adaptation training stage to increase the image size up to 1,344 pixels. It employs gated cross-attention layers (as in Flamingo) to inject the visual features into the LLM, without increasing the input sequence length. LLaVA-UHD~\cite{xu2024llava} finds the optimal partitioning scheme leading to sub-images that most resemble the native resolution and aspect ratio of the visual encoder. The number of visual tokens is compressed through a Perceiver-like adapter and employs a spatial schema in such a way that the LLM can understand the grid of sub-images.

\tit{Others} Another solution, namely OtterHD~\cite{li2023otterhd}, can seamlessly deal with any resolution or aspect ratio, as it directly feeds large image patches of $30\times30$ pixels to the LLM, without the need for a visual encoder. LLaVA-HR~\cite{luo2024feast}, instead, introduces a mixture-of-resolution adaptation to fuse into the ViT layers high-resolution features extracted with a CNN and low-resolution ones produced by the ViT itself.

\section{Additional Training Data}
\label{sec:training_data_supp}

Specific training datasets are required to empower MLLMs with visual grounding and image generation capabilities. Here we briefly describe the common choices in this domain.

\tit{Visual Grounding} To enable visual grounding, MLLMs can be trained directly on task-specific data using predetermined instruction templates. For instance, CoinIt~\cite{pramanick2023jack} is a unified set of 14 benchmarks converted into an instruction-tuning format, spanning from single-image coarse-level to multi-image region-level tasks. An additional training step is usually performed on an instruction-tuning dataset, such as LLaVa-Instruct~\cite{liu2023llava}, to preserve the conversational capabilities of the MLLM. However, some methods create their custom datasets to simultaneously improve the grounding and conversational capabilities. Specifically, Shikra~\cite{chen2023shikra}, DetGPT~\cite{pi2023detgpt}, ChatSpot~\cite{zhao2023chatspot}, and PVIT~\cite{chen2023position} leverage LLMs~\cite{achiam2023gpt,chatgpt} to combine regions and captions from datasets that present both annotations (\eg, COCO). Differently, Kosmos-2~\cite{peng2023kosmos} and Ferret~\cite{you2023ferret} exploit an open-vocabulary detector~\cite{li2022grounded} to ground noun chunks parsed from captions and then reconstruct referring expressions. ASM~\cite{wang2023all}, GLaMM~\cite{rasheed2023glamm}, and LLaVA-G~\cite{zhang2023llava} propose automated pipelines comprising multiple steps based on off-the-shelf models for generating large corpora of conversations grounded in their corresponding images.

\tit{Image Generation and Editing} To perform image generation, datasets containing both textual captions and images are required, as the one mentioned in Sec.~\ref{sec:Multimodal_Training} (\eg, LAION-400M, COYO-700M, and COCO). To enable interleaved text-image generation, MMC4, OBELICS, and VIST~\cite{huang2016visual} are popular choices. Instead, for image editing tasks, additional datasets like the one introduced in InstructPix2Pix~\cite{brooks2023instructpix2pix} and MagicBrush~\cite{Zhang2023MagicBrush} are typically used.

\begin{table*}
\centering
\setlength{\tabcolsep}{.25em}
\resizebox{\linewidth}{!}{
\begin{tabular}{lc ccccc c cc c ccccccc}
\toprule
& & \multicolumn{5}{c}{\textbf{VQA}} & & \multicolumn{2}{c}{\textbf{Captioning}} & & \multicolumn{7}{c}{\textbf{MLLM Evaluation}} \\
\cmidrule{3-7} \cmidrule{9-10} \cmidrule{12-18}
\textbf{Model} & & VQA$^\text{v2}$ & GQA & VizWiz & SQA & VQA$^\text{T}$ & & COCO & Flickr & & POPE & MME & MMB & SEED & LLaVA$^\text{W}$ & MM-Vet & Math$^\text{V}$ \\
\midrule
Flamingo~\cite{alayrac2022flamingo} & & 82.0 & - & \textbf{65.7} & - & 57.1 & & 138.1 & 75.4 & & - & - & - & - & - & - & - \\
BLIP-2~\cite{li2023blip} & & 65.0 & 41.0 & 19.6 & 61.0 & 42.5 & & \underline{144.5}  & - & & 85.3 & 1293.8 & - & 46.4 & 38.1 & 22.4 & - \\
OpenFlamingo~\cite{awadalla2023openflamingo} & & 52.7 & - & 27.5 & - & 24.2 & & 75.9  & 59.5 & & - & - & - & - & - & - & - \\
MiniGPT-4~\cite{zhu2023minigpt} && 53.7 & 32.2 & - & - & - && - & - && - & 581.7 & 23.0 & 42.8  & 45.1 & 22.1 & 23.1\\
mPLUG-Owl~\cite{ye2023mplug} && 59.5 & 40.9 & - & - & - && -  & - && - & 967.3 & 46.6 & 34.0 & - & - & -  \\
ChatBridge~\cite{zhao2023chatbridge} & & - & 41.8 & - & - & - & & - & 82.5 & & - & - & - & - & - & - & - \\
InstructBLIP~\cite{dai2023instructblip} & & 69.4 & 49.5 & 33.4 & 63.1 & 50.7 & & 102.2 & 82.8 & & 78.9 & 1212.8 & 36.0 & 53.4 & 58.2  & 25.6  & 25.3\\
Shikra~\cite{chen2023shikra} & & 77.4 & - & - & - & - & & 117.5  & - & & - & - & 58.8 & - & \textbf{79.9} & - & - \\
Emu~\cite{sun2023generative} & & 62.0 & 46.0 & 38.3 & - & - & & 117.7  & - & & - & - & - & - & - & 36.3 & - \\
SVIT~\cite{zhao2023svit} & & 80.3 & \underline{64.1} & 56.4 & 70.0 & 60.8 & & - & - & & - & 1565.8 & 69.1 & 61.9 & - & - & - \\
BLIVA~\cite{hu2024bliva} & & - & - & 42.9 & - & 58.0 & & -  & \underline{87.1} & & - & \textbf{1669.2} & - & - & - & - & - \\
IDEFICS~\cite{laurenccon2023obelisc} & & 60.0 & 45.2 & 36.0 & - & 30.9 & & -  & - & & - & - & 54.5 & - & - & - \\
Qwen-VL~\cite{bai2023qwen} & & 78.2 & 57.5 & 38.9 & 68.2 & \underline{61.5} & & 120.2  & 81.0 & & - & 1487.6 & 60.6 & 58.2 & 56.7 & - & - \\
DreamLLM~\cite{dong2023dreamllm} & & 56.6 & - & 38.1 & - & 34.9 & & 115.4  & - & & - & - & 49.9 & - & - & 35.9 & - \\
LLaVA-1.5~\cite{liu2023improved} & & 80.0 & 63.3 & 53.6 & 71.6 & 61.3 & & - & - & & 85.9 & 1531.3 & 67.7 & 61.6 & 70.7 & 35.4 & 23.6 \\
CogVLM~\cite{wang2023cogvlm} & & \underline{82.3} & - & - & - & - & & \textbf{148.7}  & \textbf{94.9} & & 87.9 & - & \textbf{77.6} & \underline{72.5} & \underline{77.8} & \textbf{51.1} & \underline{34.5}\\
LION~\cite{chen2023lion} && - & 51.6 & - & - & - && 139.3  & \underline{87.1} && 88.9 & - & - & - & - & - & - \\
mPLUG-Owl2~\cite{ye2023mplug2} && 79.4 & 56.1 & 54.5 & - & - && 137.3 & - && 86.2 & 1450.2 & 64.5 & 57.8 & 25.0 & 36.2 & 25.3 \\
SPHINX~\cite{lin2023sphinx} &&  80.2 & 62.9  & 46.8 & 69.1 & - && - & - && \textbf{90.8} & 1560.2 & 67.1 & 71.6 & 74.3 & 36.6 & 27.5\\
Emu2~\cite{sun2023generative2} & & \textbf{84.9} & \textbf{65.1} & 54.9 & - & \textbf{66.6} & & -  & - & & - & - & - & 62.8 & - &  \underline{48.5} & - \\
Honeybee~\cite{cha2023honeybee} && - & - & - & - & - && - & - && - & \underline{1632.0} & \underline{73.6} & 68.6 & 77.5 & - & - \\
Unified-IO 2~\cite{lu2023unified} & & 79.4 & - & - & \textbf{88.7} & - & & 125.4  & - & & 87.7 & - & 71.5 & 61.8 & - & - & - \\
VILA~\cite{lin2023vila} & & 80.8 & 63.3 & 60.6 & 73.7 & \textbf{66.6} & & 115.7 & 74.2 & & 84.2 & 1570.1 & 70.3 & 62.8 & 73.0 & 38.8\\
SPHINX-X~\cite{gao2024sphinxx} && 81.1 & 63.8 & \underline{61.9} & \underline{74.5} & - &&  - & - && \underline{89.6} & 1485.3 & 71.3 & \textbf{73.0} & 70.2 & 40.9 & \textbf{42.7}\\
\bottomrule
\end{tabular}
}
\vspace{-.15cm}
\caption{\label{tab:evaluation}
Performance analysis on 14 evaluation benchmarks for VQA, image captioning, and MLLM evaluation. Best scores are in bold, second best are underlined.
\vspace{-.3cm}
}
\end{table*}

\section{Evaluation}
MLLMs are evaluated across different benchmarks, taking into account both more classic visual comprehension and recognition skills and advanced multimodal conversation capabilities. Table~\ref{tab:evaluation} shows the performance of the most common MLLMs on both standard VQA and captioning datasets and benchmarks specifically designed for evaluating MLLMs. In the following, we detail the datasets reported in the table and other benchmarks typically used for the evaluation of MLLMs.

\subsection{Standard Benchmarks}
One of the most important skills of MLLMs is their ability to effectively answer questions based on the given input image. This ability is quantitatively evaluated across several visual question-answering datasets, measuring the accuracy~\citep{antol2015vqa} of the answers provided by the MLLM.

\titt{VQAv2}~\cite{goyal2017making} is an extended and balanced version of VQA~\citep{antol2015vqa} built by collecting similar images for the same question, but whose answer is different compared to the original one. This makes it difficult to perform favorably for those models that ignore visual information and only rely on language priors while answering questions. The reported results are related to the test-dev split.

\titt{GQA}~\cite{hudson2019gqa} is based on Visual Genome scene graph annotations~\citep{krishna2017visual} and comprises 113k images and 22M questions focusing on scene understanding and compositionality. We report the results over the test split, which contains 10\% of the total images.

\titt{OKVQA}~\cite{marino2019ok} is a benchmark to study how vision-and-language models can address visual questions whose answers cannot be completely found in the image, encouraging systems that also rely on external knowledge. The test set has 14,055 open-ended questions.

\titt{VizWiz}~\cite{gurari2018vizwiz} originates from authentic situations involving individuals with visual impairments who have taken images and articulated accompanying inquiries about them, together with 10 responses. The validation split consists of 4,319 images paired with their corresponding questions, while the test split encompasses roughly 8,000 instances.

\titt{ScienceQA (SQA)}~\cite{lu2022learn} evaluates models over challenging multimodal multiple-choice questions about 3 subjects (\ie, natural science, language science, and social science), 26 topics, 127 categories, and 379 skills. Each question is annotated with explanations linked to relevant lectures. The test set includes 4,241 examples.

\titt{Visual Spatial Reasoning (VSR)}~\cite{liu2023vsr} contains images from COCO, each paired with a caption mentioning two concepts and the spatial relation between them. Models have to choose if a given caption is true or false according to the picture. MLLMs are typically evaluated on the 616 samples from the zero-shot test split.

\titt{IconQA}~\cite{lu2021iconqa} tests the visual reasoning abilities of vision-and-language models on three types of questions: multiple-image-choice, multiple-text-choice, and fill-in-the-blank. The dataset stems from real-world problems found in math textbooks and focuses on abstract images (\ie, icons). There are 107,439 questions, 20\% of which makes up for the test split.

\titt{TextVQA (VQA$^\text{T}$)}~\cite{singh2019towards} is a dataset based on pictures from Open Images~\cite{kuznetsova2020open} and challenges OCR capabilities of vision-and-language models. The test set comprises 5,734 examples.

\titt{OCR-VQA}~\cite{mishra2019ocr} presents a new task in visual question answering by interpreting text within images and involves a collection of 207,572 images of book covers, accompanied by more than 1M question-answer pairs.

\medbreak
\noindent Comprehensively describing the visual input is another important skill desired in MLLMs. To evaluate this, various image captioning datasets are commonly employed. As regards the evaluation metric, the CIDEr score~\cite{vedantam2015cider}, which is the reference metric for the task, is used to compare generated image descriptions with ground-truth captions.

\titt{COCO}~\cite{lin2014microsoft} contains more than 120k images, each associated with five human-generated captions. For captioning tasks, the splits defined by~\citet{karpathy2015deep} are typically employed, with 113k, 5k, and 5k images respectively for train, validation and test.

\titt{Flickr30k}~\cite{young2014image} comprises 31,783 images, depicting diverse everyday activities, events, and scenes. Complementing these images are 158,915 captions, obtained through crowd-sourcing techniques. 

\titt{nocaps}~\cite{agrawal2019nocaps} represents a benchmark for novel object captioning, boasting an extensive collection of almost 400 novel object categories compared to the COCO dataset. The validation and test sets include approximately 4.5k and 10.6k images, obtained from Open Images~\cite{kuznetsova2020open}. Each image is annotated with 11 human-generated captions. Both validation and test sets are further categorized into in-domain, near-domain, and out-of-domain, where images from the out-of-domain subset contain object categories that are never present in COCO.

\titt{TextCaps}~\cite{sidorov2020textcaps} includes 145k captions aligned with 28k images. The goal is to recognize and understand the text in images and provide an effective caption that describes the entire visual content. This requires the model to possess OCR capabilities along with image description skills.

\subsection{MLLM-Specific Benchmarks}
Thoroughly evaluating MLLMs is challenging and remains an open frontier. While evaluating on standard datasets represents a valid choice, many benchmarks designed for MLLMs have been recently proposed. They require very strong perception and cognitive skills to succeed, and often they query for deep domain-specific knowledge. To facilitate the evaluation, many works propose to leverage state-of-the-art proprietary models (\eg, ChatGPT~\citep{chatgpt}, GPT-4~\citep{achiam2023gpt}) to automatically judge candidate answers. In Table~\ref{tab:evaluation}, we report the performance of some models on a subset of these new benchmarks.

\titt{POPE}~\cite{li2023evaluating} is a valuable benchmark for evaluating object hallucination challenges within MLLMs. This dataset encompasses several distinct subsets, namely random, popular, and adversarial, which are generated utilizing a variety of sampling methodologies. Cumulatively, it is a binary classification query dataset that comprises 8,910 entries, facilitating comprehensive investigations into the phenomenon of object hallucination within the context of MLLMs.

\titt{MME}~\cite{fu2023mme} is an evaluation benchmark that aims to assess proficiency in various communication modalities through 14 tasks covering comprehension and manipulation across modalities like quantification, spatial determination, color identification, and others. 

\titt{MMBench (MMB)}~\cite{liu2023mmbench} includes approximately 3,000 multiple-choice questions, distributed across 20 distinct domains. Questions are curated to evaluate the efficacy of MLLM across diverse task paradigms. These competencies are systematically arranged into a hierarchical taxonomy, delineating overarching categories such as perception and reasoning, while also outlining granular capabilities including object localization and attribute inference.

\titt{SEED-Bench (SEED)}~\cite{li2023seed} is specifically designed to evaluate LLMs and MLLMs across 12 dimensions spanning from scene understanding to OCR and action recognition. The benchmark consists of 19k multiple-choice questions written by human annotators.

\titt{LLaVA-Bench (LLaVA$^\text{W}$)}~\cite{liu2023visual} comprehends 24 images with 60 manually-curated questions, including indoor and outdoor scenes, memes, paintings, and sketches. GPT-4 is used to generate the reference solutions and score given answers.

\titt{MM-Vet}~\cite{yu2023mm} evaluates MLLMs over 16 tasks covering six fundamental vision-and-language capabilities such as recognition, OCR, knowledge, language generation, spatial awareness, and math. The benchmark comprises 200 images and 218 questions. The evaluation scores are obtained from GPT-4 by few-shot prompting.

\titt{MathVista (Math$^\text{V}$)}~\cite{lu2023mathvista} probes the mathematical reasoning skills of MLLMs for visual question answering. There are 6,141 questions, but only 5,141 are used for evaluation. Before computing the accuracy, the authors propose to parse the answers using an LLM such as GPT-4.

\titt{MMMU}~\cite{yue2023mmmu} is a challenging benchmark targeting domain-specific knowledge of multimodal models. It consists of 10.5k test samples drawn from university textbooks or online courses spanning six main disciplines. Questions may contain multiple images interleaved with text. Exact matching and word matching are used to assess the correctness of an answer for multiple-choice and open-ended questions respectively. Models are evaluated on zero or few-shot settings.

\titt{Tiny LVLM}~\cite{shao2023tiny} focuses on six multimodal capabilities distributed among 2.1k image-question pairs. It introduces a new evaluation metric called ChatGPT ensemble evaluation (CEE). In practice, given the question and the ground-truth solution, ChatGPT is queried with five different prompts to assign the candidate answer either 0 or 1, and the scores are eventually ensembled.

\titt{TouchStone}~\cite{bai2023touchstone} is a visual dialog benchmark with manually annotated open-world images, totaling 908 questions corresponding to five major categories of abilities and 27 sub-tasks. The evaluation score is computed by an LLM such as GPT-4, which is asked to compare a candidate answer with a reference one. The latter is computed by GPT-4 itself, with fine-grained annotations of the query image being part of the prompt.

\subsection{Visual Grounding Evaluation}
The assessment of visual grounding capabilities of MLLMs comprises a variety of standard referring tasks, including region captioning, referring expression generation (REG), and region-level question answering, as well as grounding tasks like referring expression comprehension (REC), referring expression segmentation (RES) and grounded captioning. As regards evaluation metrics, for REC the accuracy is computed by assuming as correct predictions the ones that correspond to an intersection over union with the ground-truth above 0.5 (Acc@0.5). For referring expression segmentation the cumulative intersection over union (cIoU) is considered, while for region captioning METEOR~\cite{banerjee2005meteor} and CIDEr~\cite{vedantam2015cider} are commonly used. However, few methods introduce their own benchmarks to evaluate the performance in more realistic scenarios, with grounded conversations that may involve multiple rounds. Quantitative results on the REC, RES, and region captioning tasks are respectively reported in Table~\ref{tab:evaluation_rec}, Table~\ref{tab:evaluation_res}, and Table~\ref{tab:evaluation_region_cap}.

\begin{table}[t]
\centering
\setlength{\tabcolsep}{.2em}
\resizebox{\linewidth}{!}{
\begin{tabular}{lc ccc c ccc c cc}
\toprule
& & \multicolumn{3}{c}{\textbf{RefCOCO}} & & \multicolumn{3}{c}{\textbf{RefCOCO+}} & & \multicolumn{2}{c}{\textbf{ RefCOCOg}} \\
\cmidrule{3-5} \cmidrule{7-9} \cmidrule{11-12} 
\textbf{Model} & & val & testA & testB & & val & testA & testB & & val(U) & test(U) \\
\midrule
Kosmos-2~\cite{peng2023kosmos} & & 52.3 & 57.4 & 47.3 & & 45.5 & 50.7 & 42.2 & & 60.6 & 61.7 \\
Shikra~\cite{chen2023shikra} & & 87.8 & 91.1 & 81.8 & & 82.9 & 87.8 & 74.4 & & 82.6 & 83.2 \\
Qwen-VL~\cite{bai2023qwen} & & 88.6 & 92.3 & 84.5 & & 82.8 & 88.6 & 76.8 & & 86.0 & 86.3 \\
Ferret~\cite{you2023ferret} & & 89.5 & 92.4 & 84.4 & & 82.8 & 88.1 & 75.2 & & 85.8 & 86.3 \\
MiniGPT-v2~\cite{chen2023minigpt} & & 88.7 & 91.7 & 85.3 & & 80.0 & 85.1 & 74.5 & & 84.4 & 84.7 \\
CogVLM~\cite{wang2023cogvlm} & & \textbf{92.8} & \textbf{94.8} & \textbf{89.0} & & \textbf{88.7} & \textbf{92.9} & \underline{83.4} & & \underline{89.8} & \textbf{90.8} \\
Griffon~\cite{zhan2023griffon} & & 90.1 & 93.4 & 86.1 & & 84.8 & 90.5 & 77.8 & & 86.1 & 87.2 \\
LION~\cite{chen2023lion} & & 89.8 & 93.0 & 85.6 & & 84.0 & 89.2 & 78.1 & & 85.5 & 85.7 \\
NExT-Chat~\cite{zhang2023next} & & 85.5 & 90.0 & 77.9 & & 77.2 & 84.5 & 68.0 & & 80.1 & 79.8 \\
SPHINX~\cite{lin2023sphinx} & & \underline{91.0} & 92.7 & 86.6 & & 86.6 & \underline{91.1} & 80.4 & & 88.2 & 88.4 \\
Lenna~\cite{wei2023lenna} & & 90.3 & 93.2 & \underline{87.0} & & \underline{88.1} & 90.1 & \textbf{84.0} & & \textbf{90.3} & \underline{90.3} \\
LLaVA-G~\cite{zhang2023llava} & & 89.2 & - & - & & 81.7 & - & - & & 84.8 & - \\
Unified-IO 2~\cite{lu2023unified} & & 90.7 & - & - & & 83.1 & - & - & & 86.6 & - \\
MM-Interleaved~\cite{tian2024mm} & & 89.9 & 92.6 & 86.5 & & 83.0 & 88.6 & 77.1 & & 85.2 & 84.9
 \\
SPHINX-X~\cite{gao2024sphinxx} & & 90.6 & \underline{93.7} & 86.9 & & 85.5 & 90.5 & 79.9 & & 88.3 & 88.5 \\
\bottomrule
\end{tabular}
}
\vspace{-.15cm}
\caption{\label{tab:evaluation_rec}
Performance analysis on the RefCOCO benchmarks for referring expression comprehension (REC). Best scores are in bold, second best are underlined.
\vspace{-.3cm}
}
\end{table}

\titt{RefCOCO and RefCOCO+}~\cite{mao2016generation} are collections of referring expressions based on images from the COCO dataset. They were gathered through the ReferItGame~\cite{kazemzadeh2014referitgame}, a two-player game where the first player examines an image featuring a segmented target object and formulates a natural language description referring to that object. The second player, who has access only to the image and the referring expression, selects the corresponding object. Players swap roles if they perform correctly, otherwise they receive a new object and image for description. The RefCOCO dataset has no constraints on the natural language and consists of 142,209 expressions for 50,000 objects across 19,994 images. Instead, in the RefCOCO+ players are disallowed from using location words in their referring expressions and it has 141,564 expressions for 49,856 objects in 19,992 images. Evaluation is performed on 1,500, 750, and 750 images corresponding to the validation, testA, and testB splits for both datasets.

\titt{RefCOCOg}~\cite{yu2016modeling} was collected by a set of annotators who wrote natural language referring expressions for objects in COCO images, and another set of annotators who selected objects corresponding to given referring expressions. When a selected object was correct, the corresponding referring expression was inserted in the dataset. It consists of 85,474 referring expressions for 54,822 objects in 26,711 images. Evaluation is carried out on 1,300 and 2,600 images corresponding to the validation and test splits.

\titt{Visual Genome}~\cite{krishna2017visual} connects structured image concepts to language and comprises 108,077 images along with detailed descriptions of all objects present in them, providing 5.4M region descriptions and 1.7M visual question-answer pairs. This dataset is typically used for region-level captioning and question-answering.

\titt{Visual7W}~\cite{zhu2016visual7w} is a visual question-answering dataset that combines textual descriptions with image regions through object-level grounding. It comprises 328k question-answer pairs on 47k COCO images, together with 1.3M human-generated multiple-choice and more than 560k object groundings from 36,579 categories.

\begin{table}[t]
\centering
\setlength{\tabcolsep}{.25em}
\resizebox{\linewidth}{!}{
\begin{tabular}{lc ccc c ccc c cc}
\toprule
& & \multicolumn{3}{c}{\textbf{RefCOCO}} & & \multicolumn{3}{c}{\textbf{RefCOCO+}} & & \multicolumn{2}{c}{\textbf{RefCOCOg}} \\
\cmidrule{3-5} \cmidrule{7-9} \cmidrule{11-12} 
\textbf{Model} & & val & testA & testB & & val & testA & testB & & val(U) & test(U) \\
\midrule
LISA~\cite{lai2023lisa} & & 74.9 & 79.1 & 72.3 & & 65.1 & 70.8 & 58.1 & & 67.9 & 70.6 \\
GLaMM~\cite{rasheed2023glamm} & & \textbf{79.5} & \textbf{83.2} & \textbf{76.9} & & \textbf{72.6} & \textbf{78.7} & \textbf{64.6} & & \underline{74.2} & \underline{74.9} \\
NExT-Chat~\cite{zhang2023next} & & 74.7 & 78.9 & 69.5 & & 65.1 & 71.9 & 56.7 & & 67.0 & 67.0 \\
GSVA~\cite{xia2023gsva} & & \underline{79.2} & \underline{81.7} & \underline{77.1} & & \underline{70.3} & \underline{73.8} & 63.6 & & \textbf{75.7} & \textbf{77.0} \\
LLaVA-G~\cite{zhang2023llava} & & 77.1 & - & - & & 68.8 & - & - & & 71.5 & - \\
PixelLLM~\cite{xu2023pixel} & & 76.9 & 78.5 & 74.4 & & 69.2 & 72.1 & \underline{64.5} & & 70.7 & 72.4 \\
GELLA~\cite{qi2024generalizable} & & 76.7 & 80.5 & 73.6 & & 67.0 & 73.2 & 60.6 & & 70.4 & 71.5 \\
\bottomrule
\end{tabular}
}
\vspace{-.15cm}
\caption{\label{tab:evaluation_res}
Performance analysis on the RefCOCO benchmarks for referring expression segmentation (RES). Best scores are in bold, second best are underlined.
\vspace{-.3cm}
}
\end{table}

\begin{table}[t]
\centering
\setlength{\tabcolsep}{.35em}
\resizebox{\linewidth}{!}{
\begin{tabular}{lc cc c cc}
\toprule
& & \multicolumn{2}{c}{\textbf{RefCOCO}} & & \multicolumn{2}{c}{\textbf{Visual Genome}} \\
\cmidrule{3-4} \cmidrule{6-7}
\textbf{Model} & & METEOR & CIDEr & & METEOR & CIDEr \\
\midrule
Kosmos-2~\cite{peng2023kosmos} & & 14.1 & 62.3 & & - & - \\
GPT4RoI~\cite{zhang2023gpt4roi} & & - & - & & 17.4 & 145.2 \\
ASM~\cite{wang2023all} & & \textbf{20.8} & \underline{103.0} & & 18.0 & 145.1 \\
GLaMM~\cite{rasheed2023glamm} & & \underline{16.2} & \textbf{106.0} & & \underline{19.7} & \textbf{180.5} \\
NExT-Chat~\cite{zhang2023next} & & 13.6 & 79.6 & & - & - \\
PixelLLM~\cite{xu2023pixel} & & 14.3 & 82.3 & & \textbf{19.9} & \underline{148.9} \\

\bottomrule
\end{tabular}
}
\vspace{-.15cm}
\caption{\label{tab:evaluation_region_cap}
Performance analysis on the RefCOCO and Visual Genome benchmarks for region captioning. Best scores are in bold, second best are underlined.
\vspace{-.3cm}
}
\end{table}

\titt{GRIT}~\cite{peng2023kosmos} is a large-scale dataset of grounded image-text pairs (\ie, noun phrases or referring expressions associated with regions of the image) based on a subset of COYO-700M and LAION-2B. The construction pipeline consists of two steps: (i) extracting noun chunks from the captions and grounding them to bounding boxes with an open-vocabulary detector (\eg, GLIP); (ii) expanding the noun chunks to referring expressions by exploiting their dependency relations in the original caption. The resulting dataset comprises 91M images, 115M text spans, and 137M associated bounding boxes.

\titt{ReasonSeg}~\cite{lai2023lisa} is a benchmark introduced for the reasoning segmentation task, which consists of providing segmentation masks for complex and implicit query texts. Images are from OpenImages~\cite{kuznetsova2020open} and ScanNetv2~\cite{dai2017scannet} and are annotated with text instructions and corresponding segmentation masks. The resulting dataset comprises 1,218 image-instruction pairs. Evaluation metrics are the same as the RES standard benchmark. Two extended variants, ReasonDet~\cite{wei2023lenna} and ReasonSeg-Inst~\cite{yang2023improved}, are respectively introduced for reasoning detection and reasoning instance segmentation tasks.

\titt{Grounding-anything Dataset (GranD)}~\cite{rasheed2023glamm} is a dataset designed for the grounded conversation generation (GCG) task, which aims to construct image-level captions with phrases associated with segmentation masks in the image. This dataset was built with an automated annotation pipeline composed of four stages: (i) object localization with the corresponding semantic label, segmentation mask, attributes, and depth information, (ii) extracting relationships between detected objects, (iii) combining previously collected relations to produce dense captions, (iv) enriching captions with contextual information. It comprises annotations for 11M SAM~\cite{kirillov2023segment} images. Another dataset, $\text{GranD}_f$, is introduced for further fine-tuning and evaluating over the GCG task. It was gathered by extending Flickr30k~\cite{young2014image}, RefCOCOg, and PSG~\cite{yang2022panoptic} through GPT-4 and by manually annotating a set of samples. It comprises 214k image-grounded text pairs with 2.5k validation and 5k test samples. Evaluation metrics include METEOR and CIDEr for captioning, class-agnostic mask AP for grounding, intersection over union for segmentation, and mask recall for grounded captioning.

\titt{Grounded-Bench}~\cite{zhang2023llava} is a benchmark introduced to assess the capabilities of an MLLM in carrying a grounded visual chat. It is built on top of the LLaVA-Bench~\cite{liu2023llava}, comprising conversational data generated with GPT-4 and instance annotations from COCO. It is expanded using 1,000 images with 7,000 entities from COCO annotated through an automated pipeline that involves GPT-4 to associate noun phrases from captions to ground-truth instances.

\titt{MUSE}~\cite{ren2023pixellm} is a multi-target reasoning segmentation dataset. It was created with an automated pipeline on top of 910k instance segmentation masks from the LVIS dataset~\cite{gupta2019lvis} by exploiting GPT-4V to combine instance categories with natural language descriptions. The resulting dataset comprises 246k question-answer pairs, averaging 3.7 targets per answer.

\begin{table}[t]
\centering
\setlength{\tabcolsep}{.4em}
\resizebox{0.95\linewidth}{!}{
\begin{tabular}{lc ccc}
\toprule
& & \multicolumn{3}{c}{\textbf{COCO}} \\
\cmidrule{3-5}
\textbf{Model} & & FID & CLIP-I & CLIP-T \\
\midrule
Stable Diffusion~\cite{rombach2022high} & & \textcolor{gray}{9.22} & \textcolor{gray}{0.667}	& \textcolor{gray}{0.302} \\
Stable Diffusion XL~\cite{podell2023sdxl} & & \textcolor{gray}{-} & \textcolor{gray}{0.674}	& \textcolor{gray}{0.310} \\
\midrule
GILL~\cite{koh2023generating} & & 12.20	& 0.684 & - \\
Emu~\cite{sun2023generative} & & 11.66 & 0.656 & \underline{0.286} \\
SEED~\cite{ge2023planting} & & - & 0.682 & - \\
DreamLLM~\cite{dong2023dreamllm} & & 8.46 & - & - \\
LaVIT~\cite{jin2023unified} & & \textbf{7.40} & - & - \\
NExT-GPT~\cite{wu2023next} & & 11.28 & - & - \\
Kosmos-G~\cite{pan2023kosmos} & & 10.99 & - & - \\
SEED-LLaMa~\cite{ge2023making} & & - & \textbf{0.707} & - \\
Emu2~\cite{sun2023generative2} & & - & \underline{0.686} & \textbf{0.297} \\
VL-GPT~\cite{zhu2023vl} & & 11.53 & - & - \\
Unified-IO 2~\cite{lu2023unified} & & 13.39 & - & - \\
MM-Interleaved~\cite{tian2024mm} & & \underline{7.90} & - & - \\
\bottomrule
\end{tabular}
}
\vspace{-.15cm}
\caption{\label{tab:eval_generation} 
Image generation results on the COCO dataset. Best scores are in bold, second best are underlined.
}
\vspace{-.1cm}
\end{table}

\begin{table}[t]
\centering
\setlength{\tabcolsep}{.4em}
\resizebox{0.9\linewidth}{!}{
\begin{tabular}{lc ccc}
\toprule
& & \multicolumn{3}{c}{\textbf{MagicBrush}} \\
\cmidrule{3-5}
\textbf{Model} & & DINO & CLIP-I & CLIP-T \\
\midrule  
InstructPix2Pix~\cite{brooks2023instructpix2pix} & & \textcolor{gray}{0.698} & \textcolor{gray}{0.854} & \textcolor{gray}{0.292} \\
MagicBrush~\cite{Zhang2023MagicBrush} & & \textcolor{gray}{0.868} & \textcolor{gray}{0.934} & \textcolor{gray}{0.302} \\    
\midrule 	
MGIE~\cite{fu2023guiding} & & \textbf{0.903} & \textbf{0.943} & \textbf{0.317} \\
SmartEdit~\cite{huang2023smartedit} & & 0.815 & 0.914 & 0.305 \\
\bottomrule
\end{tabular}
}
\vspace{-.15cm}
\caption{\label{tab:eval_editing} 
Image editing results on the MagicBrush benchmark.
}
\vspace{-.35cm}
\end{table}

\titt{ChatterBox-300k}~\cite{tian2024chatterbox} is a benchmark established to evaluate models on multimodal dialogue systems in multi-round referring and grounding. The dataset is built on images from Visual Genome~\cite{krishna2017visual} providing bounding boxes, object relationships, and object attributes information to GPT-4 to generate question-answer pairs.

\subsection{Image Generation and Editing Evaluation}
To evaluate image generation and editing results, a set of different benchmarks is usually utilized. In terms of evaluation metrics, Fréchet Inception Distance (FID)~\cite{heusel2017gans} is the reference metric to evaluate generated images. It quantitatively assesses the congruence between the distribution of synthetically generated images and the distribution of real ones. A diminution in the FID score indicates an enhanced alignment between the two distributions, denoting a superior visual quality and realism within the generated images. 

Other metrics measure the coherence of the generated image with the input prompt and the real ground-truth image corresponding to it. Specifically, CLIP-I and DINO scores consist of computing the cosine similarity between generated and ground-truth images leveraging CLIP~\cite{radford2021learning} and DINO~\cite{caron2021emerging} as visual backbones. CLIP-T, instead, measures image-text alignment through cosine similarity between input captions and generated images, using CLIP to encode both images and textual prompts. 

\begin{table}[t]
\centering
\setlength{\tabcolsep}{.4em}
\resizebox{0.85\linewidth}{!}{
\begin{tabular}{lc ccc}
\toprule
& & \multicolumn{3}{c}{\textbf{DreamBench}} \\
\cmidrule{3-5}
\textbf{Model} & & DINO & CLIP-I & CLIP-T \\
\midrule
DreamBooth~\cite{ruiz2023dreambooth} & & \textcolor{gray}{0.668} & \textcolor{gray}{0.803} & \textcolor{gray}{0.305} \\
\midrule
Kosmos-G~\cite{pan2023kosmos} & & 0.694 & 0.847 & 0.287 \\
CoDi-2~\cite{tang2023codi} & & 0.703 & \textbf{0.852} & \textbf{0.311} \\
Emu2~\cite{sun2023generative2} & & \textbf{0.766} & 0.850 & 0.287 \\
\bottomrule
\end{tabular}
}
\vspace{-.15cm}
\caption{\label{tab:eval_driven_gen} 
Subject-driven image generation results on the DreamBench dataset.
}
\vspace{-.35cm}
\end{table}

\titt{COCO} is employed for evaluating text-to-image generation. The evaluation is conducted using either the original validation set comprising 41k samples or a subset of 30k samples randomly selected from the same set. Results on this dataset of MLLMs with image generation capabilities are reported in Table~\ref{tab:eval_generation}.

\titt{VIST}~\cite{huang2016visual} is specifically curated for the task of interleaved image-text generation. It includes 34k and 5k samples for training and evaluation. Each sample is a sequence consisting of 5 images accompanied by 5 textual narratives that collectively form a coherent story.

\titt{MagicBrush}~\cite{Zhang2023MagicBrush} is a benchmark in the area of image editing and contains a collection of 10,000 manually annotated triplets, each consisting of a source image, an editing instruction, and the corresponding target image. Performances on this benchmark are reported in Table~\ref{tab:eval_editing}.

\titt{DreamBench}~\cite{ruiz2023dreambooth} is a benchmark that evaluates the generative capabilities of the models on subject-driven generation. Specifically, it contains 30 subjects, each illustrated with 4 to 6 images, and 25 template prompts enabling modification and accessorization of the given subjects. Results on this benchmark are shown in Table~\ref{tab:eval_driven_gen}.

\begin{table}[t]
\centering
\setlength{\tabcolsep}{.35em}
\resizebox{0.92\linewidth}{!}{
\begin{tabular}{l cc cl cl}
\toprule
\textbf{Model} & \textbf{Hardware Type} & \textbf{\#} \\
\midrule
Flamingo\cite{alayrac2022flamingo} & TPUv4 & 1,535 \\
PaLI~\cite{chen2023pali} & TPUv4 & 1,024 \\
IDEFICS~\cite{laurenccon2023obelisc} & A100 & 512\\
SPHINX~\cite{lin2023sphinx} & A100 & 32\\
Emu~\cite{sun2023generative} & A100 & 128\\
VILA~\cite{lin2023vila} & A100 & 128\\
BLIP-2~\cite{li2023blip} & A100 & 16\\
SEED-LLaMA~\cite{ge2023making} & A100 & 64\\
Shikra~\cite{chen2023shikra} & A100 & 8\\
MiniGPT-v2~\cite{chen2023minigpt} & A100 & 8\\
InstructBLIP~\cite{dai2023instructblip} & A100 & 16\\
BLIVA~\cite{hu2024bliva} & A6000 & 8\\
CleverFlamingo~\cite{chen2023visual} & A100 & 8\\
LLaVA 1.5~\cite{liu2023improved} & A100 & 8\\
LLaVA~\cite{liu2023visual} & A100 & 8\\
MiniGPT-4~\cite{zhu2023minigpt} & A100 & 4\\
FROMAGe~\cite{koh2023grounding} & A100 & 1\\
LaVIN~\cite{luo2023cheap} & A100 & 8 \\
\bottomrule
\end{tabular}
}
\vspace{-0.15cm}
\caption{\label{tab:gpu_2}
Summary of the hardware required to train common MLLMs.}
\vspace{-.3cm}
\end{table}

\section{Computational Requirements}
To provide a quantification of the computational requirements necessary to train an MLLM, we compare some of the most common models in Table~\ref{tab:gpu_2} and indicate for each of them the type and number of GPUs/TPUs employed during training. Except for Flamingo and PaLI, which are trained on a large amount of TPUs, all other models employ A100 or A6000 GPUs. As it can be seen, most MLLMs distribute training across 8 A100s. 

\begin{figure}[t]
\centering
\includegraphics[width=0.95\linewidth]{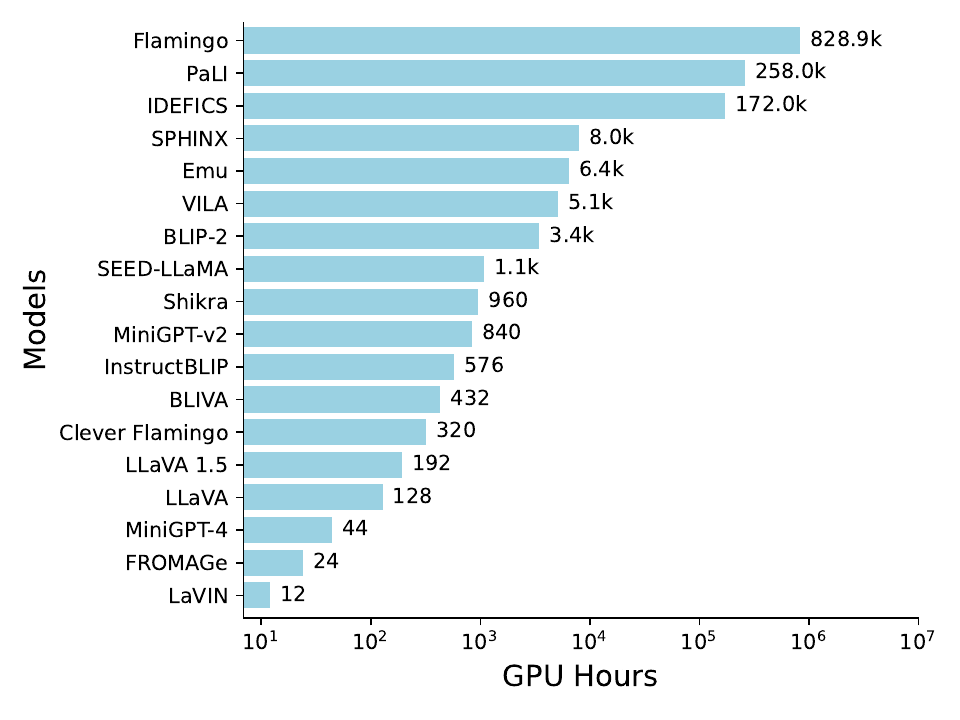}
\vspace{-.2cm}
\caption{Number of GPU training hours for various MLLMs. Here 1 TPU hour is approximated as 1.5 GPU hours following public benchmarks.}
\label{fig:gpus}
\vspace{-.35cm}
\end{figure}

Moreover, in Figure~\ref{fig:gpus} we show for each MLLM the total amount of GPU training hours, approximating 1 TPU hour as 1.5 GPU hours. Notably, models like Flamingo, PaLI, and IDEFICS require a significant amount of GPU time (in the order of magnitude of a few hundred thousand GPU hours). Instead, lighter models like LLaVA only require a few hundred GPU hours to complete training.

\begin{table*}
\centering
\setlength{\tabcolsep}{.35em}
\resizebox{\linewidth}{!}{
\begin{tabular}{l cc cl cl}
\toprule
\textbf{Model} & \textbf{LLM} & \textbf{Visual Encoder} & & \textbf{Main Tasks \& Capabilities} \\
\midrule
VideoChat~\cite{li2023videochat} & StableVicuna-13B$^\bigstar$ & EVA ViT-g & & Visual Dialogue, VQA, Captioning \\
Video-ChatGPT~\cite{maaz2023video} & Vicuna-7B$^\bigstar$ & CLIP ViT-L & & Visual Dialogue, VQA, Captioning \\
Video-LLaMA~\cite{zhang2023video} & Vicuna-7B$^\bigstar$ & EVA ViT-g & & Visual Dialogue, Captioning, VQA, Audio Understanding \\
BT-Adapter~\cite{liu2023one} & Vicuna-7B$^\bigstar$ & CLIP ViT-L & & Visual Dialogue, Captioning, VQA, Retrieval \\ 
LLaMA-VID~\cite{li2023llama} & Vicuna-13B$^\blacklozenge$ & EVA ViT-g & & Visual Dialogue, VQA, Captioning \\
PG-Video-LLaVA~\cite{munasinghe2023pg} & LLaVA-1.5-13B$^\bigstar$ & CLIP ViT-L & & Visual Dialogue, Captioning, VQA, Grounding \\ 
TimeChat~\cite{ren2023timechat} & LLaMA-2-7B$^\blacktriangle$ & EVA ViT-g & & Visual Dialogue, Captioning, Temporal Grounding, Highlight Detection \\
Vista-LLaMA~\cite{ma2023vista} & Vicuna-7B$^\bigstar$ & EVA ViT-g & & Visual Dialogue, VQA, Captioning \\
\bottomrule
\end{tabular}
}
\vspace{-0.15cm}
\caption{\label{tab:video}
Summary of video-based MLLMs. For each model, we indicate the LLM used in its best configuration, in some cases initialized with the weights of a pre-trained MLLM ($\bigstar$: frozen LLM; $\blacklozenge$: LLM fine-tuning; $\blacktriangle$: LLM fine-tuning with PEFT techniques).
}
\vspace{-.3cm}
\end{table*}

\section{Additional Details on Other Modalities and Applications} 

\tinytit{Video Understanding} As a complement of Sec.~\ref{sec:others}, we report in Table~\ref{tab:video} a summary of the main characteristics of video-based MLLMs. For each model, we indicate the LLM used as starting point, which in some cases is initialized with the parameters of a pre-trained MLLM, the visual encoder, and the main tasks and capabilities of the MLLM. Additionally, we specify whether the LLM is kept frozen, is entirely fine-tuned, or is fine-tuned with PEFT-based strategies.  

\tit{3D Understanding}
MLLMs are also applied to 3D data for solving complex tasks like 3D VQA, 3D conversation, and 3D dense captioning. Differently from standard visual encodings which exploit 2D pre-trained embeddings, in the context of 3D data, appropriate strategies are designed to project them to the LLM space. In 3D-LLM~\cite{hong20233d}, 3D scenes are rendered in different views and 3D features are built using an EVA-CLIP backbone connected to a fine-tuned BLIP-2 model. Similarly,~\citet{xu2023pointllm} employ a pre-trained Point-BERT~\cite{yu2022point} as 3D encoder and conducts a two-stage training that initially aligns the input features via an MLP projection layer, and then performs an instruction tuning phase of the model. Differently, in Point-Bind~\cite{guo2023point}, 3D point-clouds are aligned with ImageBind~\cite{girdhar2023imagebind} and by leveraging I2P-MAE~\cite{zhang2023learning} as 3D encoder. This alignment allows the introduction of new tasks such as any-to-3D generation and 3D embedding-space arithmetic. Recently, LL3DA~\cite{chen2023ll3da} introduces the Interactor3D module, which consists of a frozen 3D scene encoder, a visual prompt encoder, and a Q-Former to address 3D captioning and VQA. 

\begin{table*}
\centering
\setlength{\tabcolsep}{.25em}
\resizebox{\linewidth}{!}{
\begin{tabular}{l cc cl}
\toprule
\textbf{Model} & \textbf{LLM} & \textbf{Visual Encoder} & & \textbf{Main Tasks \& Capabilities} \\
\midrule
\textit{Document Analysis} \\
\hspace{0.3cm}mPLUG-DocOwl~\cite{ye2023mplugdoc} & mPLUG-Owl-7B$^\blacktriangle$ & CLIP ViT-L & & Visual Dialogue, Captioning, VQA \\
\hspace{0.3cm}Kosmos-2.5~\cite{lv2023kosmos} & Magneto-1.3B$^\blacklozenge$ & Pix2Struct ViT-L & & Text Recognition, Image-to-Markdown Generation\\
\hspace{0.3cm}UReader~\cite{ye2023ureader} & mPLUG-Owl-7B$^\blacktriangle$ & CLIP VIT-L & & Visual Dialogue, VQA, Captioning, Information Extraction \\
\hspace{0.3cm}mPLUG-PaperOwl~\cite{hu2023mplugpaper} & mPLUG-Owl-7B$^\blacktriangle$ & CLIP ViT-L & & Visual Dialogue, VQA, Captioning, Diagram Analysis \\
\rowcolor{Gray}
\hspace{0.3cm}LLaMA-SciTune~\cite{horawalavithana2023scitune} & LLaMA-13B$^\blacklozenge$ & CLIP ViT-L & & Visual Dialogue, VQA, Captioning, Diagram Analysis \\
\rowcolor{Gray}
\hspace{0.3cm}DocPedia~\cite{feng2023docpedia} & Vicuna-7B$^\blacklozenge$ & Swin-B & & Visual Dialogue, VQA, Information Extraction \\
\midrule
\textit{Embodied AI} \\
\hspace{0.3cm}EmbodiedGPT~\cite{mu2023embodiedgpt} & LLaMA-7B$^\bigstar$ & EVA ViT/g, RN50 & & Visual Dialogue, VQA, Captioning, Task Planning \\
\rowcolor{Gray}
\hspace{0.3cm}PaLM-E~\cite{driess2023palm} & PaLM-540B$^\blacklozenge$ & ViT-22B & & Visual Dialogue, VQA, Captioning, Task Planning, Manipulation\\
\midrule
\textit{Medical Vision Learning} \\
\hspace{0.3cm}PMC-VQA~\cite{zhang2023pmc} & PMC-LLaMA-7B$^\bigstar$ & PMC-CLIP RN50 & & VQA \\
\hspace{0.3cm}LLaVA-Med~\cite{li2023llava} & LLaVA-7B$^\blacklozenge$ & CLIP ViT-L & & Visual Dialogue, VQA \\
\hspace{0.3cm}Qilin-Med-VL~\cite{liu2023qilin} &  CN-LLaMA2-13B$^\blacklozenge$ & CLIP ViT-L & & Visual Dialogue, VQA \\
\midrule 
\textit{Autonomous Driving} \\
\hspace{0.3cm}Dolphins~\cite{ma2023dolphins} & OpenFlamingo-7B$^\blacktriangle$ & CLIP ViT-L & & Visual Dialogue, VQA, Captioning, Traffic Condition Understanding \\
\rowcolor{Gray}
\hspace{0.3cm}DriveGPT4~\cite{xu2023drivegpt4} & LLaMA-2-7B$^\blacklozenge$ & CLIP ViT-L & & Visual Dialogue, VQA, Captioning \\
\midrule 
\textit{Food Understanding} \\
\hspace{0.3cm}FoodLLM~\cite{yin2023foodlmm} & LISA-7B$^\blacktriangle$ & CLIP ViT-L & & Visual Dialogue, VQA, Nutrition Estimation, RES \\
\bottomrule
\end{tabular}
}
\vspace{-0.15cm}
\caption{\label{tab:models_domains}
Summary of MLLMs designed for domain-specific applications. For each model, we indicate the LLM used in its best configuration, in some cases initialized with the weights of a pre-trained MLLM ($\bigstar$: frozen LLM; $\blacklozenge$: LLM fine-tuning; $\blacktriangle$: LLM fine-tuning with PEFT techniques). Gray color indicates models not publicly available. 
}
\vspace{-.3cm}
\end{table*}

\tit{Any-Modality Models}
Several studies focus on extending the reasoning capabilities of the MLLMs by including multiple modalities, such as video, 3D, and audio. A line of research investigates the usage of dedicated pathways to input the different modalities to the LLM. UniVAL~\cite{shukor2023unival} maps the features from each modality encoder into the shared representation space of the LLM through dedicated linear projections. X-LLM~\cite{chen2023x} leverages Q-Former interfaces for the image and video modalities, interpreting the video as a sequence of independent frames, each one encoded as an image. For the speech modality, it uses a C-Former interface that compresses the feature sequence from the speech encoder into token-level embeddings. X-InstructBLIP~\cite{panagopoulou2023x} and ChatBridge~\cite{zhao2023chatbridge} propose to freeze both the modality encoders and the LLM and to leverage, respectively, dedicated Q-Former or Perceiver adapters for each modality. To maximize feature compatibility, AnyMAL~\cite{moon2023anymal} uses an encoder that has already been aligned to a text embedding space for each modality, including also IMU signals, and a dedicated adapter, which is a Perceiver for the visual modality and linear layers for the others. On the other hand, PandaGPT~\cite{su2023pandagpt} and NExT-GPT~\cite{wu2023next} exploit a single frozen multimodal encoder (\ie, ImageBind) to extract features from different modalities. OneLLM~\cite{han2024onellm} builds a unified universal encoder and a universal projection module by mixing multiple image projection modules and a modality router to align input signals with language. CAT~\cite{ye2024cat} adds a clue aggregator to aggregate question-aware audio-visual hidden features and produce clue tokens that are provided to the LLM.

In addition to handling different modalities in input to the LLM, some works investigate the generation of outputs of different modalities. For example, NExT-GPT~\cite{wu2023next} introduces signal tokens in the LLM that indicate whether the diffusion-based decoder for a specific modality has to be activated. Moreover, the signal tokens are provided to a transformer-based output projector to condition the generation. Similarly, M2UGen~\cite{hussain2023m} handles the music modality by using the LLM output corresponding to signal tokens, along with unimodal music features from a music encoder, to condition the generation of an audio encoder. LLMBind~\cite{zhu2024llmbind} indicates the conditioning text to generate image, video, or audio by wrapping it in special tokens. Thus, this text is provided to the corresponding modality-specific diffusion model. Unified-IO2~\cite{lu2023unified} uses VQ-GAN decoders for both image and audio modalities to decode output discrete tokens and can generate surface normals, depth, and segmentation masks for the input images. AnyGPT~\cite{zhan2024anygpt} interprets all the continuous non-text modalities as discrete tokens in both input and output, using, respectively, multimodal tokenizers and de-tokenizers. To enable the 3D modality, LAMM~\cite{yin2023lamm} introduces a novel instruction tuning dataset and benchmark that comprise both image-text and point cloud-text instruction-response pairs, covering a wide range of 2D and 3D tasks.

\tit{Interactive and Compositional Systems}
A different trend is to build systems that can combine multiple tools (\ie, existing vision-only or vision-and-language models), usually through ChatGPT or another LLM. In particular, these approaches aim to let the user interact with the LLM which is in charge of selecting the useful tools to carry out complex tasks. In this context, some solutions study how to prompt ChatGPT~\cite{wu2023visual,yang2023mm} to invoke visual foundation models. GPT4Tools~\cite{yang2023gpt4tools}, instead, employs open-source LLMs such as LLaMA and OPT, that are fine-tuned with PEFT techniques to use tools for performing a wide range of visual tasks. Differently,~\citet{liu2023interngpt} introduce more sophisticated user-chatbot interactions, through the incorporation of mouse-based pointing instructions on images or videos. 

While in all these approaches the LLM does not directly handle the visual input which is instead processed by other external tools, in LLaVA-Plus~\cite{liu2023llava} the query image is directly input to the MLLM (\ie, LLaVA) and is therefore involved during the selection and invocation of the most helpful tool according to the user needs. This is achieved also thanks to the introduction of a new instruction-following use tool dataset, which is employed to fine-tune the MLLM.

\tit{Domain-Specific MLLMs} Finally, in Table~\ref{tab:models_domains} we summarize the main characteristics of domain-specific MLLMs, also in this case indicating for each model the LLM used as starting point.

\end{document}